\documentclass[journal,transmag]{IEEEtran}
\usepackage{times}
\usepackage{graphicx} 
\usepackage{enumerate}
\usepackage{subcaption}
\usepackage{amssymb}
\usepackage{amsmath}
\usepackage{algorithm}
\usepackage{algorithmic}

\usepackage{multirow}
\usepackage{bm}
\usepackage[framed,numbered]{mcode}
\usepackage{hyperref}
\hypersetup{pdfborder={0 0 0}, colorlinks=true,linkcolor=red}

\newcommand{\sumi}{\sum_{i=1}^m}
\newcommand{\sumj}{\sum_{j=1}^p}

\newtheorem{theo}{Theorem}
\newtheorem{lem}{Lemma}
\newtheorem{definition}{Definition}

\DeclareMathOperator*{\Diag}{Diag}
\DeclareMathOperator*{\X}{{\mathbf{X}}}
\DeclareMathOperator*{\Xt}{\mathcal{X}}

\DeclareMathOperator*{\M}{\mathbf{M}}
\DeclareMathOperator*{\Y}{\mathbf{Y}}
\DeclareMathOperator*{\x}{\mathbf{x}}
\DeclareMathOperator*{\y}{\mathbf{y}}

\DeclareMathOperator*{\PP}{{\mathbf{P}}}

\begin{document}
\title{Nonconvex Nonsmooth Low-Rank Minimization via Iteratively Reweighted Nuclear Norm}
\author{Canyi Lu,~
	Jinhui Tang,~\IEEEmembership{Senior Member,~IEEE},
Shuicheng Yan,~\IEEEmembership{Senior Member,~IEEE},\\
        and~Zhouchen Lin,~\IEEEmembership{Senior Member,~IEEE}
        
\thanks{C. Lu and S. Yan are with the Department of Electrical and Computer Engineering, National University of Singapore, Singapore (e-mail: canyilu@gmail.com; eleyans@nus.edu.sg).}
\thanks{J. Tang is with the School of Computer Science, Nanjing University of Science and Technology, China (e-mail: jinhuitang@mail.njust.edu.cn).}
\thanks{Z. Lin is with the Key Laboratory of Machine Perception (MOE), School of EECS, Peking University, China (e-mail: zlin@pku.edu.cn).}
\thanks{This paper is an extended version of \cite{lu2014generalized} published in CVPR 2014.}}


\maketitle
\begin{abstract}
	The nuclear norm is widely used as a convex surrogate of the rank function in compressive sensing for low rank matrix recovery with its applications in image recovery and signal processing. However, solving the nuclear norm based relaxed convex problem usually leads to a suboptimal solution of the original rank minimization problem. In this paper, we propose to perform a family of nonconvex surrogates of $L_0$-norm on the singular values of a matrix to approximate the rank function. This leads to a nonconvex nonsmooth minimization problem. Then we propose to solve the problem by Iteratively Reweighted Nuclear Norm (IRNN) algorithm. IRNN iteratively solves a Weighted Singular Value Thresholding (WSVT) problem, which has a closed form solution due to the special properties of the nonconvex surrogate functions. We also extend IRNN to solve the nonconvex problem with two or more blocks of variables. In theory, we prove that IRNN decreases the objective function value monotonically, and any limit point is a stationary point. Extensive experiments on both synthesized data and real images demonstrate that IRNN enhances the low-rank matrix recovery compared with state-of-the-art convex algorithms.
\end{abstract}
\IEEEpeerreviewmaketitle

\IEEEpeerreviewmaketitle

\section{Introduction}
\label{sec:intro}

\IEEEPARstart{B}ENEFITING from the success of Compressive Sensing (CS) \cite{candes2008introduction}, the {sparse} and {low rank} matrix structures have attracted considerable research interests from the computer vision and machine learning communities. There have been many applications which exploit these two structures. For instance, sparse coding has been widely used for face recognition \cite{SRC}, image classification \cite{wang2010locality} and super-resolution \cite{yang2010image}, while low rank models are applied for background modeling \cite{RPCA}, motion segmentation \cite{robustlrr,lu2013correlation} and collaborative filtering \cite{candes2009exact}. 

\begin{figure*}
	\begin{subfigure}[b]{0.235\textwidth}
		\centering
		\includegraphics[width=\textwidth]{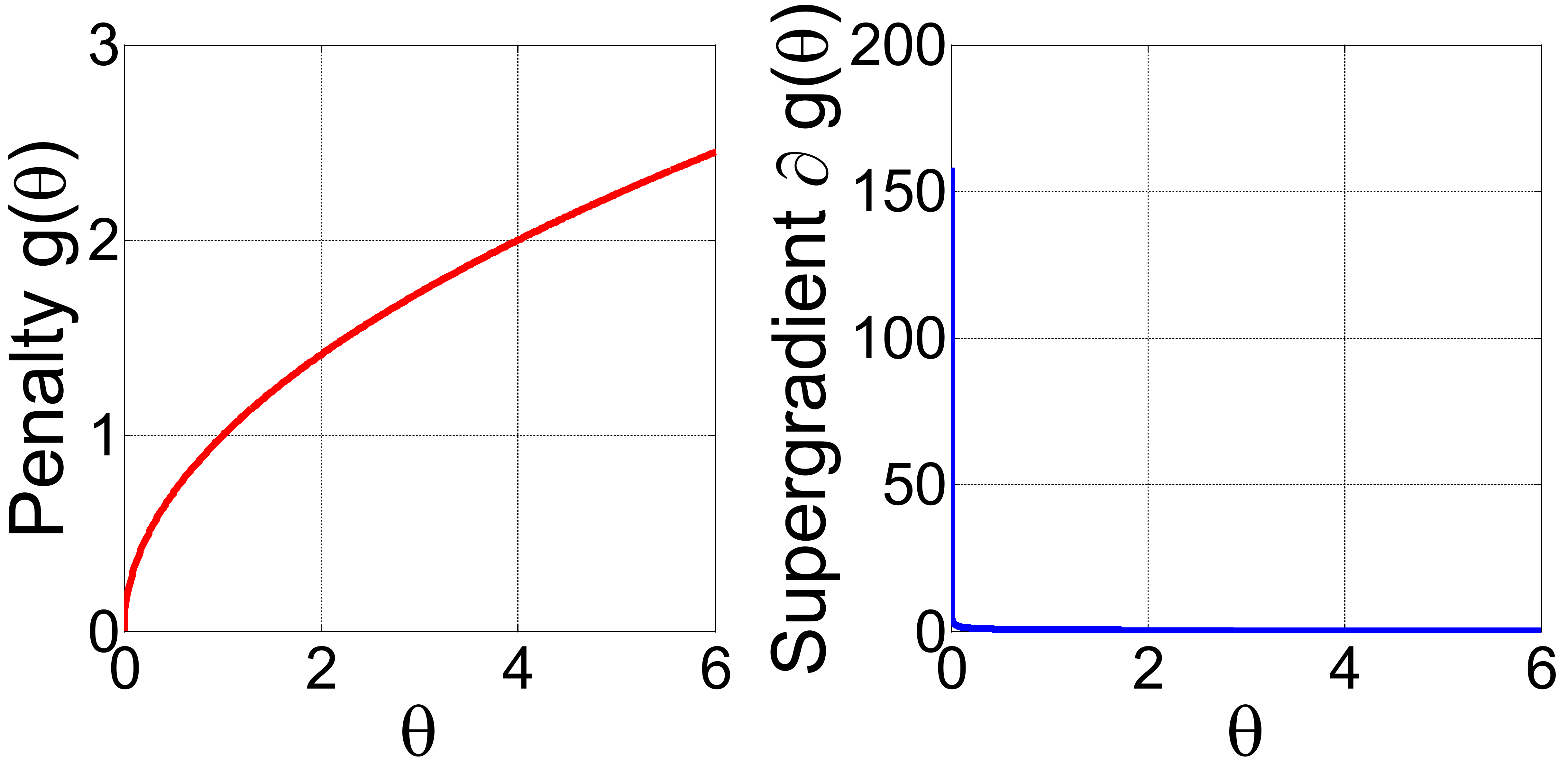}
		\caption{\scriptsize{$L_p$ Penalty \cite{frank1993statistical}}}
	\end{subfigure}
	\begin{subfigure}[b]{0.235\textwidth}
		\centering
		\includegraphics[width=\textwidth]{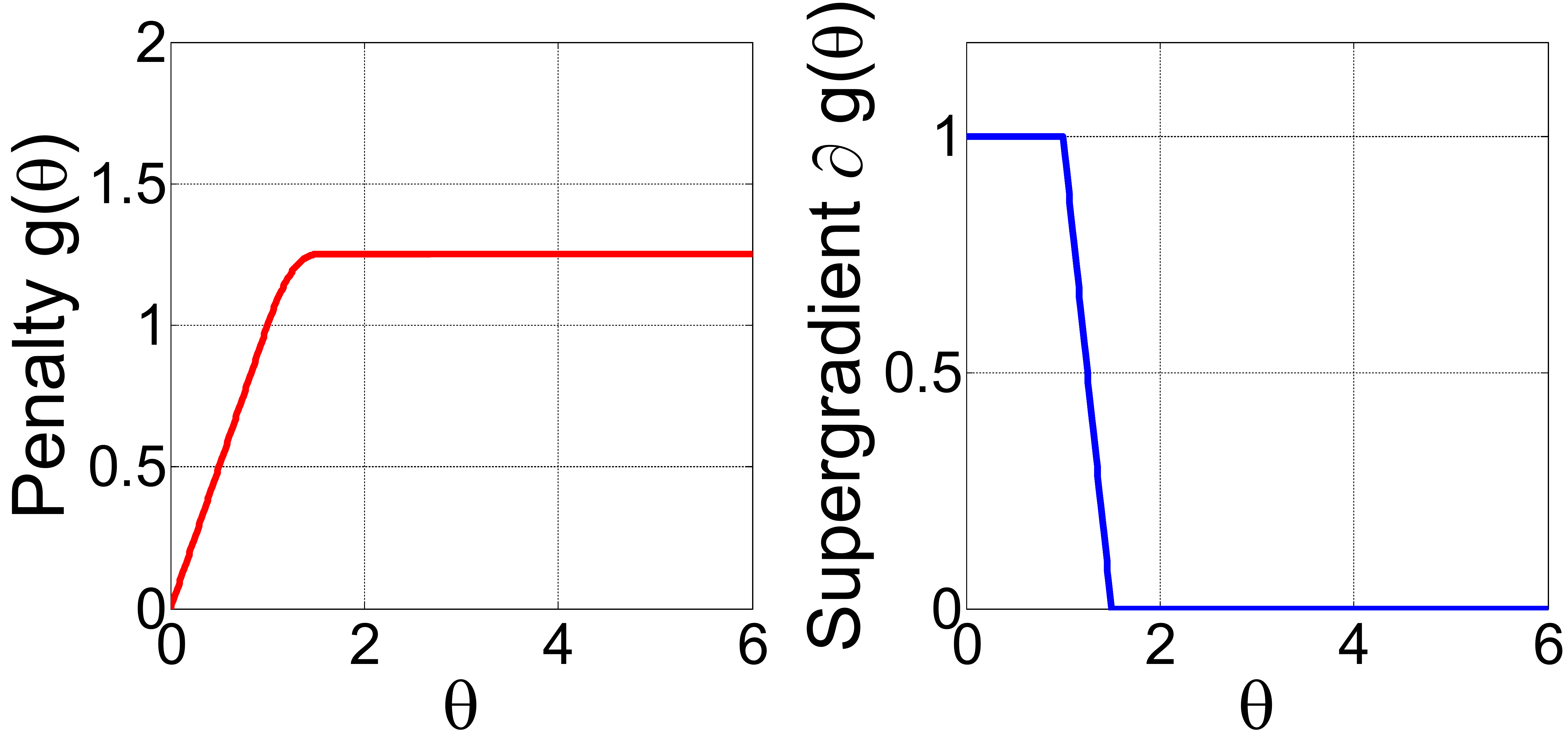}
		\caption{\scriptsize{SCAD Penalty \cite{fan2001variable}}}
	\end{subfigure}
	\begin{subfigure}[b]{0.235\textwidth}
		\centering
		\includegraphics[width=\textwidth]{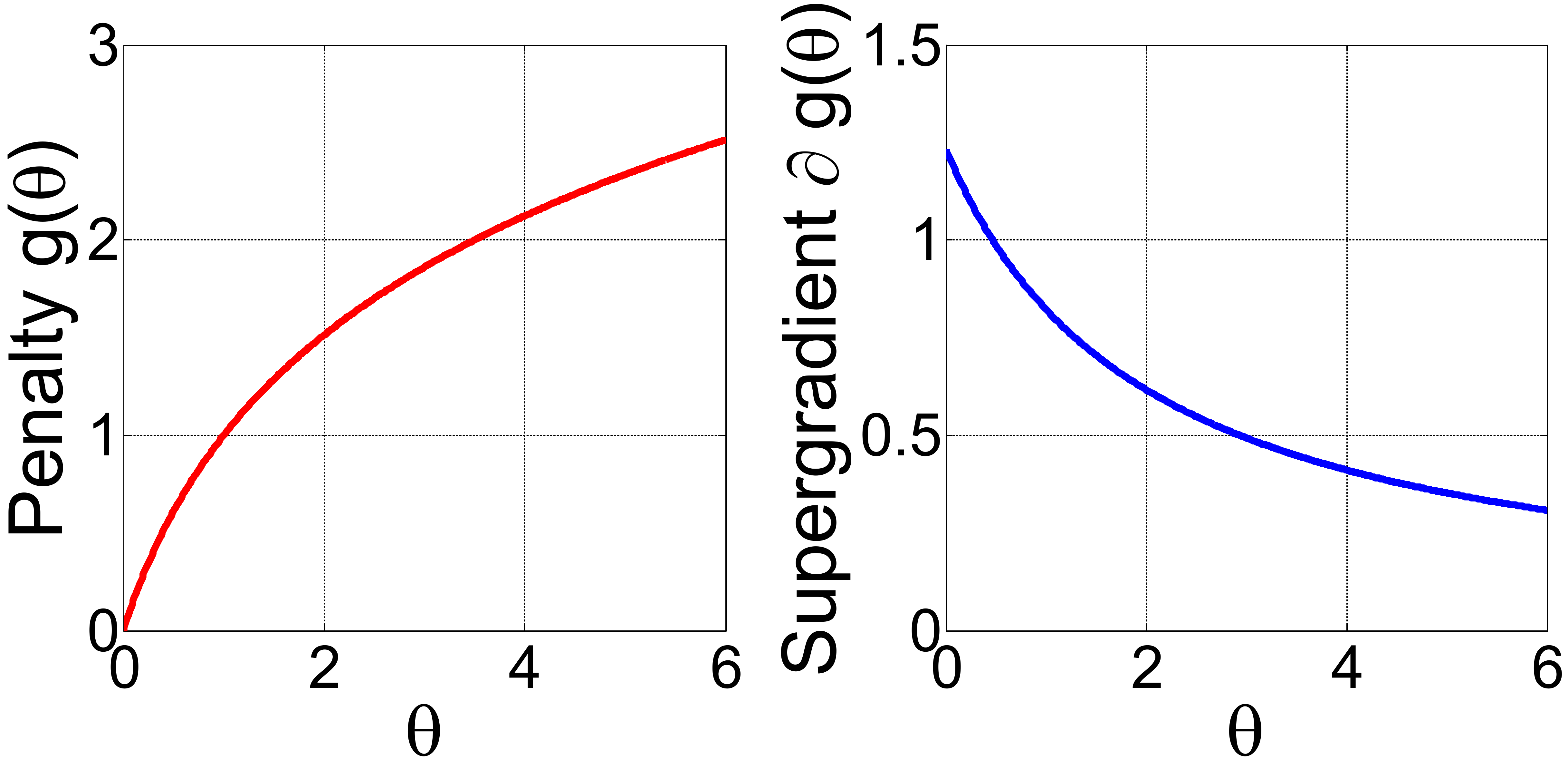}
		\caption{\scriptsize{Logarithm Penalty \cite{friedman2012fast}}}
	\end{subfigure}
	\begin{subfigure}[b]{0.235\textwidth}
		\centering
		\includegraphics[width=\textwidth]{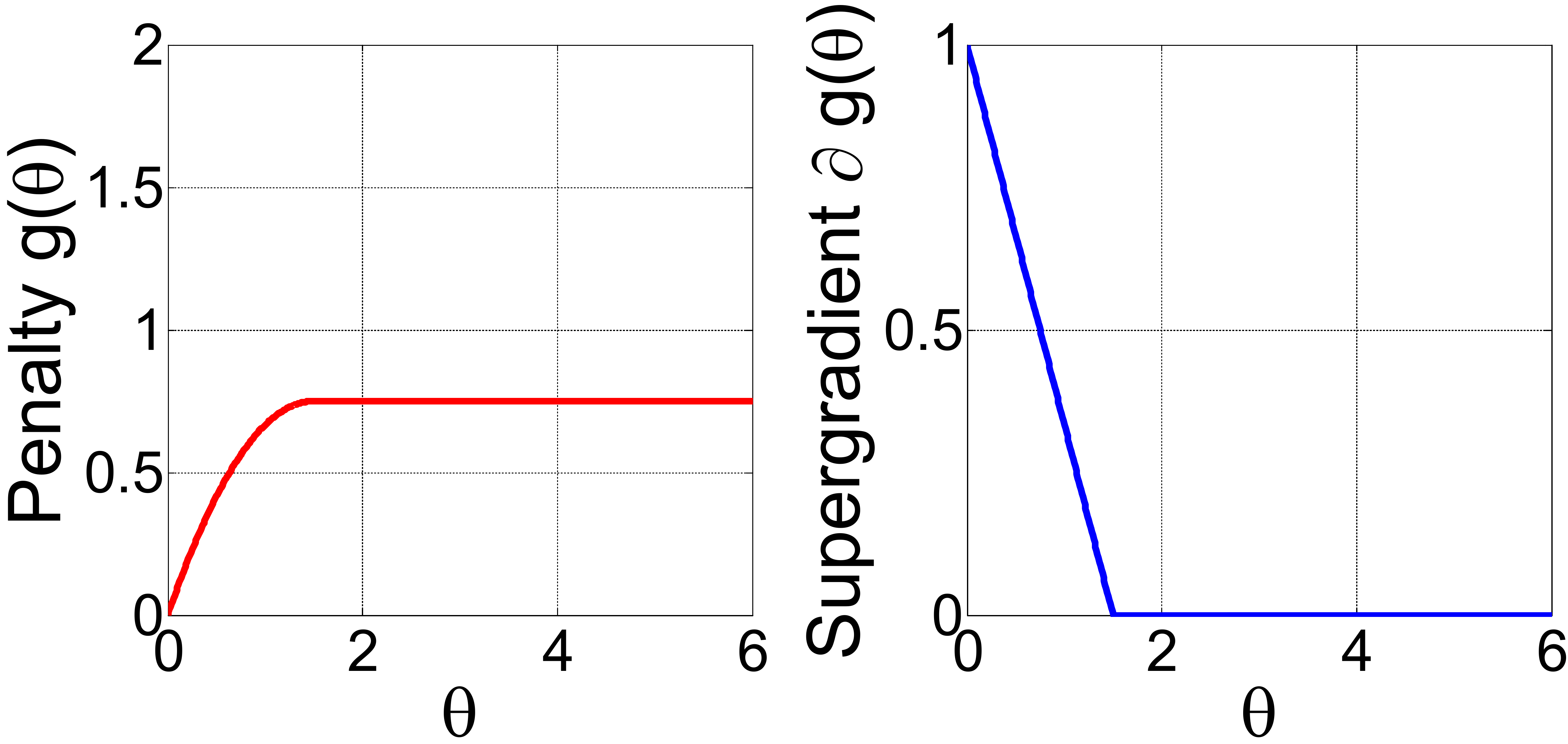}
		\caption{\scriptsize{MCP Penalty \cite{zhang2010nearly}}}
	\end{subfigure}
	\begin{subfigure}[b]{0.235\textwidth}
		\centering
		\includegraphics[width=\textwidth]{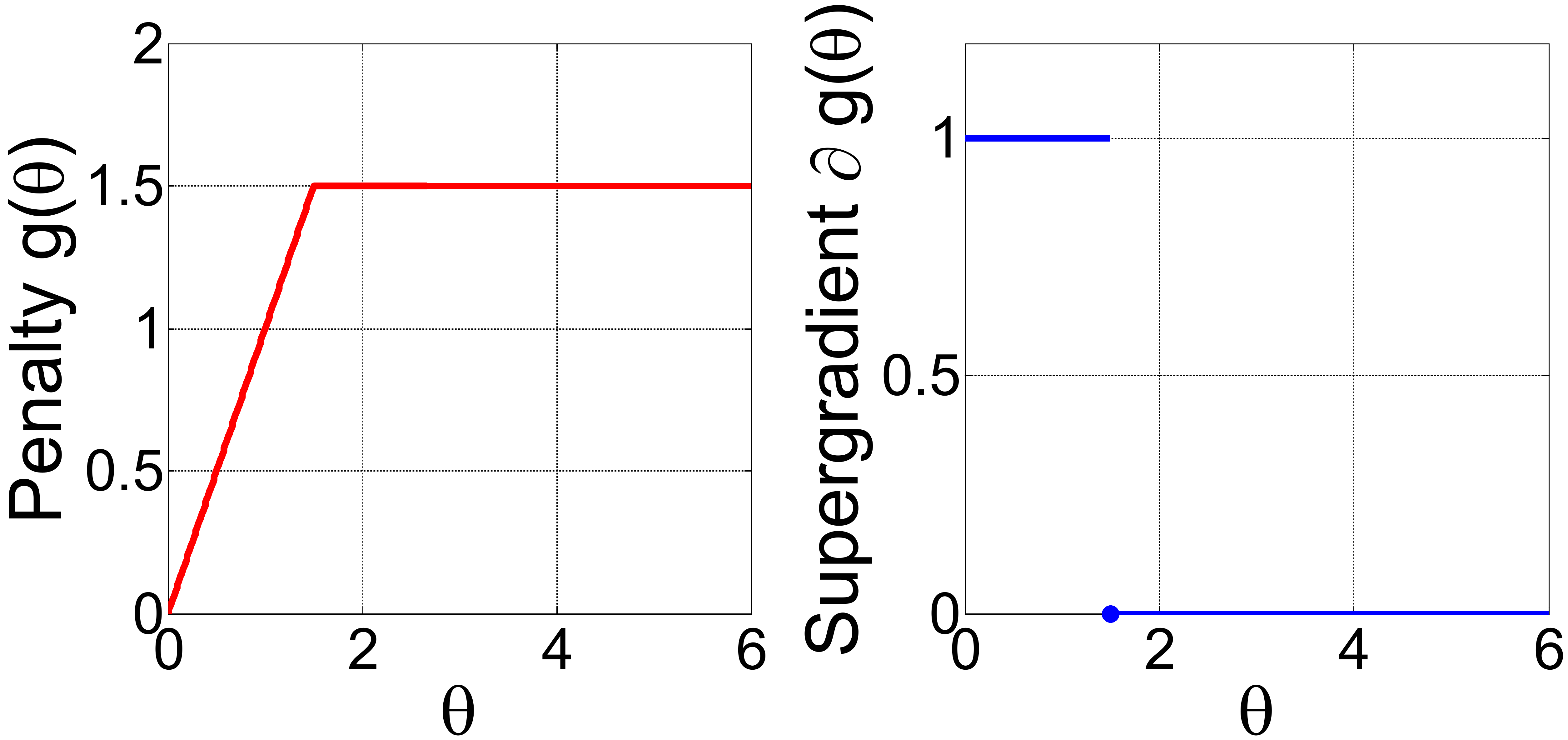}
		\caption{\scriptsize{Capped $L_1$ Penalty \cite{zhang2010analysis}}}
	\end{subfigure}
	\begin{subfigure}[b]{0.235\textwidth}
		\centering
		\includegraphics[width=\textwidth]{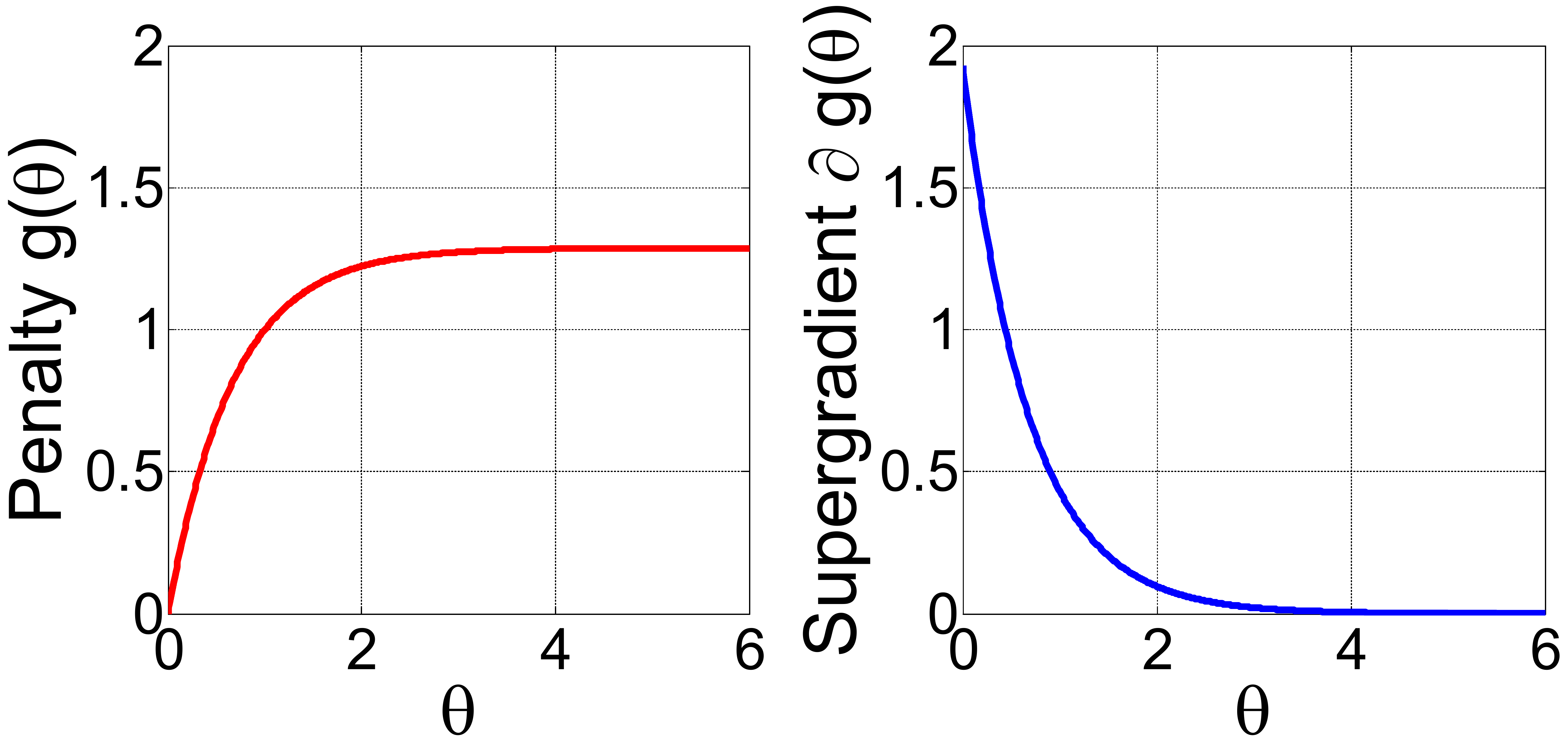}
		\caption{\scriptsize{ETP Penalty \cite{gao2011feasible}}}
	\end{subfigure}
	\begin{subfigure}[b]{0.235\textwidth}
		\centering
		\includegraphics[width=\textwidth]{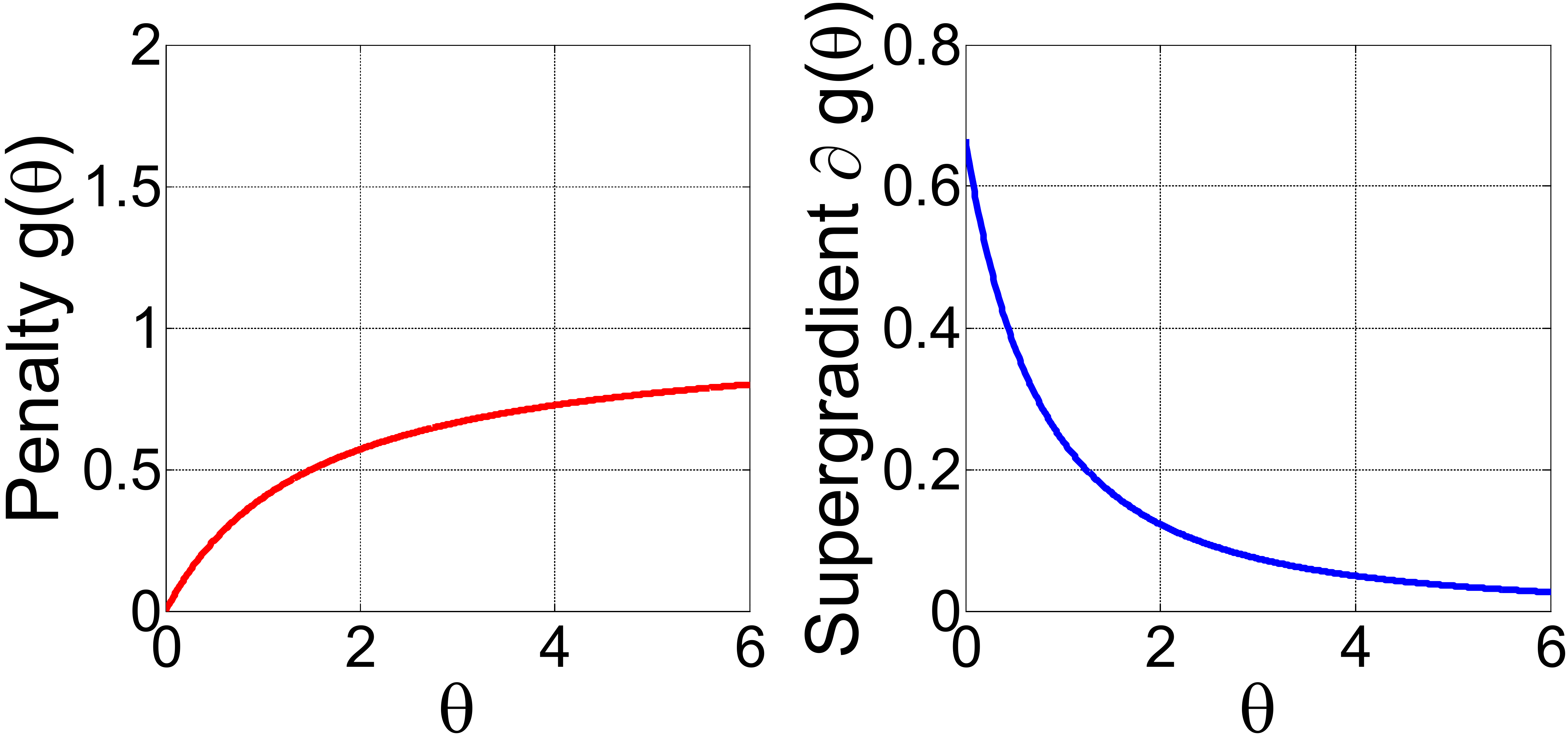}
		\caption{\scriptsize{Geman Penalty \cite{geman1995nonlinear}}}
	\end{subfigure}\ \ \ \ 
	\begin{subfigure}[b]{0.235\textwidth}
		\centering
		\includegraphics[width=\textwidth]{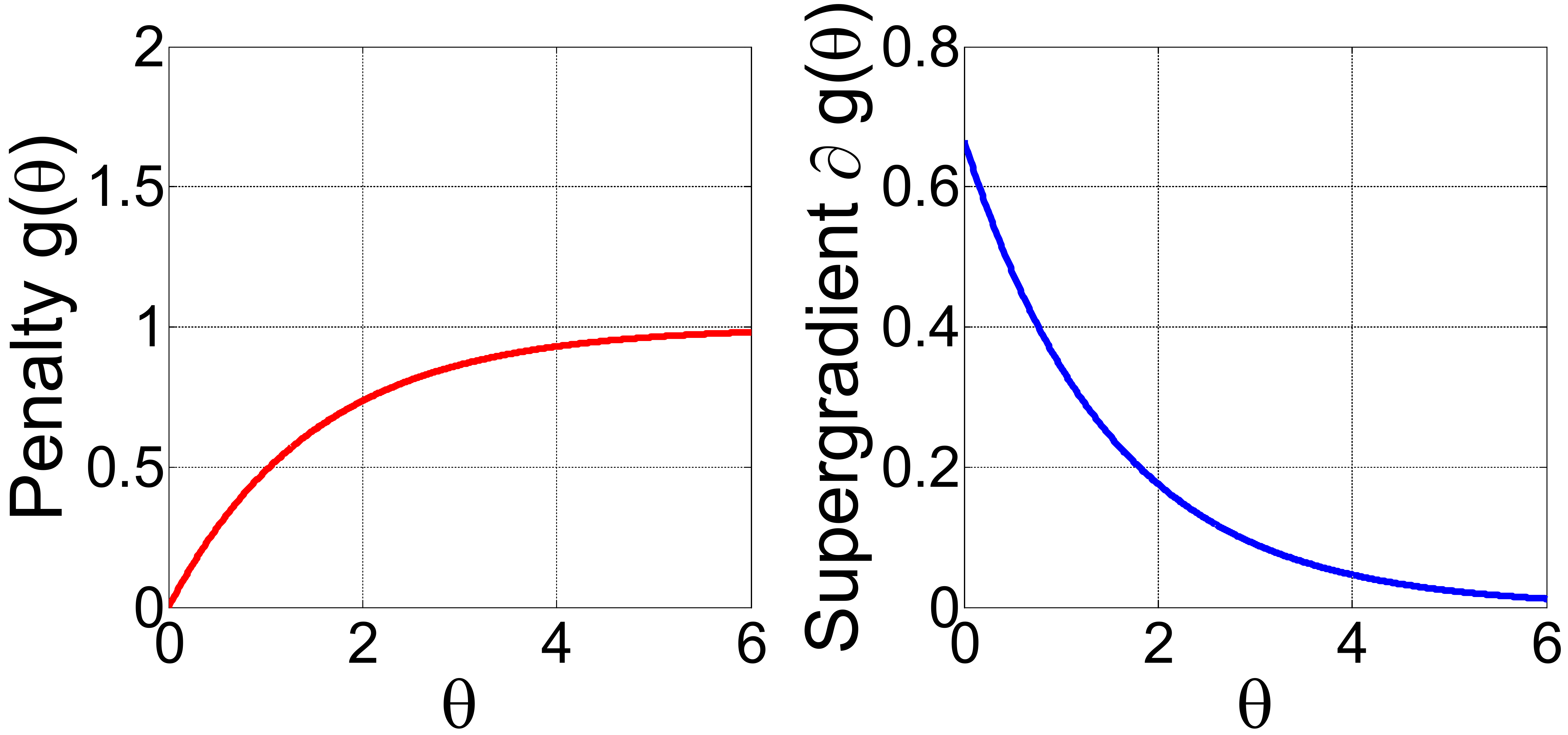}
		\caption{\scriptsize{Laplace Penalty \cite{trzasko2009highly}}}
	\end{subfigure}
	\caption{{Illustration of the popular nonconvex surrogate functions of $||\theta||_0$ (left) and their supergradients (right). For the $L_p$ penalty, $p=0.5$. For all these penalties, $\lambda=1$ and $\gamma=1.5$. }}\label{fig_nonconfun}
	\vspace{-1em}
\end{figure*}

Conventional CS recovery uses the $L_1$-norm, i.e., $\|\x\|_1=\sum_i|x_i|$, as the surrogate of the $L_0$-norm, i.e., $\|\x\|_0=\#\{x_i\neq0\}$, and the resulting convex problem can be solved by fast first-order solvers \cite{beck2009fast,donoho2008fast}.  
Though for certain problems, the $L_1$-minimization is equivalent to the $L_0$-minimization under certain incoherence conditions \cite{donoho2006most}, the obtained solution by $L_1$-minimization is usually suboptimal to the original $L_0$-minimization since the $L_1$-norm is a loose approximation of the $L_0$-norm. This motivates to approximate the $L_0$-norm by nonconvex continuous surrogate functions. 
Many known nonconvex surrogates of $L_0$-norm have been proposed, including $L_p$-norm ($0<p<1$) \cite{frank1993statistical}, Smoothly Clipped Absolute Deviation (SCAD)~\cite{fan2001variable}, Logarithm \cite{friedman2012fast}, Minimax Concave Penalty (MCP) \cite{zhang2010nearly}, Capped $L_1$ \cite{zhang2010analysis}, Exponential-Type Penalty (ETP) \cite{gao2011feasible}, Geman \cite{geman1995nonlinear} and Laplace \cite{trzasko2009highly}. We summarize their definitions in Table \ref{tab_nonpenlty} and visualize them in Figure \ref{fig_nonconfun}. Numerical studies \cite{candes2008enhancing,lai2013improved} have shown that the nonconvex sparse optimization usually outperforms convex models in the areas of signal recovery, error correction and image processing. 


\begin{table}[!t]
\tiny
\centering
\caption{{Popular nonconvex surrogate functions of $||\theta||_0$ and their supergradients.}}
\label{tab_nonpenlty}
\centering
\begin{tabular}{r| l| l }
\hline
Penalty & Formula $g(\theta)$, $\theta\geq0$, $\lambda>0$ & Supergradient $\partial g(\theta)$\\ \hline
$L_p$ \cite{frank1993statistical} & $\lambda\theta^p$
&  $\begin{cases}
		+\infty, &	\text{ if } \theta=0,\\
		\lambda p\theta^{p-1}, & \text{ if } \theta>0.
		\end{cases}$\\\hline
SCAD \cite{fan2001variable} & $\begin{cases}
			 \lambda\theta, & \text{ if } \theta\leq\lambda, \\
			 \frac{-\theta^2+2\gamma\lambda\theta-\lambda^2}{2(\gamma-1)}, & \text{ if } \lambda<\theta\leq\gamma\lambda, \\
			\frac{\lambda^2(\gamma+1)}{2}, &\text{ if }\theta>\gamma\lambda.
			\end{cases}$
	&		$\begin{cases}
			 \lambda, & \text{ if } \theta\leq\lambda, \\
			 \frac{\gamma\lambda-\theta}{\gamma-1}, & \text{ if } \lambda<\theta\leq\gamma\lambda, \\
			0, &\text{ if }\theta>\gamma\lambda.
			\end{cases}$ \\\hline
Logarithm \cite{friedman2012fast}& $\frac{\lambda}{\log(\gamma+1)}\log(\gamma\theta+1)$ & $\frac{\gamma\lambda}{(\gamma\theta+1)\log(\gamma+1)}$ \\\hline
MCP \cite{zhang2010nearly}& $\begin{cases}
			 \lambda\theta-\frac{\theta^2}{2\gamma}, & \text{ if } \theta<\gamma\lambda, \\
			 \frac{1}{2}\gamma\lambda^2, & \text{ if } \theta\geq\gamma\lambda.
			\end{cases}$
	& $\begin{cases}
		\lambda-\frac{\theta}{\gamma}, &	\text{ if } \theta<\gamma\lambda,\\
		0, & \text{ if } \theta\geq\gamma\lambda.
		\end{cases}$\\\hline	
Capped $L_1$ \cite{zhang2010analysis}& 	
$\begin{cases}
	\lambda\theta, &	\text{ if } \theta<\gamma,\\	
	\lambda\gamma, & \text{ if } \theta\geq\gamma.
\end{cases}$	
& 	
$\begin{cases}
	\lambda, &	\text{ if } \theta<\gamma,\\
	[0,\lambda], & \text{ if } \theta=\gamma,\\
	0, & \text{ if } \theta>\gamma.
\end{cases}$\\\hline
ETP \cite{gao2011feasible}& $\frac{\lambda}{1-\exp(-\gamma)}(1-\exp(-\gamma\theta))$ & $\frac{\lambda\gamma}{1-\exp(-\gamma)}\exp(-\gamma\theta)$\\\hline
Geman \cite{geman1995nonlinear} & $\frac{\lambda\theta}{\theta+\gamma}$ &$\frac{\lambda\gamma}{(\theta+\gamma)^2}$\\\hline
Laplace \cite{trzasko2009highly}& $\lambda(1-\exp(-\frac{\theta}{\gamma}))$ & $\frac{\lambda}{\gamma}\exp(-\frac{\theta}{\gamma})$ \\\hline
\end{tabular}
\vspace{-1em}
\end{table}

\begin{figure*}
	\begin{subfigure}[b]{0.19\textwidth}
		\centering
		\includegraphics[width=\textwidth]{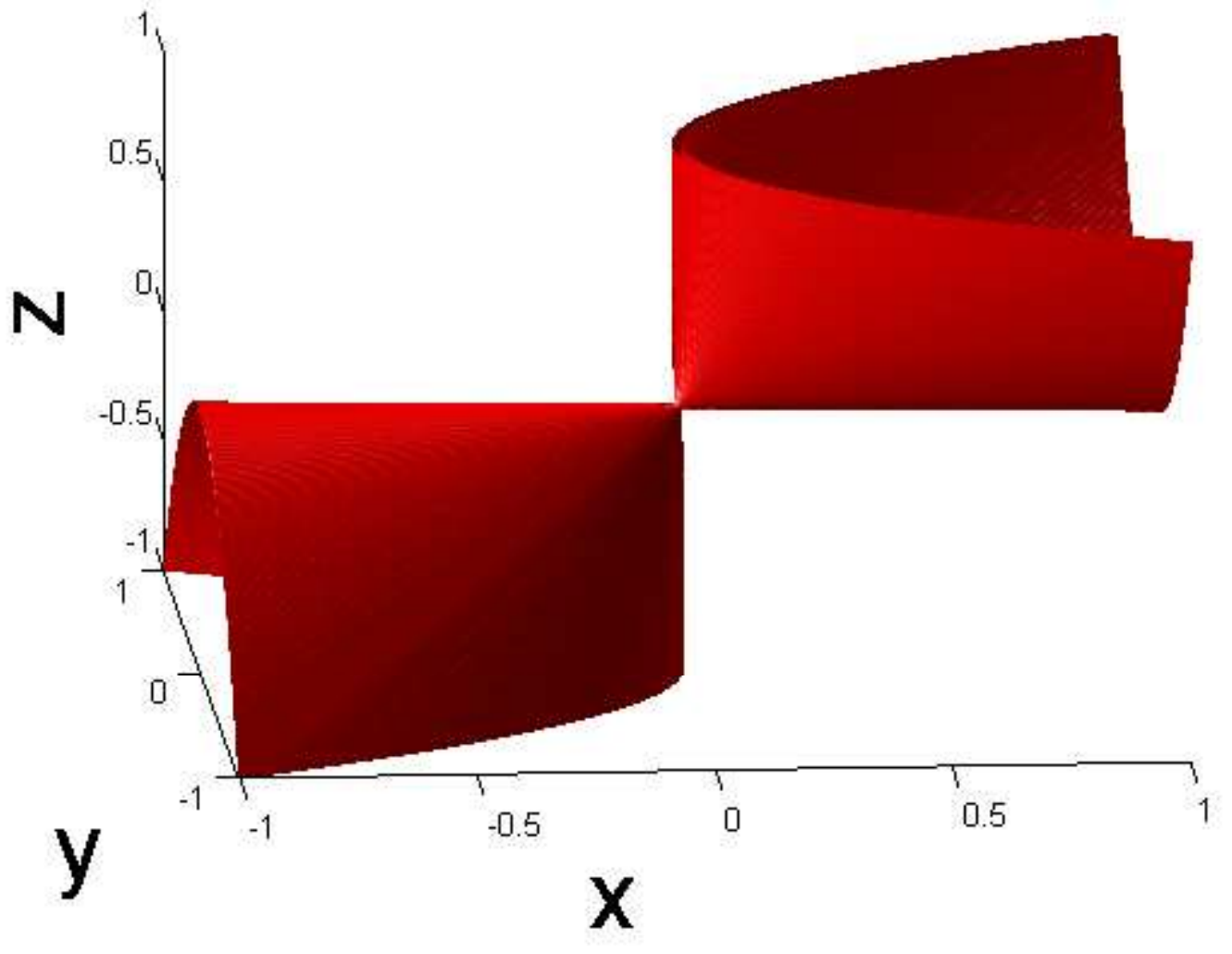}
		\caption{\scriptsize{ Rank }}
	\end{subfigure}
	\begin{subfigure}[b]{0.19\textwidth}
		\centering
		\includegraphics[width=\textwidth]{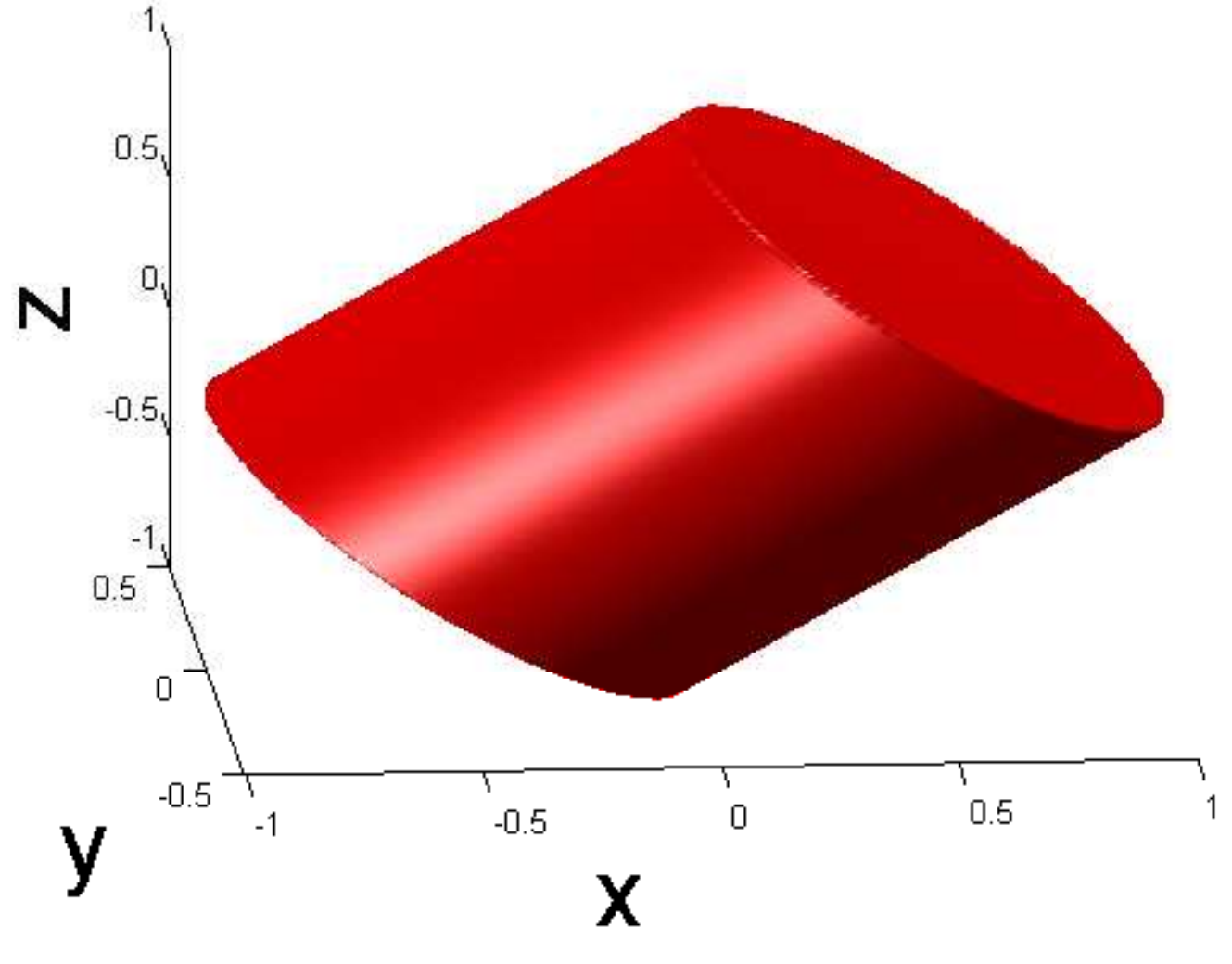}
		\caption{\scriptsize{Nuclear norm }}
	\end{subfigure}
	\begin{subfigure}[b]{0.19\textwidth}
		\centering
		\includegraphics[width=\textwidth]{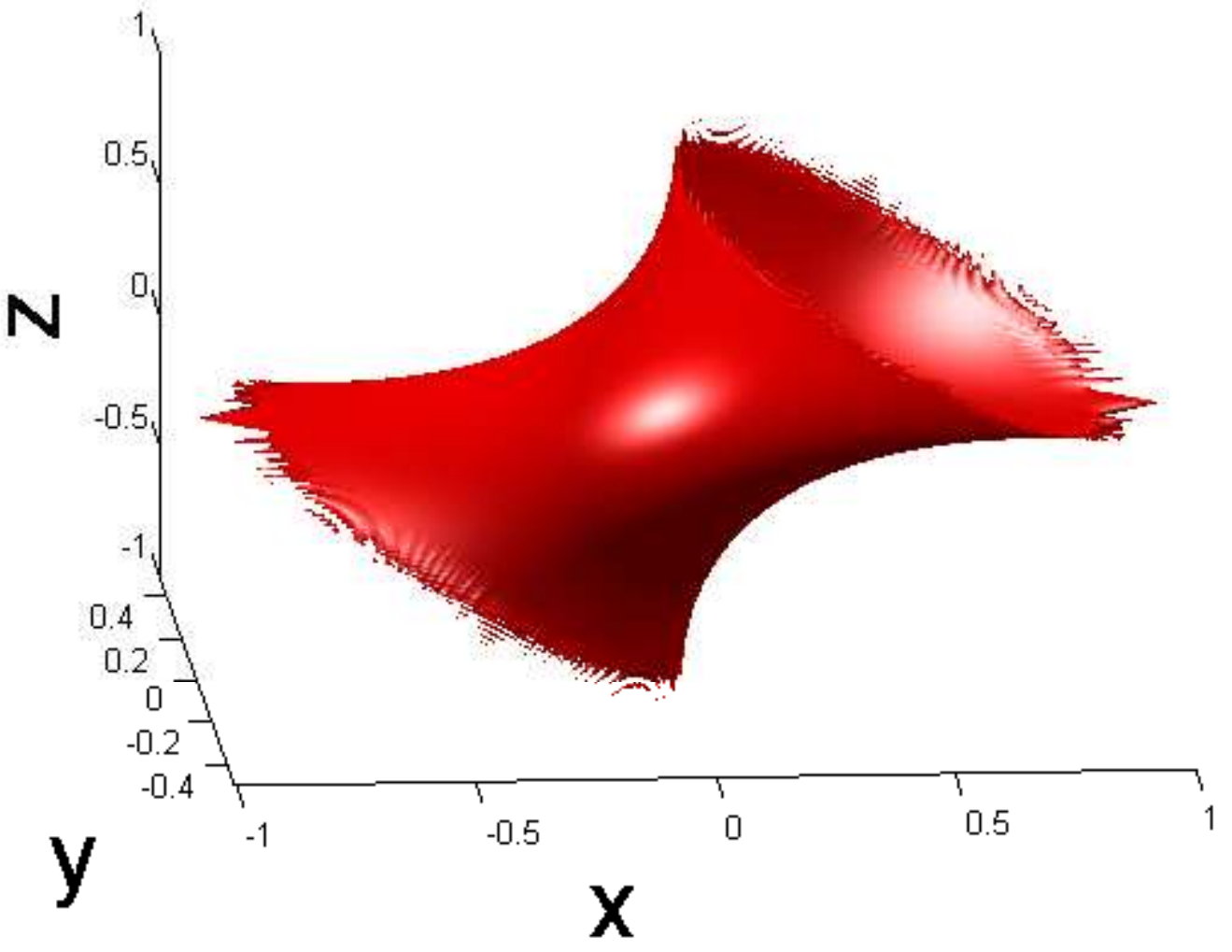}
		\caption{\scriptsize{ $L_p$  }}
	\end{subfigure}
	\begin{subfigure}[b]{0.19\textwidth}
		\centering
		\includegraphics[width=\textwidth]{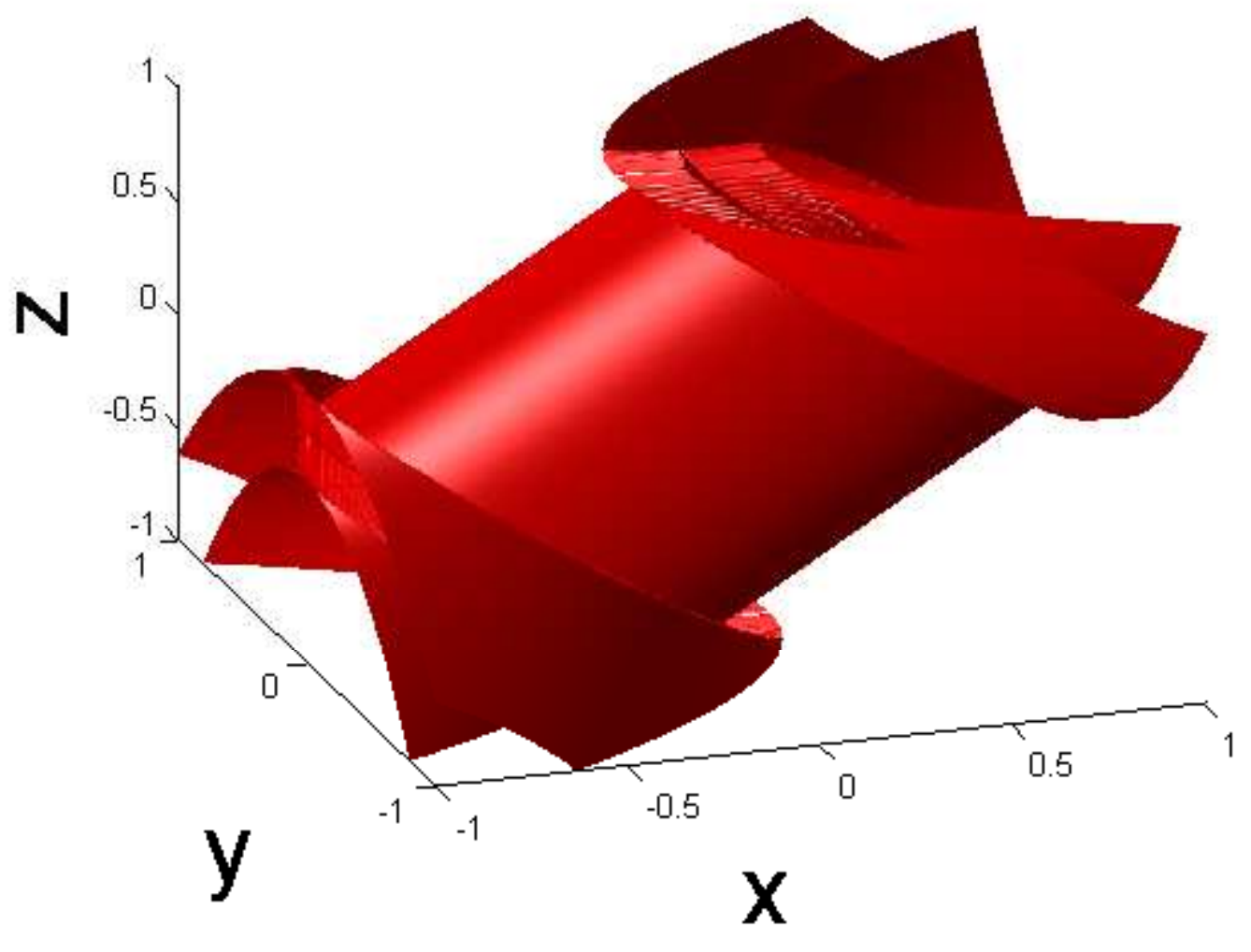}
		\caption{\scriptsize{SCAD  }}
	\end{subfigure}
	\begin{subfigure}[b]{0.19\textwidth}
		\centering
		\includegraphics[width=\textwidth]{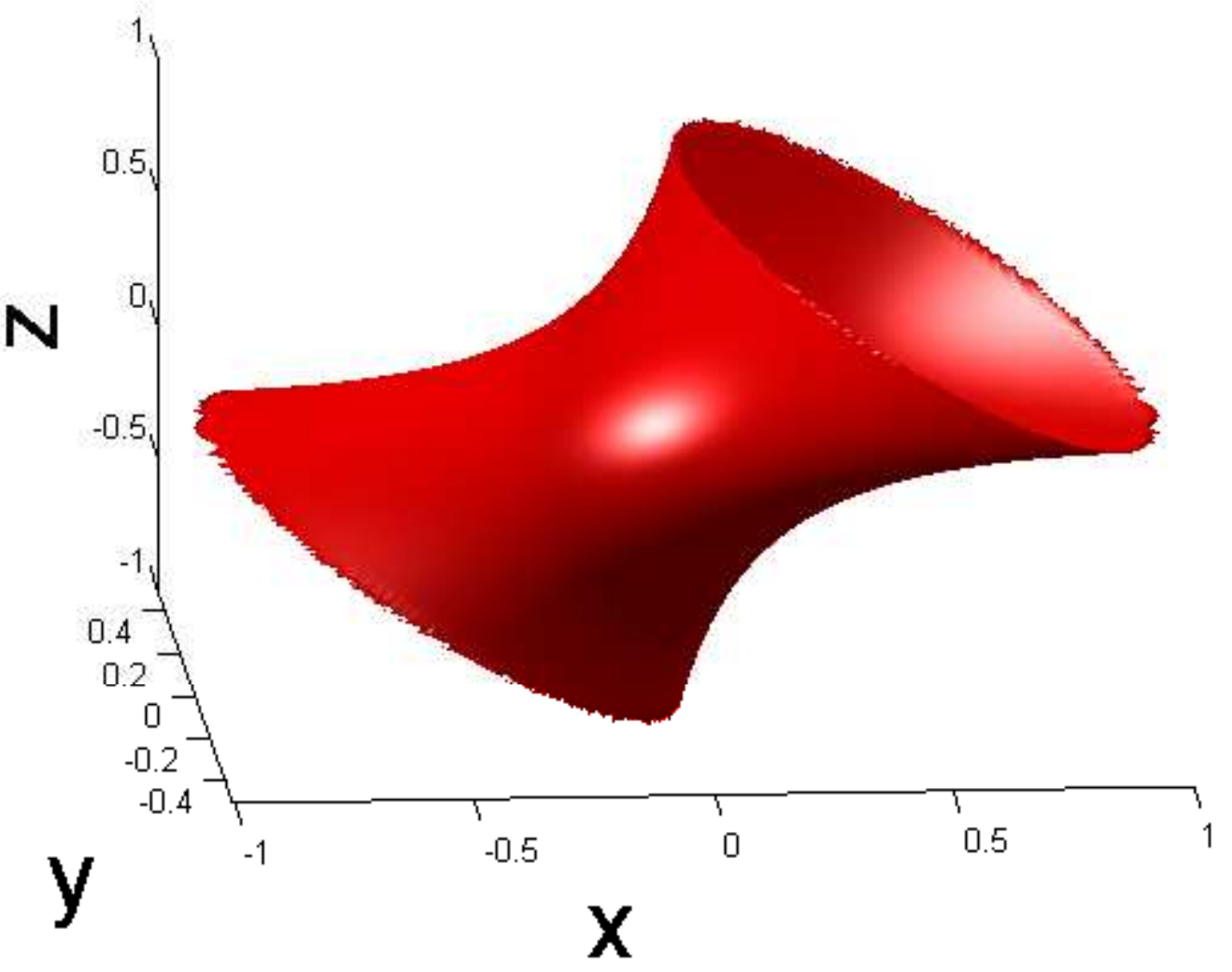}
		\caption{\scriptsize{Logarithm  }}
	\end{subfigure}
	\begin{subfigure}[b]{0.2\textwidth}
		\centering
		\includegraphics[width=\textwidth]{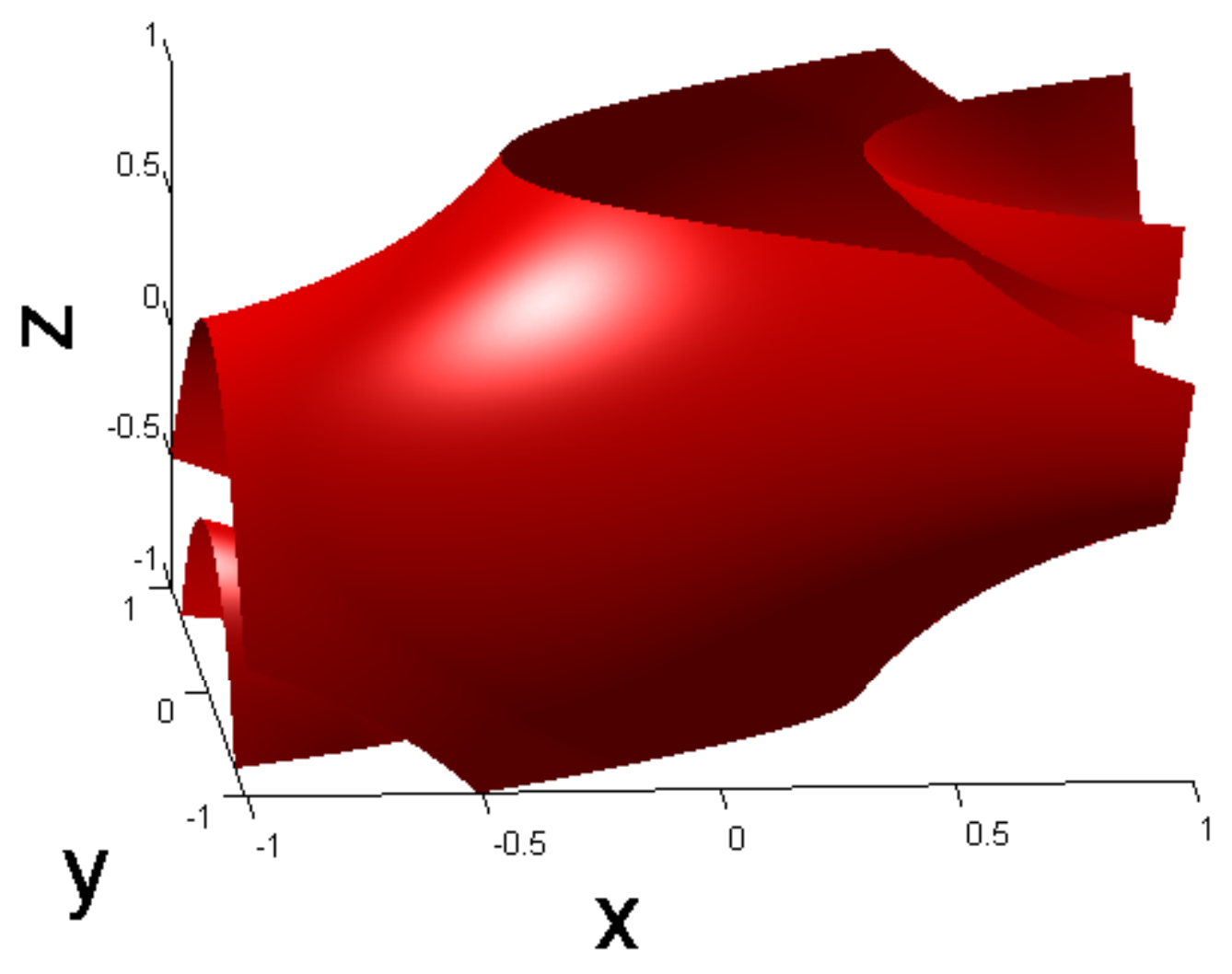}
		\caption{\scriptsize{MCP }}
	\end{subfigure}
	\begin{subfigure}[b]{0.19\textwidth}
		\centering
		\includegraphics[width=\textwidth]{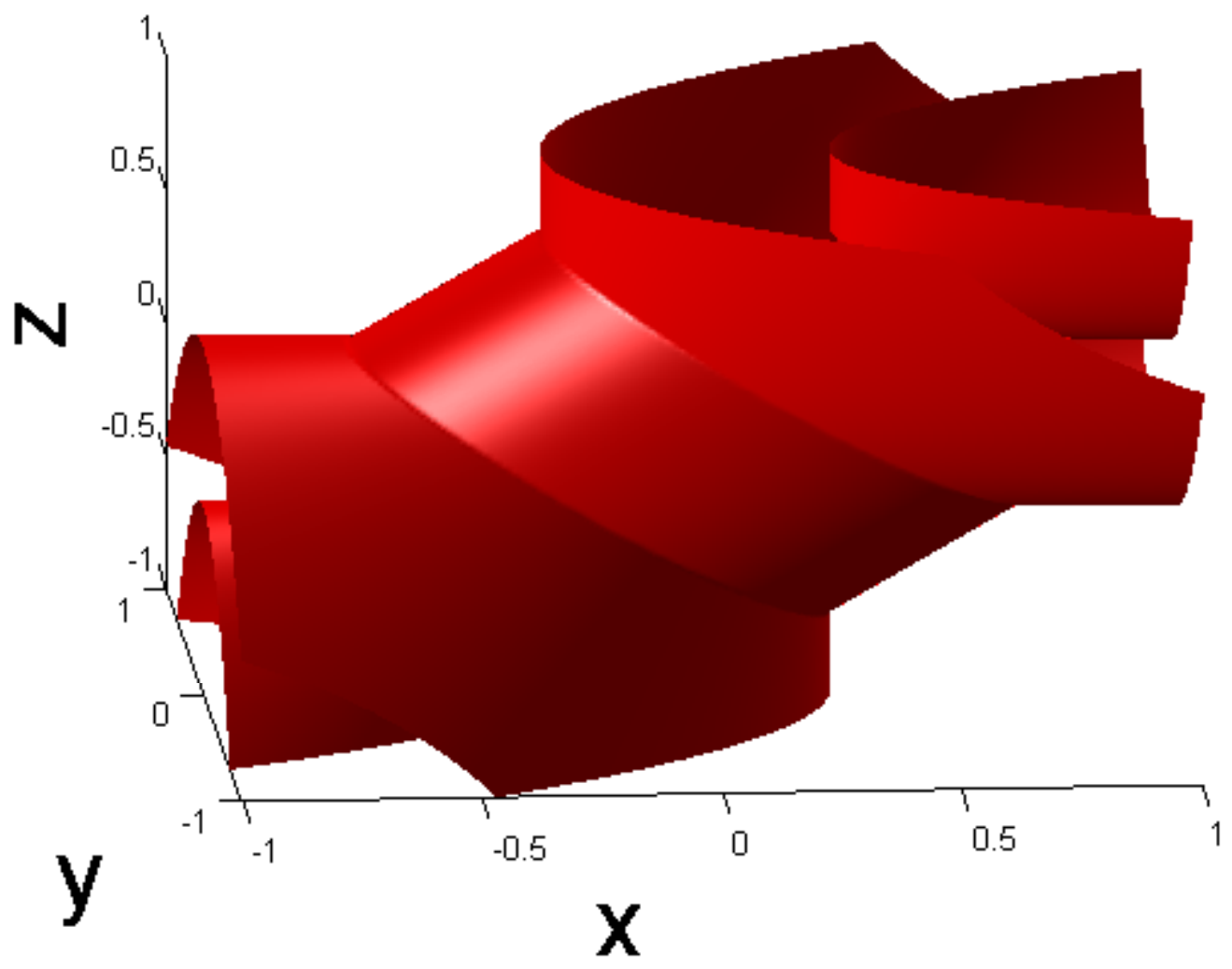}
		\caption{\scriptsize{Capped $L_1$ }}
	\end{subfigure}
	\begin{subfigure}[b]{0.2\textwidth}
		\centering
		\includegraphics[width=\textwidth]{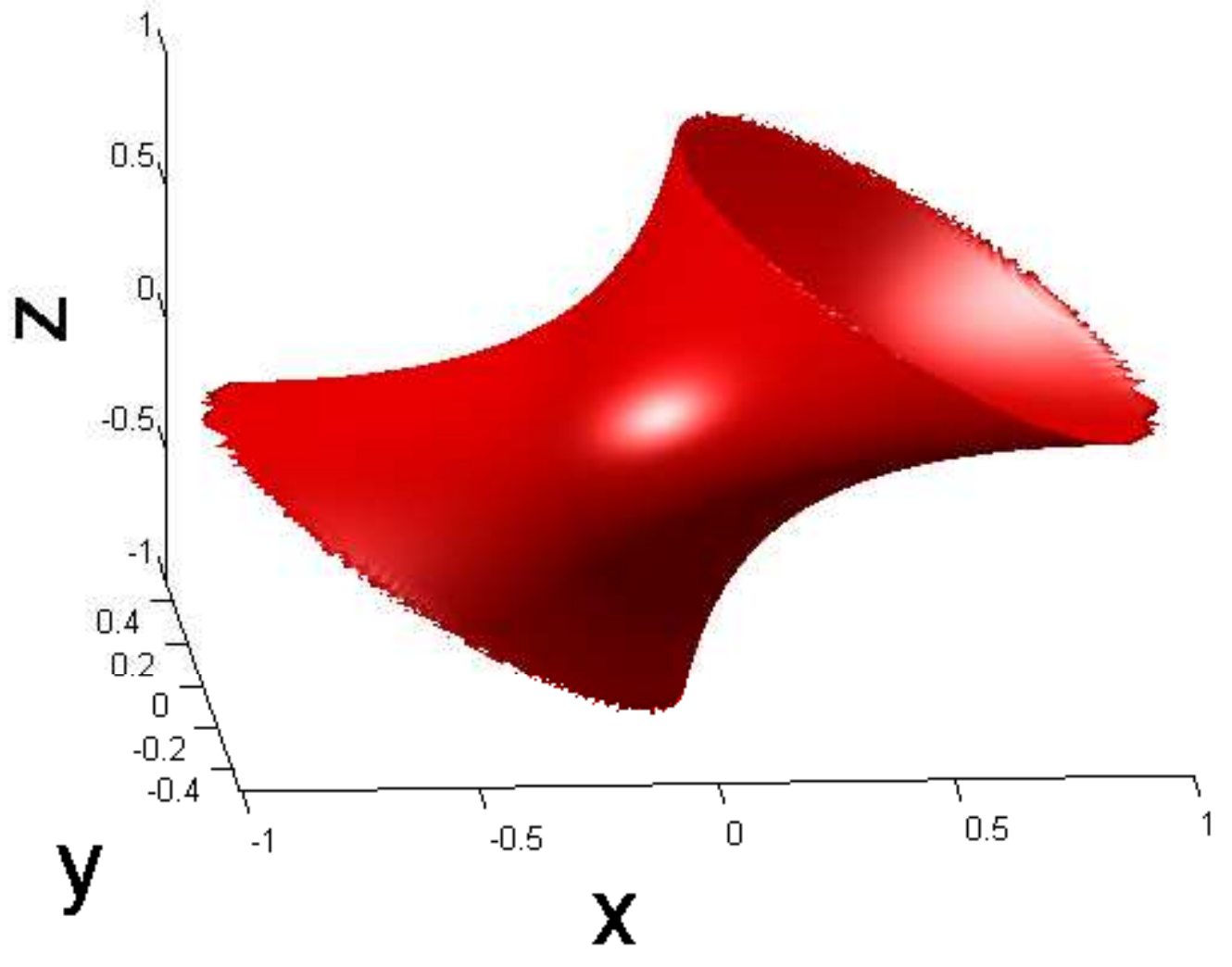}
		\caption{\scriptsize{ETP }}
	\end{subfigure}
	\begin{subfigure}[b]{0.19\textwidth}
		\centering
		\includegraphics[width=\textwidth]{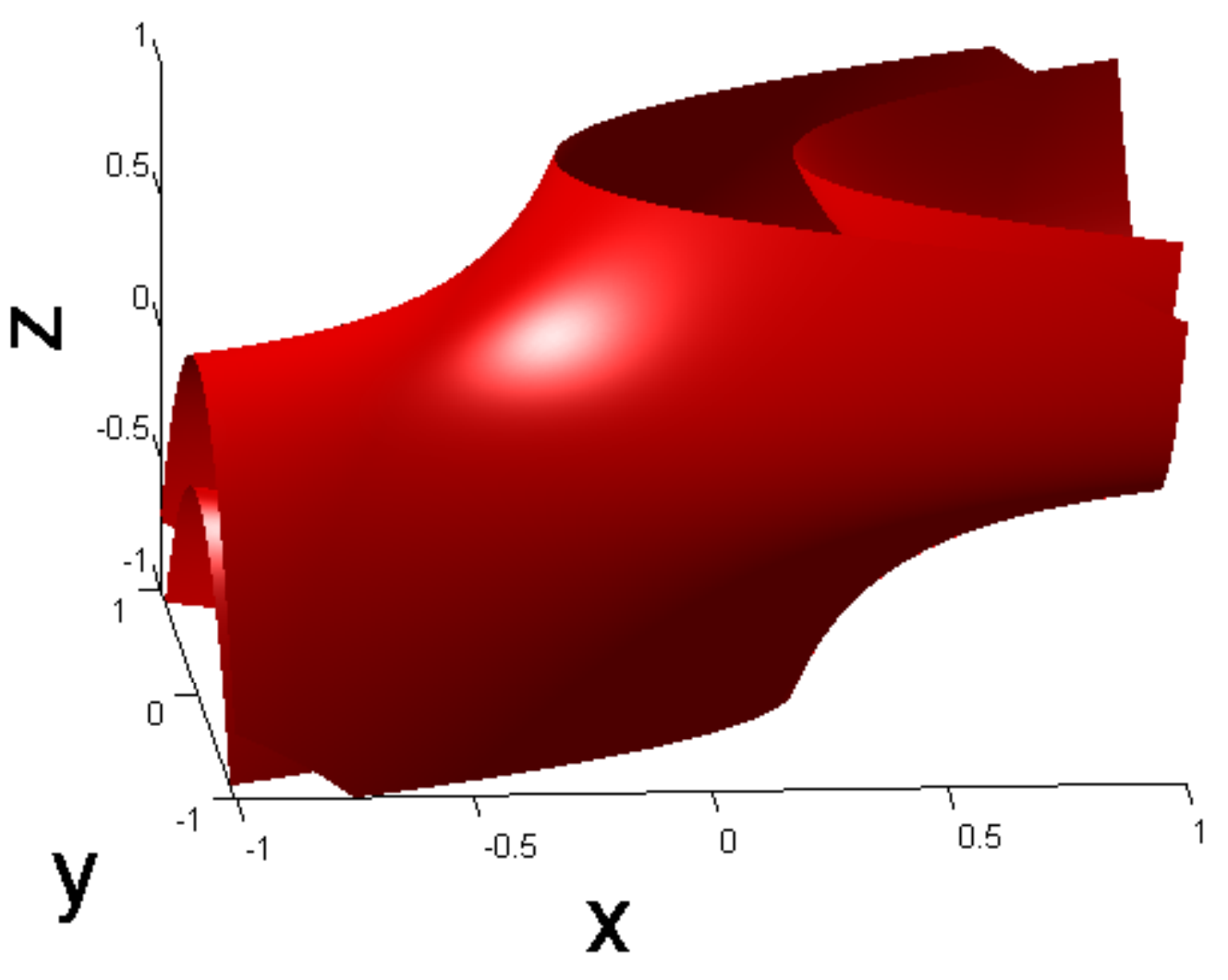}
		\caption{\scriptsize{Geman  }}
	\end{subfigure}
	\begin{subfigure}[b]{0.19\textwidth}
		\centering
		\includegraphics[width=\textwidth]{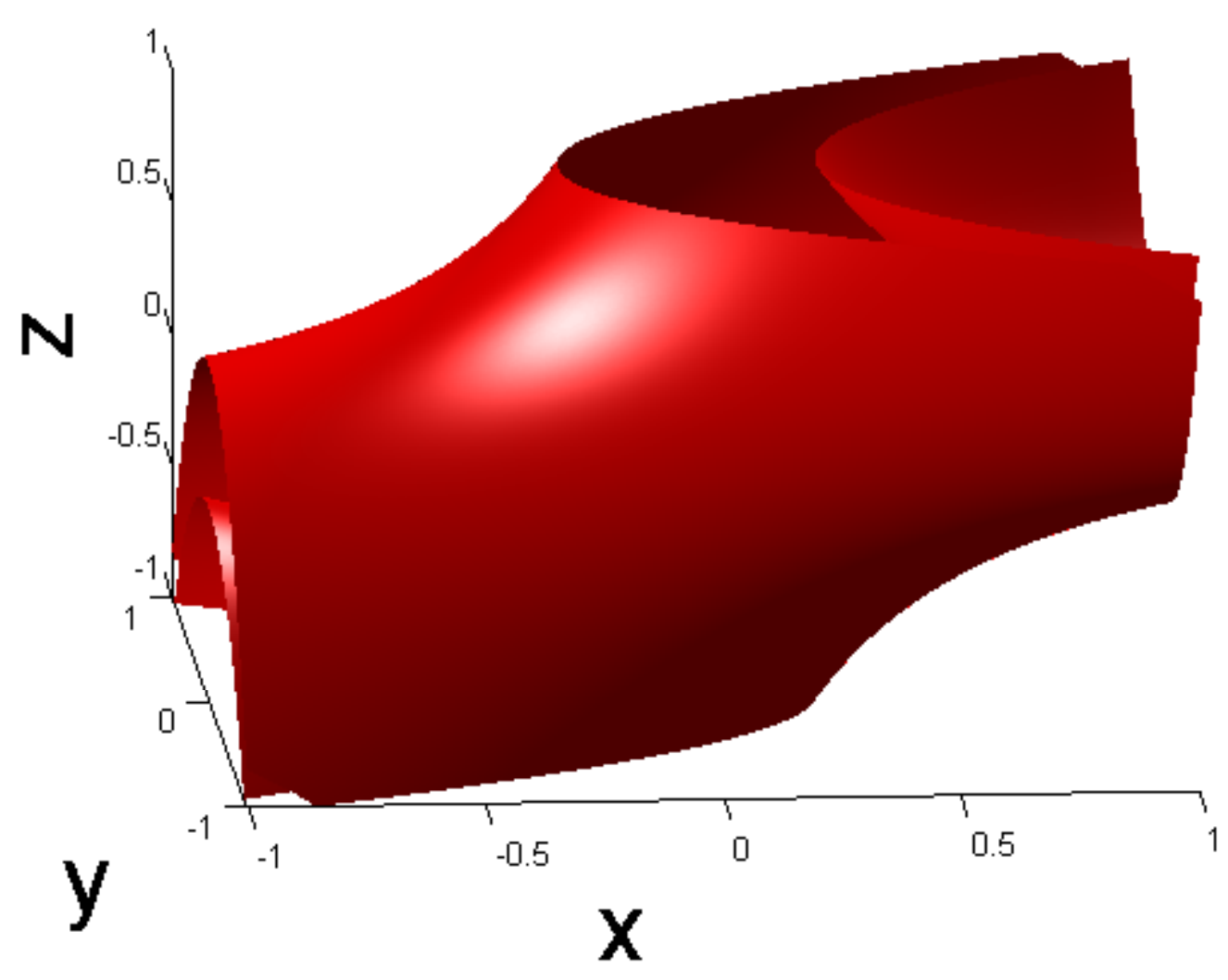}
		\caption{\scriptsize{Laplace  }}
	\end{subfigure}
	\caption{{Manifold of constant penalty for a symmetric $2\times 2$ matrix $\X$ for the (a) rank penalty, (b) nuclear norm, (c-j) $\sum_ig(\sigma_i(\X))$, where the choices of the nonconvex $g$ are listed in Table \ref{tab_nonpenlty}. For $\lambda$ in $g$, we set $\lambda=1$. For other parameters, we set (c) $p=0.5$, (d) $\gamma=0.6$, (e) $\gamma = 5$, (f) $\gamma=1.5$, (g)  $\gamma=0.7$, (h)  $\gamma=2$, (i)  $\gamma=0.5$ and (j)  $\gamma=0.8$. Note that the manifold will be different for $g$ with different parameters.  } }  \label{fig_nonconvexlowrank_matrix}
	\vspace{-1em}
\end{figure*}

The low rank structure of a matrix is the sparsity defined on its singular values. A particularly interesting model is the low rank matrix recovery problem
\begin{equation}\label{pro_rankmi}
\min_{\X} \lambda\text{rank}(\X)+\frac{1}{2}||\mathcal{A}(\X)-\mathbf{b}||_F^2,
\end{equation}
where $\mathcal{A}$ is a linear mapping and $\lambda>0$. The above low rank minimization problem arises in many computer vision tasks such as
multiple category classification \cite{amit2007uncovering},
matrix completion \cite{toh2010accelerated}, multi-task learning \cite{argyriou2008convex} and low-rank representation with squared loss for subspace segmentation \cite{robustlrr}. Similar to the $L_0$-minimization, the rank minimization problem (\ref{pro_rankmi}) is also challenging to solve. Thus, the rank function is usually replaced by the convex nuclear norm, $\|\X\|_*=\sum_i{\sigma}_i(\X)$, where $\sigma_i(\X)$'s denote the singular values of $\X$. This leads to 
a relaxed convex formulation of (\ref{pro_rankmi}):
\begin{equation}\label{pro_nuclear}
\min_{\X} \lambda\|\X\|_*+\frac{1}{2}||\mathcal{A}(\X)-\mathbf{b}||_F^2.
\end{equation}
The above convex problem can be efficiently solved by many known solvers \cite{APG,boyd2011distributed}. 
However, the obtained solution by solving (\ref{pro_nuclear}) is usually suboptimal to (\ref{pro_rankmi}) since the nuclear norm is also a loose approximation of the rank function. Such a phenomenon is similar to the difference between $L_1$-norm and $L_0$-norm for sparse vector recovery. However, different from the nonconvex surrogates of $L_0$-norm, the nonconvex rank surrogates and the optimization solvers have not been well studied before.

In this paper, to achieve a better approximation of the rank function, we extend the nonconvex surrogates of $L_0$-norm shown in Table \ref{tab_nonpenlty} onto the singular values of the matrix, and show how to solve the following general nonconvex nonsmooth low rank minimization problem \cite{lu2014generalized} 
\begin{equation}\label{eq_genpro}
\min_{\mathbf{X}\in\mathbb{R}^{m\times n}} F(\X)=\sum_{i=1}^mg(\sigma_i(\mathbf{X}))+f(\mathbf{X}),
\end{equation}
where $\sigma_i(\X)$ denotes the $i$-th singular value of $\X\in\mathbb{R}^{m\times n}$ (we assume that $m\leq n$ in this work). 
The penalty function $g$ and loss function $f$ satisfy the following assumptions:
\begin{itemize}
\item[\textbf{A1}] $g:$ $\mathbb{R}^+\rightarrow\mathbb{R}^+$ is continuous, concave and monotonically increasing on $[0,\infty)$. It is possibly nonsmooth.
\item[\textbf{A2}] $f$: $\mathbb{R}^{m\times n}\rightarrow\mathbb{R}^+$ is a smooth function of type $C^{1,1}$, i.e., the gradient is Lipschitz continuous,
\begin{equation}\label{deflip1}
||\nabla f(\mathbf{X})-\nabla f(\mathbf{Y})||_F\leq L(f)||\mathbf{X}-\mathbf{Y}||_F,
\end{equation}
for any $\mathbf{X}, \mathbf{Y}\in\mathbb{R}^{m\times n}$, $L(f)>0$ is called Lipschitz constant of $\nabla f$. $f(\X)$ is possibly nonconvex.
\end{itemize}

Note that problem (\ref{eq_genpro}) is very general. All the nonconvex surrogates $g$ of $L_0$-norm  in Table \ref{tab_nonpenlty} satisfy the assumption \textbf{A1}. So $\sum_{i=1}^mg(\sigma_i(\mathbf{X}))$ is the nonconvex surrogate of the rank function\footnote{Note that the singular values of a matrix are always nonegative. So we only consider the nonconvex $g$ definted on $\mathbb{R}^+$.}. It is expected that it approximates the rank function better than the convex nuclear norm. To see this more intuitively, we show the balls of constant penalties for a symmetric $2\times 2$ matrix in Figure \ref{fig_nonconvexlowrank_matrix}. For the loss function $f$ in assumption \textbf{A2}, the most widely used one is the squared loss $\frac{1}{2}\|\mathcal{A}(\X)-\mathbf{b}\|_F^2$.

There are some related work which consider the nonconvex rank surrogates. But they are different from this work. The work \cite{IRLSrank,fornasier2011low} extend the $L_p$-norm of a vector to the Schatten-$p$ norm ($0<p<1$) and use the iteratively reweighted least squares (IRLS) algorithm to solve the nonconvex rank minimization problem with affine constraint. IRLS is also applied for the unconstrained problem with the smoothed Schatten-$p$ norm regularizer \cite{lai2011unconstrained}. However, the obtained solution by IRLS may not be
naturally of low rank, or it may require a lot of iterations to get a low rank solution. One may perform the singular value thresholding appropriately to achieve a low rank solution, but there has no theoretically sound rule to suggest a correct threshold. Another nonconvex rank surrogate is the truncated nuclear norm \cite{hu2012fast}. Their proposed alternating updating optimization algorithm may not be efficient due to double loops of iterations and cannot be applied to solve (\ref{eq_genpro}). The nonconvex low rank matrix completion problem considered in \cite{todeschini2013probabilistic} is a special case of our problem (\ref{eq_genpro}). Our solver shown later for (\ref{eq_genpro}) is also much more general. 
The work \cite{dong2014compressive} uses the nonconvex log-det heuristic in \cite{fazel2003log} for image recovery. But their augmented Lagrangian multiplier based solver lacks of the convergence guarantee. 
A possible method to solve (\ref{eq_genpro}) is the proximal gradient algorithm \cite{lu2015generalized}, which requires to compute the proximal mapping of the nonconvex function $g$. However, computing the proximal mapping requires solving a nonconvex problem exactly. To the best of our knowledge, without additional assumptions on $g$ (e.g., the convexity of $\nabla g$ \cite{lu2015generalized}), there does not exist a general solver for computing the proximal mapping of the general nonconvex $g$ in assumption \textbf{A1}.

In this work, we observe that all the existing nonconvex surrogates in Table \ref{tab_nonpenlty} are concave and monotonically increasing on $[0,\infty)$. Thus their gradients (or supergradients at the nonsmooth points) are nonnegative and monotonically decreasing. Based on this key fact, we propose an Iteratively Reweighted Nuclear Norm (IRNN) algorithm to solve (\ref{eq_genpro}). It computes the proximal operator of the weighted nuclear norm, which has a closed form solution due to the nonnegative and monotonically decreasing supergradients. The cost is the same as the computing of singular value thresholding which is widely used in convex nuclear norm minimization. In theory, we prove that IRNN monotonically decreases the objective function value and any limit point is a stationary point. 

Furthermore, note that problem (\ref{eq_genpro}) contains only one block of variable. But there are also some work which aim at finding several low rank matrices simultaneously, e.g., \cite{liu2011latent}. So we further extend IRNN to solve the following problem with $p\ \geq2$  blocks of variables
\begin{equation}\label{eq_genpro2}
\min_{\X} F(\X)=\sum_{j=1}^p\sum_{i=1}^{m_j}g_j(\sigma_i(\mathbf{X}_j))+f(\X),
\end{equation}
where $\X=\{{\X}_1,\cdots,{\X}_p\}$, ${\X}_j\in\mathbb{R}^{m_j\times n_j}$ (assume $m_j\leq n_j$), $g_j$'s satisfy the assumption \textbf{A1}, and $\nabla f$ is Lipschitz continuous defined as follows.
\begin{definition}\label{Lem_lips}
	Let $f: \mathbb{R}^{n_1}\times\cdots\times\mathbb{R}^{n_p}\rightarrow\mathbb{R}$ be differentiable. Then $\nabla f$ is called Lipschitz continuous if there exist $L_i(f)>0, i=1,\cdots,n$, such that
	\begin{equation}\label{lipmuv}
	|f(\x)-f(\y)-\langle\nabla f(\y),\x-\y\rangle|\leq\sum_{i=1}^{n}\frac{L_i(f)}{2}\|{\x}_i-{\y}_i\|_2^2,
	\end{equation}
	for any $\x=[\x_1;\cdots;\x_n]$ and $\y=[\y_1;\cdots;\y_n]$ with $\x_i, \y_i\in\mathbb{R}^{n_i}$. We call $L_i(f)$'s as Lipschitz constants of $\nabla f$. 
\end{definition}
Note that the Lipschitz continuity of the multivariable function is
crucial for the extension of IRNN for (\ref{eq_genpro2}). This definition is completely new and it is different from the one block variable case defined in (\ref{deflip1}). For $n = 1$, (\ref{lipmuv}) holds if (\ref{deflip1}) holds (Lemma 1.2.3 in \cite{nesterov2004introductory}). This motivates the above definition. But note that (\ref{deflip1}) does not guarantee to hold based on (\ref{lipmuv}). So the definition of the Lipschitz continuity of the multivariable function is different from (\ref{deflip1}). This makes the extension of IRNN for problem (\ref{eq_genpro2}) nontrivial. A widely used function which satisfies (\ref{lipmuv}) is $f(\x)=\frac{1}{2}\left\|\sumi\mathbf{A}_i\x_i-\mathbf{b}\right\|^2_2$. Its Lipschitz constants are $L_i(f)={m}\|\mathbf{A}_i\|_2^2$, $i=1,\cdots,n$, where $\|\mathbf{A}_i\|_2$ denotes the spectral norm of matrix $\mathbf{A}_i$. This is easy to verified by using the property $\left\|\sumi\mathbf{A}_i(\x_i-\y_i)\right\|^2_2\leq m\left\|\mathbf{A}_i(\x_i-\y_i)\right\|^2_2\leq m\|\mathbf{A}_i\|_2^2\|\x_i-\y_i\|_2^2$, where $\y_i$'s are of compatible size.

In theory, we prove that IRNN for (\ref{eq_genpro2}) also has the convergence guarantee. In practice, we propose a new nonconvex low rank tensor representation problem which is a special case of (\ref{eq_genpro2}) for subspace clustering. The results demonstrate the effectiveness of nonconvex models over the convex counterpart. 

In summary, the contributions of this paper are as follows.
\begin{itemize}
\item Motivated from the nonconvex surrogates $g$ of $L_0$-norm in Table \ref{tab_nonpenlty}, we propose to use a new family of nonconvex surrogates $\sumi g(\sigma_i(\X))$ to approximate the rank function. Then we propose the Iteratively Reweighted Nuclear Norm (IRNN) method to solve the nonconvex nonsmooth low rank minization problem (\ref{eq_genpro}). 
\item We further extend IRNN to solve the nonconvex nonsmooth low rank minimization problem (\ref{eq_genpro2}) with $p\ \geq2$ blocks of variables. Note that such an extension is nontrivial based on our new definition of Lipschitz continuity of the multivariable function in (\ref{lipmuv}). In theory, we prove that IRNN converges with decreasing objective function values and any limit point is a stationary point.
\item For applications, we apply the nonconvex low rank models on image recovery and subspace clustering. Extensive experiments on both synthesized and real-world data well demonstrate the effectiveness of the nonconvex models. 
\end{itemize}

The remainder of this paper is organized as follows: Section \ref{sec_IRNN} presents the IRNN method for solving problem (\ref{eq_genpro}). Section \ref{sec_extension} extends IRNN for solving problem (\ref{eq_genpro2}) and provides the convergence analysis. The experimental results are presented in Section \ref{sec_exp}. Finally, we conclude this paper in Section \ref{sec_con}.

\section{Nonconvex Nonsmooth Low-Rank Minimization}\label{sec_IRNN}
In this section, we show how to solve the general problem (\ref{eq_genpro}). Note that $g$ in (\ref{eq_genpro}) is not necessarily smooth. An known example is the Capped $L_1$ norm, see Figure \ref{fig_nonconfun}. To handle the nonsmooth penalty $g$, we first introduce the concept of supergradient defined on the concave function.
\subsection{Supergradient of a Concave Function}
\label{sec_21}
\begin{figure}[!t]
	\centering
	\includegraphics[width=0.45\textwidth]{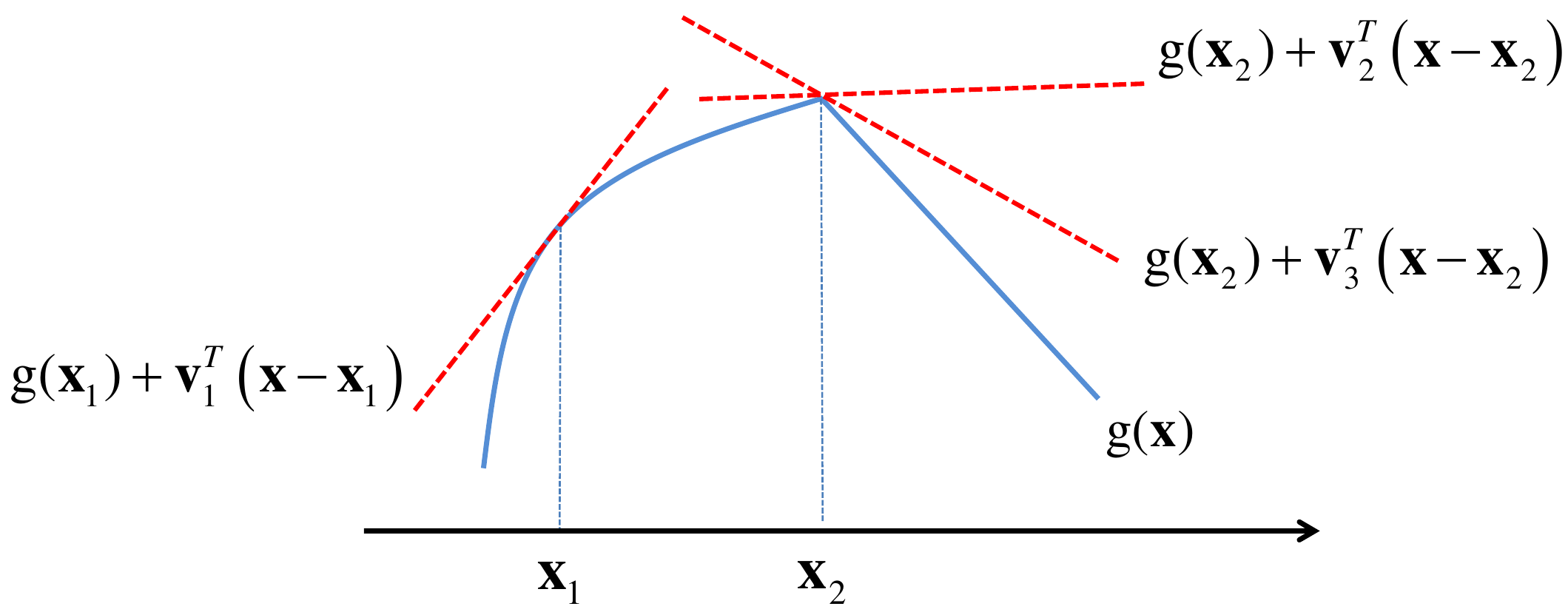}
	\caption{\small{Supergraidients of a concave function. $\mathbf{v}_1$ is a supergradient at $\mathbf{x}_1$, and $\mathbf{v}_2$ and $\mathbf{v}_3$ are supergradients at $\mathbf{x}_2$.}}
	\label{fig_supergradient}
	\vspace{-1em}
\end{figure}

If $g$ is convex but nonsmooth, its subgradient $\mathbf{u}$ at $\x$ is defined as
\begin{equation}\label{eq_convexfun}
g(\x)+\langle\mathbf{u},\y-\x\rangle\leq g(\y).
\end{equation}
If $g$ is concave and differentiable at $\x$, it is known that
\begin{equation}\label{eq_concavefun}
g(\x)+\langle\nabla g(\x),\y-\x\rangle\geq g(\y).
\end{equation}
Inspired by (\ref{eq_concavefun}), we can define the supergradient of concave $g$ at the nonsmooth point $\x$  \cite{border20011}.
\begin{definition}
	Let $g: \mathbb{R}^n\rightarrow\mathbb{R}$ be concave. A vector $\mathbf{v}$ is a supergradient of $g$ at the point $\x\in\mathbb{R}^n$ if for every $\y\in\mathbb{R}^n$, the following inequality holds
	\begin{equation}\label{eq_defsuper}
	g(\x)+\langle \mathbf{v},\y-\x\rangle\geq g(\y).
	\end{equation}
\end{definition}
The supergradient at a nonsmooth point may not be unique. All supergradients of $g$ at $\x$ are called the superdifferential of $g$ at $\x$. We denote the set of all the supergradients at $\x$ as $\partial g(\x)$. If $g$ is differentiable at $\x$, then $\nabla g(\x)$ is the unique supergradient, i.e., $\partial g(\x)=\{\nabla g(\x)\}$. Figure \ref{fig_supergradient} illustrates the supergradients of a concave function at both differentiable and nondifferentiable points.

For concave $g$, $-g$ is convex, and vice versa. From this fact, we have the following relationship between the supergradient of $g$ and the subgradient of $-g$.
\begin{lem}\label{lem_super_sub}
	Let $g(\x)$ be concave and $h(\x)=-g(\x)$. For any $\mathbf{v}\in\partial g(\x)$, $\mathbf{u}=-\mathbf{v}\in\partial h(\x)$, and vice versa.
\end{lem}

It is trivial to prove the above fact by using (\ref{eq_convexfun}) and (\ref{eq_defsuper}). The relationship of the supergradient and subgradient shown in Lemma \ref{lem_super_sub} is useful for exploring some properties of the supergradient. It is known that the subdiffierential of a convex function $h$ is a
monotone operator, i.e.,
\begin{equation}\label{eq_monotone}
\langle\mathbf{u}-\mathbf{v},\x	-\y\rangle\geq 0,
\end{equation}
for any $\mathbf{u}\in\partial h(\x)$, $\mathbf{v}\in\partial h(\y)$. Now we show that the superdifferential of a concave function is an antimonotone operator.
\begin{lem}\label{lem_antimo}
	The superdifferential of a concave function $g$ is an antimonotone operator, i.e.,
	\begin{equation}\label{eq_antimo}
	\langle\mathbf{u}-\mathbf{v},\x	-\y\rangle\leq 0,
	\end{equation}
	for any $\mathbf{u}\in\partial g(\x)$ and $\mathbf{v}\in\partial g(\y)$.
\end{lem}
The above result can be easily proved by Lemma \ref{lem_super_sub} and (\ref{eq_monotone}).

The antimonotone property of the supergradient of concave function in Lemma \ref{lem_antimo} is important in this work. Suppose that $g: \mathbb{R}\rightarrow\mathbb{R}$ satisfies the assumption \textbf{A1}, then (\ref{eq_antimo}) implies that
\begin{equation}\label{eq_antimono}
u\geq v, \text{ for any } u\in\partial g(x) \text{ and } v\in\partial g(y),
\end{equation}
when $x\leq y$. That is to say, the supergradient of $g$ is monotonically decreasing on $[0,\infty)$. The supergradients of some usual concave functions are shown in Table \ref{tab_nonpenlty}. We also visualize them in Figure \ref{fig_nonconfun}. Note that for the $L_p$ penalty, we further define that $\partial g(0)=+\infty$. This will not affect our algorithm and convergence analysis as shown later. The Capped $L_1$ penalty is nonsmooth at $\theta=\gamma$ with its superdifferential $\partial g(\gamma)=[0,\lambda]$.

\subsection{Iteratively Reweighted Nuclear Norm Algorithm}
\label{sec_22}
In this subsection, based on the above concept of the supergradient of concave function, we show how to solve the general nonconvex and possibly nonsmooth  problem (\ref{eq_genpro}). For the simplicity of notation, we denote $\sigma_1\geq\sigma_2\geq\cdots\geq\sigma_m$ as the singular values of $\X$. The variable $\X$ in the $k$-th iteration is denoted as $\X^k$ and $\sigma_i^k=\sigma_i(\X^k)$ is the $i$-th singular value of $\X^k$.

In assumption \textbf{A1}, $g$ is concave on $[0,\infty)$. So, by the definition (\ref{eq_defsuper}) of the supergradient, we have
\begin{equation}\label{eq_concaveg}
g(\sigma_i)\leq g(\sigma^k_i)+w_i^k(\sigma_i-\sigma_i^k),
\end{equation}
where
\begin{equation}\label{eq_updatew}
w_i^k\in\partial g(\sigma_i^k).
\end{equation}
Since $\sigma_1^k\geq\sigma_2^k\geq\cdots\geq\sigma_m^k\geq0$, by the antimonotone property of supergradient (\ref{eq_antimono}), we have
\begin{equation}\label{eq_incresw}
0\leq w_1^k\leq w_2^k\leq\cdots\leq w_m^k.
\end{equation}
In (\ref{eq_incresw}), the nonnegativeness of $w_i^k$'s is due to the monotonically increasing property of $g$ in assumption \textbf{A1}. As we will see later, property (\ref{eq_incresw}) plays an important role for solving the subproblem of our proposed IRNN.

Motivated by (\ref{eq_concaveg}), we may use its right hand side as a surrogate of $g(\sigma_i)$ in (\ref{eq_genpro}). Thus we may solve the following relaxed problem to update $\X^{k+1}$:
\begin{equation}\label{eq_updatexold}
\begin{split}
{\X}^{k+1}=&\arg\min_{\X} \sum_{i=1}^m g(\sigma^k_i)+w_i^k(\sigma_i-\sigma_i^k)+f(\X)\\
=&\arg\min_{\X}\sum_{i=1}^m w_i^k\sigma_i+f(\X).
\end{split}
\end{equation}
Problem (\ref{eq_updatexold}) is a weighted nuclear norm regularized problem. The updating rule (\ref{eq_updatexold}) can be regarded as an extension of the Iteratively Reweighted $L_1$ (IRL1) algorithm \cite{candes2008enhancing}  for the weighted $L_1$-norm problem
\begin{equation}\label{eqwl1}
\min_{\x} \sumi w_i^k|x_i|+l(\x).
\end{equation}
However, the weighted nuclear norm  in (\ref{eq_updatexold}) is nonconvex (it is convex if and only if $w_1^k\geq w_2^k\geq\cdots\geq w_m^k\geq0$ \cite{chen2012reduced}), while the weighted $L_1$-norm in (\ref{eqwl1}) is convex. For convex $f$ in (\ref{eq_updatexold}) and $l$ in (\ref{eqwl1}), solving the nonconvex problem (\ref{eq_updatexold}) is much more challenging than the convex weighted $L_1$-norm problem. In fact, it is not easier than solving the original problem (\ref{eq_genpro}).

\begin{algorithm}[t]
	\caption{Solving problem (\ref{eq_genpro}) by IRNN}
	\textbf{Input:} $\mu>L(f)$ - A Lipschitz constant of $\nabla f$.\\
	\textbf{Initialize:} $k=0$, ${\X}^k$, and $w^k_i$, $i=1,\cdots,m$.\\
	\textbf{Output:} $\X^*$. \\
	\textbf{while} not converge \textbf{do}
	\begin{enumerate}
		\item Update ${\X}^{k+1}$ by solving problem (\ref{eq_updatexlin}).
		\item Update the weights $w_i^{k+1}$, $i=1,\cdots,m$, by
		\begin{equation}
		w_i^{k+1}\in\partial g\left(\sigma_i({\X}^{k+1})\right).
		\end{equation}
		\textbf{end while}
	\end{enumerate}
	\label{alg_lirnn}
\end{algorithm}

Instead of updating ${\X}^{k+1}$ by solving (\ref{eq_updatexold}), we linearize $f(\X)$ at ${\X^k}$ and add a proximal term: 
\begin{equation}\label{eqappf}
f({\X})\approx f({\X}^k)+\langle\nabla f({\X}^k),{\X}-{\X}^k\rangle+\frac{\mu}{2}||{\X}-{\X}^k||_F^2,
\end{equation}
where $\mu> L(f)$. Such a choice of $\mu$ guarantees the convergence of our algorithm as shown later. Then we use the right hand sides of (\ref{eq_concaveg}) and (\ref{eqappf}) as surrogates of $g$ and $f$ in (\ref{eq_genpro}), and update ${\X}^{k+1}$ by solving
\begin{equation}\label{eq_updatexlin}
\begin{split}
{\X}^{k+1}=&\arg\min_{{\X}} \sum_{i=1}^m g(\sigma^k_i)+w_i^k(\sigma_i-\sigma_i^k)\\
&+f({\X}^k)+\langle\nabla f({\X}^k),{\X}-{\X}^k\rangle+\frac{\mu}{2}||{\X}-{\X}^k||_F^2\\
=& \arg\min_{{\X}} \sum_{i=1}^mw_i^k\sigma_i+\frac{\mu}{2}\left\|{\X}-\left({\X}^k-\frac{1}{\mu}\nabla f({\X}^k)\right)\right\|_F^2.
\end{split}
\end{equation}
Solving (\ref{eq_updatexlin}) is equivalent to computing the proximity operator of the weighted nuclear norm. Due to (\ref{eq_incresw}), the solution to (\ref{eq_updatexlin}) has a closed form  despite that it is nonconvex.

\begin{lem}\label{Lem_ineq1}
	\cite[Theorem 2.3]{chen2012reduced} 
	For any $\lambda>0$, ${\Y}\in\mathbb{R}^{m\times n}$ and $0\leq w_1\leq w_2\leq\cdots\leq w_s \ (s=\min(m,n))$, a globally optimal solution to the following problem
	\begin{equation}
	\min \lambda\sum_{i=1}^sw_i\sigma_i({\X})+\frac{1}{2}||{\X}-{\Y}||_F^2,
	\end{equation}
	is given by the Weighted Singular Value Thresholding (WSVT)
	\begin{equation}\label{wsvtsolution}
	{\X}^*=\bm{U}\mathcal{S}_{\lambda \bm{w}}(\bm{\Sigma})\bm{V}^T,
	\end{equation}
	where ${\Y}=\bm{U}\bm{\Sigma}\bm{V}^T$ is the SVD of ${\Y}$, and $\mathcal{S}_{\lambda \bm{w}}(\bm{\Sigma})=\Diag\{(\bm{\Sigma}_{ii}-\lambda w_i)_+\}$.
\end{lem}
From Lemma \ref{Lem_ineq1}, it can be seen that to solve (\ref{eq_updatexlin}) by using  (\ref{wsvtsolution}), (\ref{eq_incresw}) plays an important role and it holds for all $g$ satisfying the assumption \textbf{A1}. If $g(x)=x$, then $\sumi g(\sigma_i)$ reduces to the convex nuclear norm $\|\X\|_*$. In this case, $w_i^k=1$ for all $i=1,\cdots,m$. Then WSVT reduces to the conventional Singular Value Thresholding (SVT) \cite{cai2010singular}, which is an important subroutine in convex low rank optimization. The updating rule (\ref{eq_updatexlin}) then reduces  to the known proximal gradient method \cite{beck2009fast}.

After updating ${\X}^{k+1}$ by solving (\ref{eq_updatexlin}), we then update the weights $w_i^{k+1}\in\partial g\left(\sigma_i({\X}^{k+1})\right)$, $i=1,\cdots,m$. Iteratively updating $\X^{k+1}$ and the weights corresponding to its singular values leads to the proposed Iteratively Reweighted Nuclear Norm (IRNN) algorithm. The whole procedure of IRNN is shown in Algorithm \ref{alg_lirnn}. If the Lipschitz constant $L(f)$ is not known or computable, the backtracking rule can be used to estimate $\mu$ in each iteration \cite{beck2009fast}.

It is worth mentioning that for the $L_p$ penalty, if $\sigma_i^k=0$, then $w_i^k\in\partial g(\sigma_i^k)=\{+\infty\}$. By the updating rule of $\X^{k+1}$ in (\ref{eq_updatexlin}), we have $\sigma_i^{k+1}=0$. This guarantees that the rank of the sequence $\{\X^k\}$ is nonincreasing.


In theory, we can prove that IRNN converges. Since IRNN is a special case of IRNN with Parallel Splitting (IRNN-PS) in Section \ref{sec_IRNNPS}, so we only give the convergence results of IRNN-PS later.

At the end of this section, we would like to remark some more differences between previous work and ours. 
\begin{itemize}
	\item Our IRNN and IRNN-PS for nonconvex low rank minimization are different from previous iteratively reweighted solvers for nonconvex sparse minimization, e.g., \cite{candes2008enhancing,lai2011unconstrained}. The key difference is that the weighted nuclear norm regularized problem is nonconvex while the weighted $L_1$-norm regularized problem is convex. This  makes the convergence analysis different.
	\item Our IRNN and IRNN-PS utilize the common properties instead of specific ones of the nonconvex surrogates of $L_0$-norm. This makes them much more general than many previous nonconvex low rank  solvers, e.g., \cite{lai2013improved,hu2012fast,dong2014compressive}, which target for some special nonconvex problems.
\end{itemize}

\section{Extensions of IRNN and the Convergence Analysis}\label{sec_extension}
\label{sec_IRNNPS}
In this section, we extend IRNN to solve two types of problems which are more general than (\ref{eq_genpro}). The first one is to solve some similar problems as (\ref{eq_genpro}) but with more general nonconvex penalties. The second one is to solve problem (\ref{eq_genpro2}) which has $p \geq2$ blocks of variables.

\subsection{IRNN for the Problems with More General Nonconvex Penalties}
\label{sec_4}
IRNN can be extended to solve the following problem
\begin{equation}
\min_{\X} \sum_{i=1}^m g_i(\sigma_i(\X))+f(\X),
\end{equation}
where $g_i$'s are concave and their supergradients satisfy $0\leq v_1\leq v_2\leq\cdots\leq v_m$ for any $v_i\in\partial g_i(\sigma_i(\X))$, $i=1,\cdots,m$. The truncated nuclear norm $||\X||_r=\sum_{i=r+1}^m\sigma_i(\X)$ \cite{hu2012fast} is an interesting example. Indeed, let
\begin{equation}
g_i(x)=\begin{cases} 0, & i=1,\cdots,r,\\ x, & i=r+1,\cdots,m.\end{cases}
\end{equation}
Then  $||\X||_r=\sum_{i=1}^mg_i(\sigma_i(\X))$ and its supergradients is
\begin{equation}
\partial g_i(x)=\begin{cases} 0, & i=1,\cdots,r,\\ 1, & i=r+1,\cdots,m.\end{cases}
\end{equation}
Compared with the alternating updating algorithm in \cite{hu2012fast}, which require double loops, our IRNN will be more efficient and with stronger convergence guarantee.

\subsection{IRNN for the Multi-Blocks Problem (\ref{eq_genpro2})}
The multi-blocks problem (\ref{eq_genpro2}) also has some applications in computer vision. An example is the Latent Low Rank Representation (LatLRR) problem \cite{liu2011latent} 
\begin{equation}\label{latlrr}
\min_{\mathbf{L},\mathbf{R}} \|\mathbf{L}\|_*+\|\mathbf{R}\|_*+\frac{\lambda}{2}\|\mathbf{L}\X+\X\mathbf{R}-\X\|_F^2.
\end{equation}
Here we propose a more general Tensor Low Rank Representation (TLRR) as follows
\begin{equation}\label{tlrr}
\min_{\PP_j\in\mathbb{R}^{m_j\times m_j}} \sumj\lambda_j\|{\PP}_j\|_*+\frac{1}{2}\left\|\Xt-\sumj\Xt\times_j{\PP}_j\right\|_F^2,
\end{equation}
where $\Xt\in\mathbb{R}^{m_1\times\cdots\times m_p}$ is an $p$-way tensor and $\Xt\times_j{\PP}_j$ denotes the $j$-mode product \cite{kolda2009tensor}. TLRR is an extension of LRR \cite{robustlrr} and LatLRR. It can also be applied for subspace clustering, see Section \ref{sec_exp}. If we replace $\|{\PP}_j\|_*$ in (\ref{latlrr}) as $\sum_{i=1}^{m_j} g_j(\sigma_i(\PP_j))$ with $g_j$'s satisfying the assumption \textbf{A1}, then we have the Nonconvex TLRR (NTLRR) model which is a special case of (\ref{eq_genpro2}). 

Now we show how to solve (\ref{eq_genpro2}). Similar to (\ref{eq_updatexlin}), we update $\X_j$, $j=1,\cdots,p$, by 
\begin{align}\label{equpdatex2}
{\X}_j^{k+1}=&\arg\min_{\X_j}\sum_{i=1}^{m_j}w_{ji}^k\sigma_i({\X}_j)+\langle\nabla_jf({\X}^k),{\X}_j-{\X}_j^k\rangle\notag\\
&+\frac{\mu_j}{2}\|{\X}_j-{\X}_j^k\|^2_F,
\end{align}
where $\mu_j>L_i(f)$, the notation $\nabla_j f$ denotes the gradient of $f$ w.r.t. $\X_j$, and
\begin{equation}\label{eq_updatew2}
w_{ji}^k\in\partial g_j(\sigma_i({\X}_j^k)).
\end{equation}
Note that (\ref{equpdatex2}) and (\ref{eq_updatew2}) can be computed in parallel for $j=1,\cdots,p$. So we call such a method as IRNN with Parallel Splitting (IRNN-PS). 



\subsection{Convergence Analysis}

In this section, we give the convergence analysis of IRNN-PS for (\ref{eq_genpro2}). For the simplicity of notation, we denote $\sigma_{ji}^k=\sigma_{i}(\X_j^k)$ as the $i$-th singular value of $\X_j$ in the $k$-th iteration.


\begin{theo}\label{thm_pro3}
In problem (\ref{eq_genpro2}), assume that $g_j$'s satisfies the assumption \textbf{A1} and $\nabla f$ is Lipschitz continuous. Then the sequence $\{{\X}^k\}$ generated by IRNN-PS satisfies the following properties:
\begin{enumerate}[(1)]
\item $F({\X}^k)$ is monotonically decreasing. 
Indeed,
\begin{equation*}
F({\X}^k)-F({\X}^{k+1})\geq\sumj\frac{\mu_j-L_j(f)}{2}||{\X}_j^k-{\X}_j^{k+1}||_F^2\geq0;
\end{equation*}
\item $\lim\limits_{k\rightarrow+\infty}({\X}^k-{\X}^{k+1})=\bm{0}$;
\end{enumerate}
\end{theo}
\textit{Proof.} First, since ${\X}_j^{k+1}$ is optimal to (\ref{equpdatex2}), we have
\begin{equation*}\label{eq_proof1}
\begin{split}
&\sum_{i=1}^mw_{ji}^k\sigma_{ji}^{k+1}+\langle\nabla_j f({\X}^k),{\X}_j^{k+1}-{\X}_j^k\rangle+\frac{\mu_j}{2}||{\X}_j^{k+1}-{\X}_j^k||^2_F\\
\leq&\sum_{i=1}^mw_{ji}^k\sigma_{ji}^{k}+\langle\nabla_j f({\X}^k),{\X}^{k}_j-{\X}^k_j\rangle+\frac{\mu_j}{2}||{\X}_j^{k}-{\X}_j^k||^2_F.
\end{split}
\end{equation*}
It can be rewritten as
\begin{equation*}\label{eq_proof4}
\begin{split}
&\langle\nabla_j f({\X}^k),{\X}_j^k-{\X}_j^{k+1}\rangle\\
\geq&-\sum_{i=1}^mw_{ji}^k(\sigma_{ji}^k-\sigma_{ji}^{k+1})+\frac{\mu_j}{2}||{\X}^k-{\X}^{k+1}||_F^2.
\end{split}
\end{equation*}
Second, since $\nabla f$ is Lipschitz continuous, by (\ref{lipmuv}), we have
\begin{equation*}\label{eq_proof5}
\begin{split}
&f({\X}^k)-f({\X}^{k+1})\\
\geq&\sumj\left(\langle\nabla_j f({\X}^k),{\X}_j^k-{\X}_j^{k+1}\rangle-\frac{L_j(f)}{2}||{\X}_j^k-{\X}_j^{k+1}||_F^2\right).
\end{split}
\end{equation*}
Third, by (\ref{eq_updatew2}) and (\ref{eq_defsuper}), we have
\begin{equation*}\label{eq_proof6}
g_j(\sigma_{ji}^k)-g_j(\sigma_{ji}^{k+1})\geq w_{ji}^k(\sigma_{ji}^k-\sigma_{ji}^{k+1}).
\end{equation*}
Summing the above three equations for all $j$ and $i$ leads to
\begin{equation*}\label{eq_proof7}
\begin{split}
& F({\X}^k)-F({\X}^{k+1})\\
=&\sum_{j=1}^{p}\sum_{i=1}^{n_j}\left(g_j(\sigma_{ji}^k)-g(\sigma_{ji}^{k+1}) \right)+f({\X}^k)-f({\X}^{k+1}) \\
\geq&\sumj\frac{\mu_j-L_j(f)}{2}||{\X}^{k+1}_j-{\X}_j^k||_F^2\geq0.\\
\end{split}
\end{equation*}
Thus $F({\X}^k)$ is monotonically decreasing. Summing the above inequality for $k\geq1$, we get
\begin{equation*}\label{eq_proof9}
F({\X}^1)\geq\sumj\frac{\mu_j-L_j(f)}{2}\sum_{k=1}^{+\infty}||{\X}_j^{k+1}-{\X}_j^k||_F^2,
\end{equation*}
This implies that $\lim\limits_{k\rightarrow+\infty}({\X}^k-{\X}^{k+1})=\bm{0}$. 
$\hfill\blacksquare$

\begin{figure*}
	\centering
	\begin{subfigure}[b]{0.24\textwidth}
		\centering
		\includegraphics[width=\textwidth]{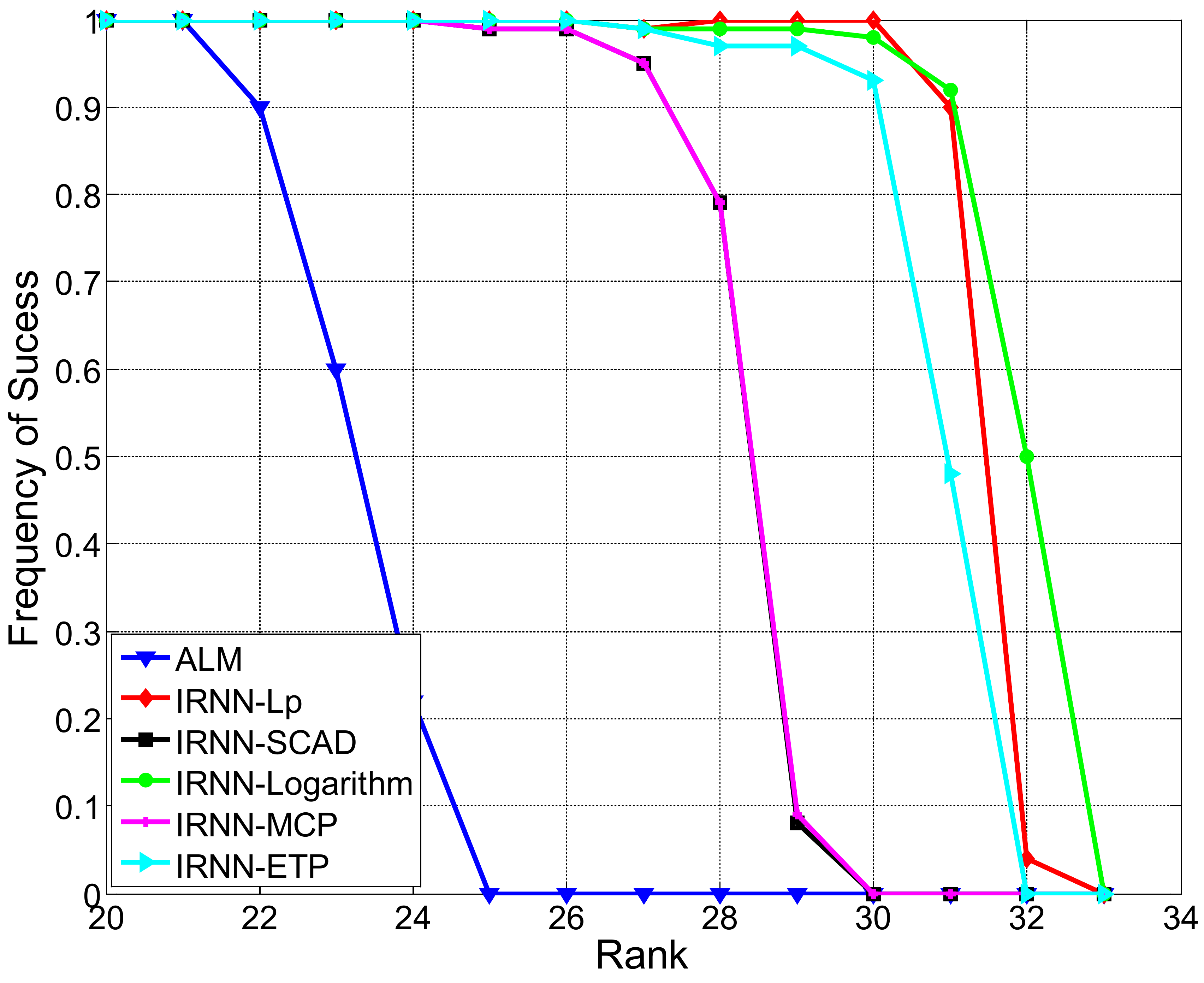}
		\caption{Random data without noise}
		\label{fig_randrecov_noiseless}
	\end{subfigure}
	\begin{subfigure}[b]{0.24\textwidth}
		\centering
		\includegraphics[width=\textwidth]{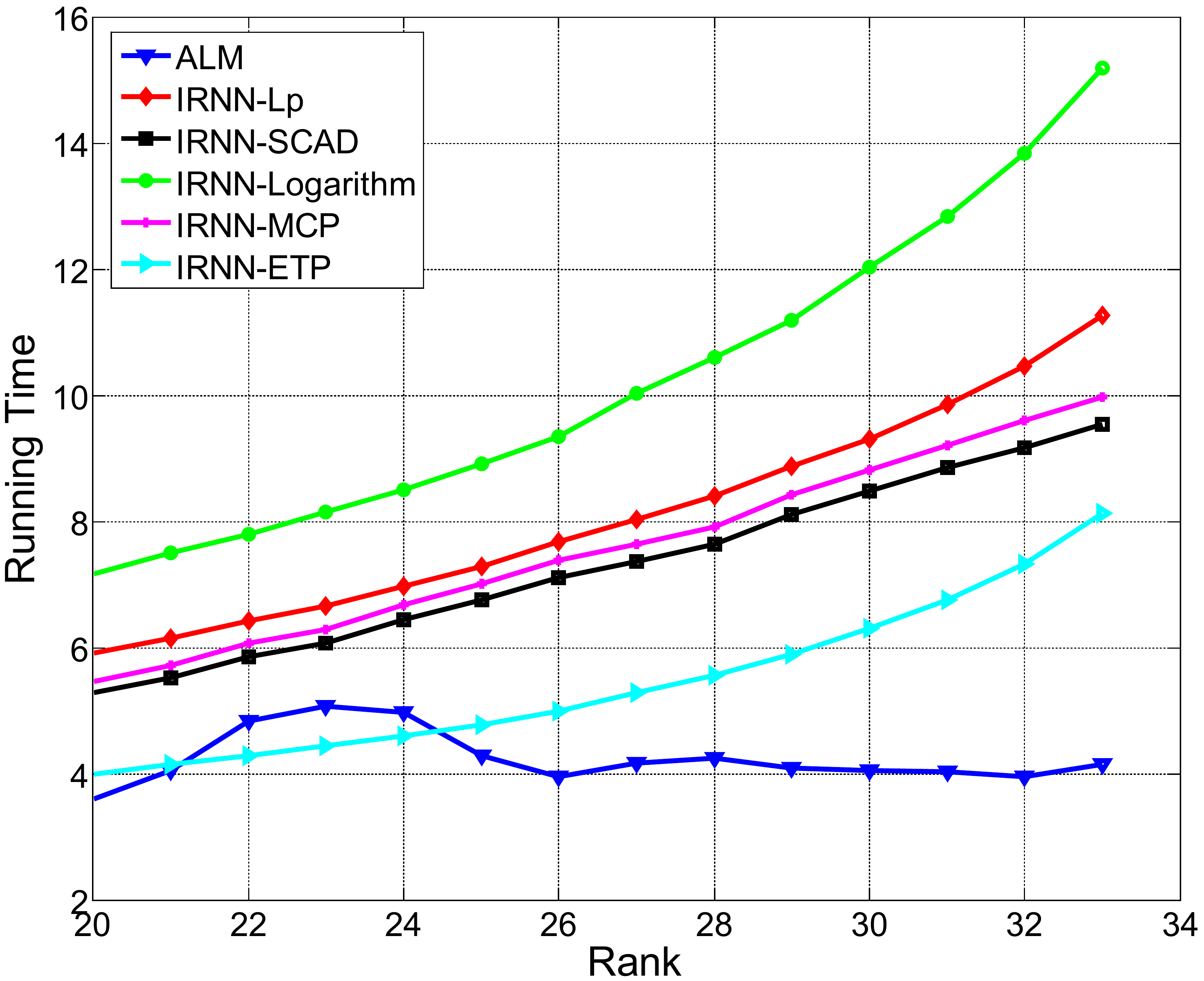}
		\caption{Running time}
		\label{fig_randrecov_time}
	\end{subfigure}    
		\begin{subfigure}[b]{0.24\textwidth}
			\centering
			\includegraphics[width=\textwidth]{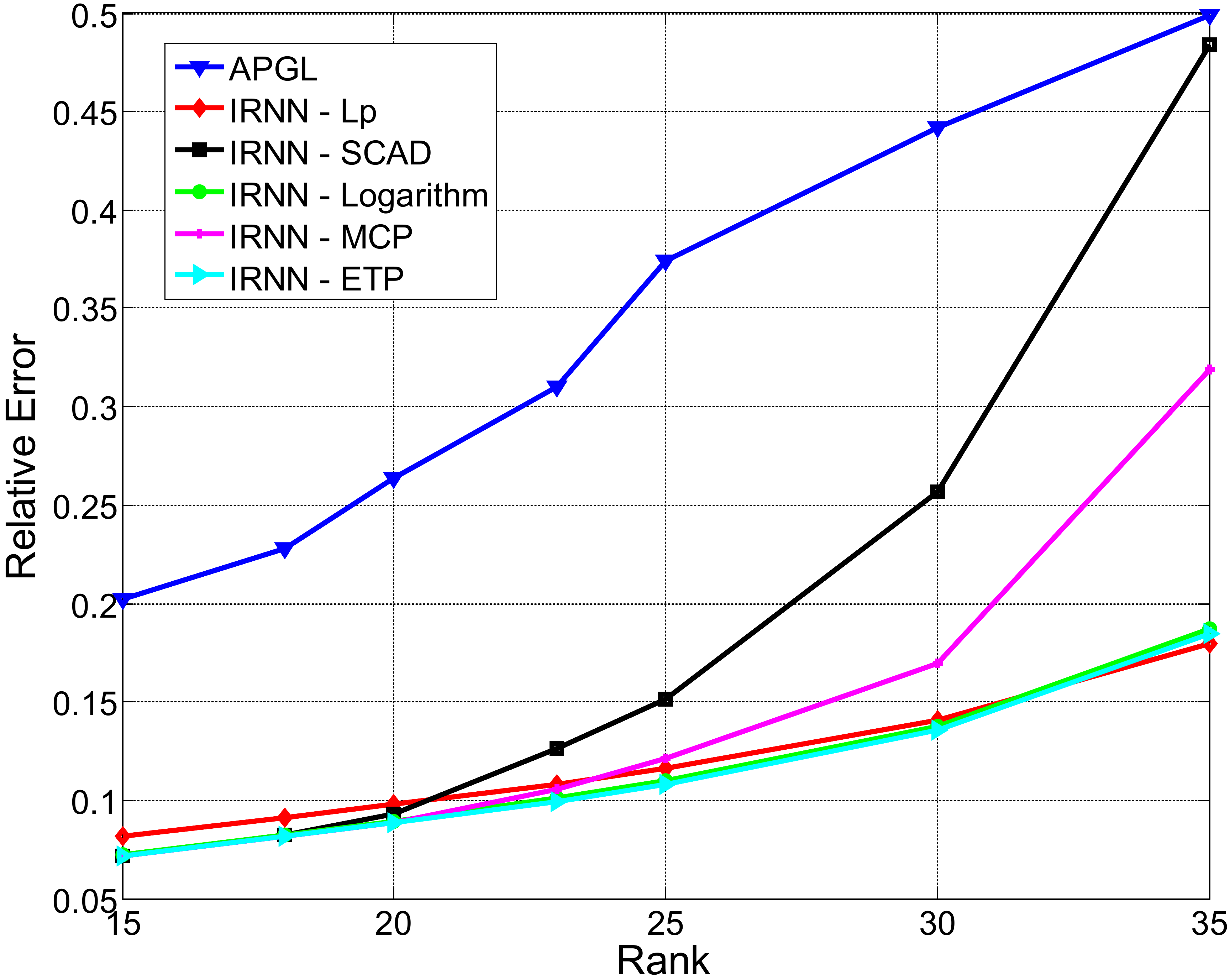}
			\caption{Random data with noises}
			\label{fig_randrecov_noise}
		\end{subfigure}
	\begin{subfigure}[b]{0.24\textwidth}
		\centering
		\includegraphics[width=\textwidth]{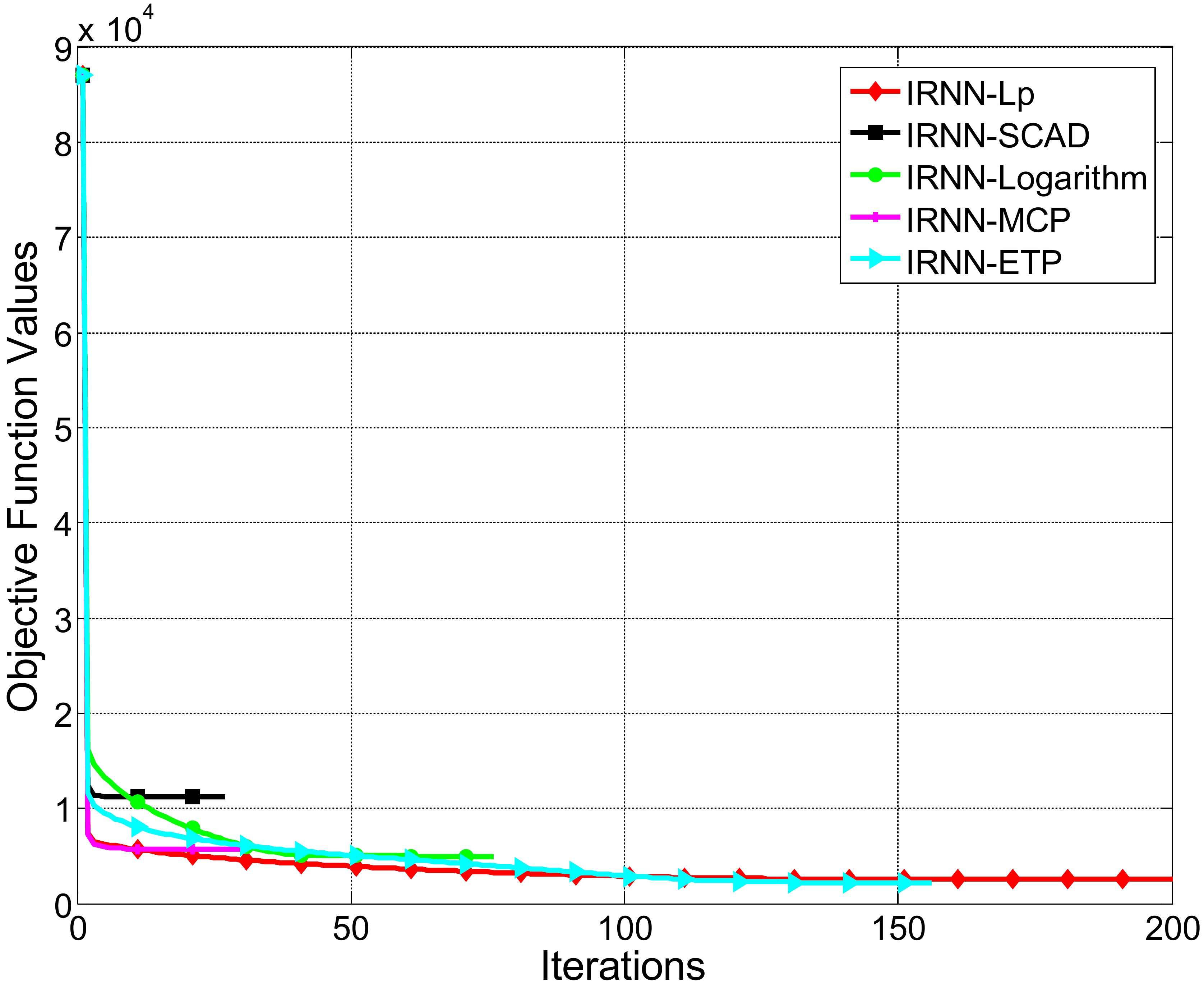}
		\caption{Convergence curves}
		\label{fig_randrecov_conv}
	\end{subfigure}  
	\caption{Low-rank matrix recovery comparison of (a) frequency of successful recovery and (b) running time on random data without noise; (c) relative error and (d) convergence curves on random data with noises.}\label{fig_randrecov}
\end{figure*}

\begin{theo}\label{thm_pro4}
In problem (\ref{eq_genpro2}), assume
 $F(\X)\rightarrow+\infty$ iff $||\X||_F\rightarrow+\infty$. Then any accumulation point ${\X}^*$ of $\{{\X}^k\}$ generated by IRNN-PS is a stationary point to (\ref{eq_genpro2}).
\end{theo}
\textit{Proof.} Due to the above assumption, $\{{\X}^k\}$ is bounded. Thus there exists a matrix ${\X}^*$ and a subsequence $\{{\X}^{k_{t}}\}$ such that ${\X}^{k_{t}}\rightarrow{\X}^*$. Note that ${\X}^{k}-{\X}^{k+1}\rightarrow\bm{0}$ in Theorem \ref{thm_pro3}, we have ${\X}^{k_{j}+1}\rightarrow{\X}^*$. Thus $\sigma_i({\X}_j^{k_t+1})\rightarrow\sigma_i({\X}_j^*)$ for $j=1,\cdots,p$ and $i=1,\cdots,n_j$. By Lemma \ref{lem_super_sub}, $w_{ji}^{k_t}\in\partial g_j(\sigma_i({\X}_j^{k_t}))$ implies that $-w_{ji}^{k_t}\in\partial \left(-g_j(\sigma_i({\X}_j^{k_t}))\right)$. From the upper semi-continuous property of the subdifferential \cite[Proposition 2.1.5]{clarke1983nonsmooth}, there exists $-w_{ji}^*\in\partial \left(-g_j(\sigma_i(\X_j^*))\right)$ such that $-w_{ji}^{k_t}\rightarrow -w_{ji}^*$. Again by Lemma \ref{lem_super_sub}, $w_{ji}^*\in\partial g_j(\sigma_i(\X_j^*))$ and $w_{ji}^{k_t}\rightarrow w_{ji}^*$.

Denote $h({\X_j},\bm{w}_j)=\sum_{i=1}^{n_j}w_{ji}\sigma_i({\X_j})$. Since ${\X}_j^{k_{t}+1}$ is optimal to (\ref{equpdatex2}), there exists
$\bm{G}_j^{k_t+1}\in\partial h({\X}_j^{k_t+1},\bm{w}_j^{k_t})$, such that
\begin{equation}\label{eq_proof11}
\bm{G}_j^{k_t+1}+\nabla_j f({\X}^{k_t})+\mu_j({\X}_j^{k_t+1}-{\X}_j^{k_t})=\bm{0}.
\end{equation}
Let $t\rightarrow+\infty$ in (\ref{eq_proof11}). Then there exists $\bm{G}_j^*\in\partial h({\X}_j^*,\bm{w}_j^*)$, such that
\begin{equation}\label{eq_proof12}
\bm{0}=\bm{G}_j^*+\nabla_j f({\X}^{*})\in\partial_j F({\X}^*).
\end{equation}
Thus ${\X}^*$ is a stationary point to (\ref{eq_genpro2}).  $\hfill\blacksquare$

\begin{figure*}
	\begin{subfigure}[b]{0.138\textwidth}
		\centering
        \includegraphics[width=\textwidth]{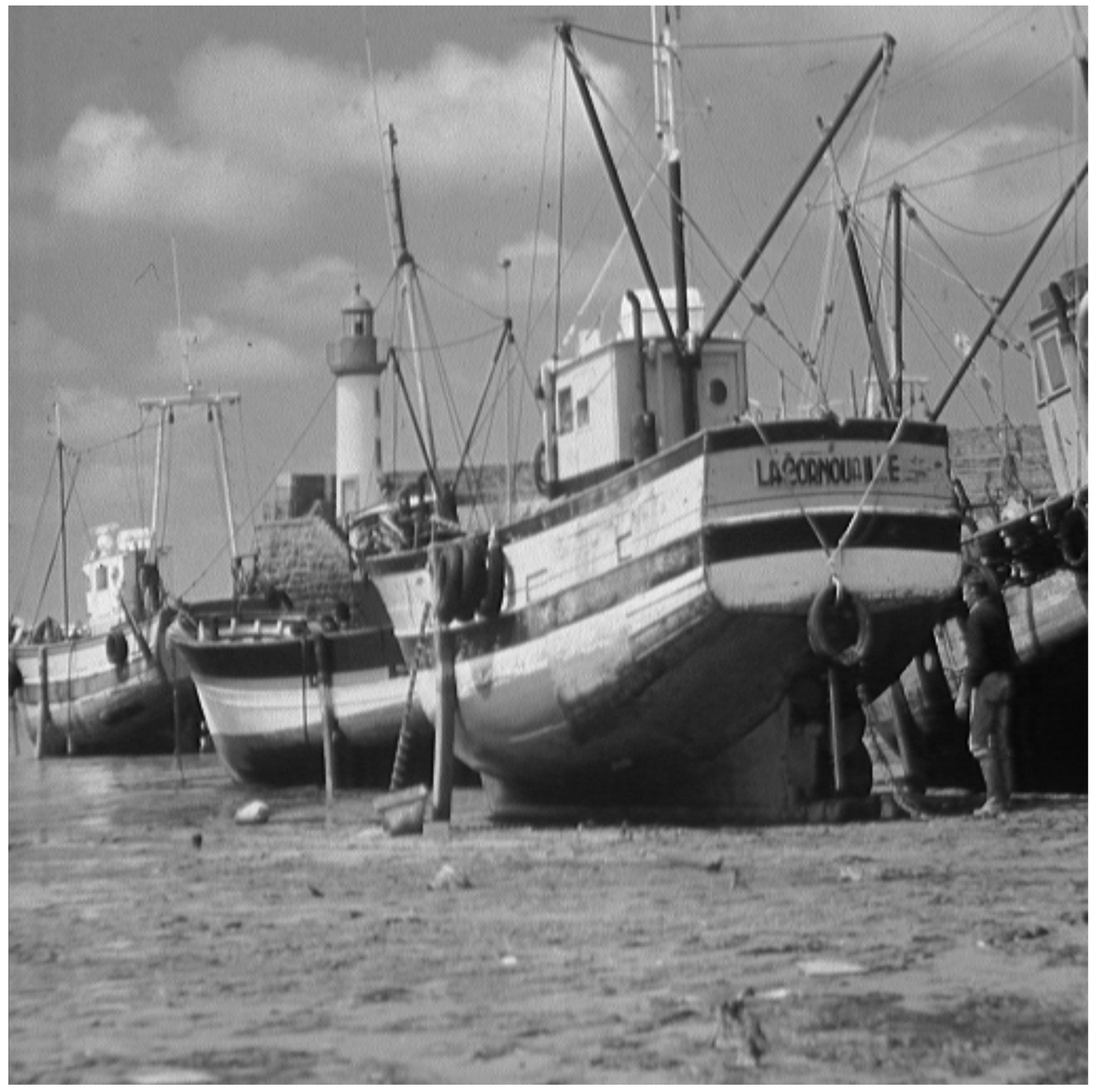}
    \end{subfigure}
        \begin{subfigure}[b]{0.138\textwidth}
		\centering
		\includegraphics[width=\textwidth]{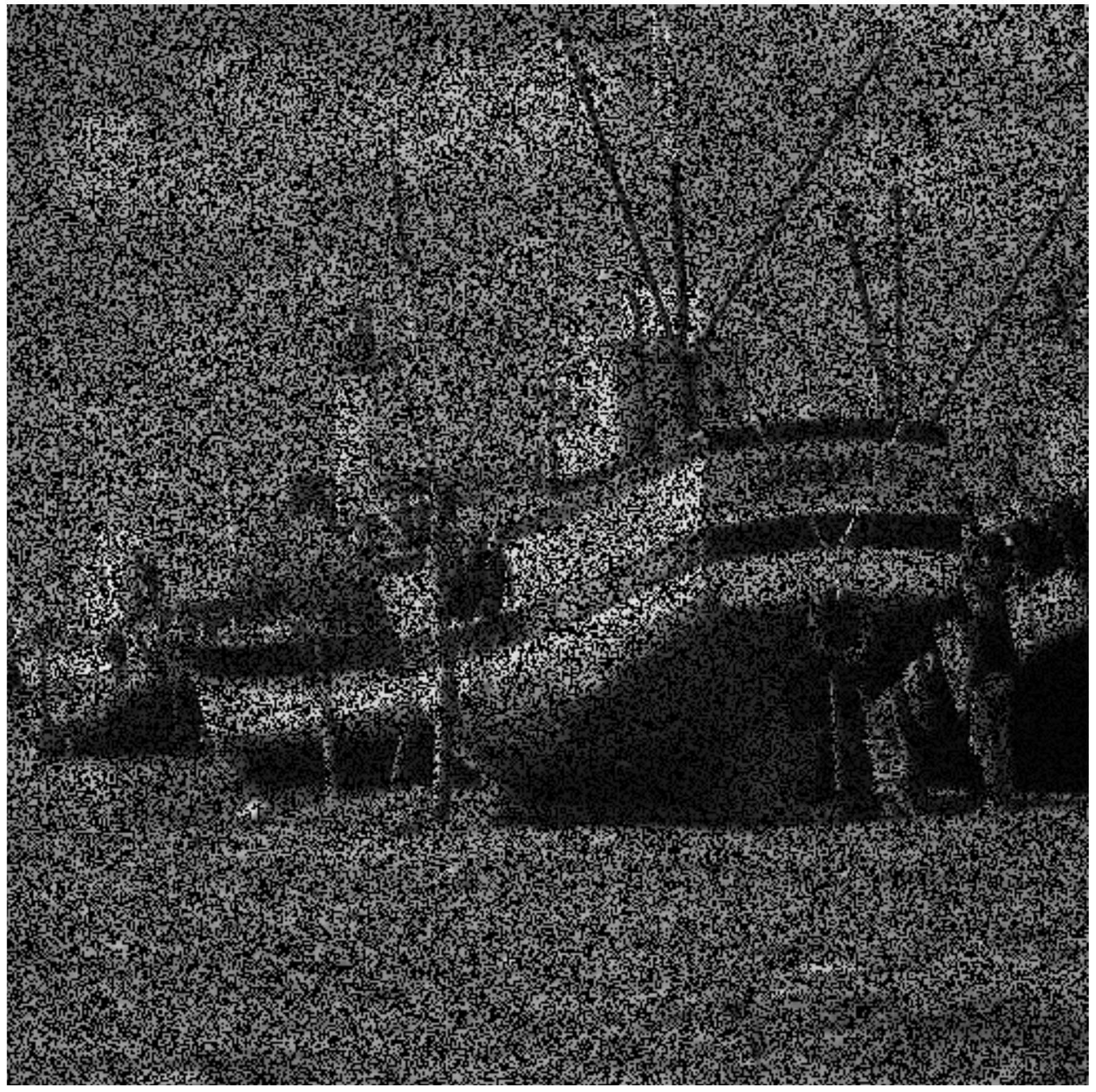}
    \end{subfigure}
    \begin{subfigure}[b]{0.138\textwidth}
		\centering
        \includegraphics[width=\textwidth]{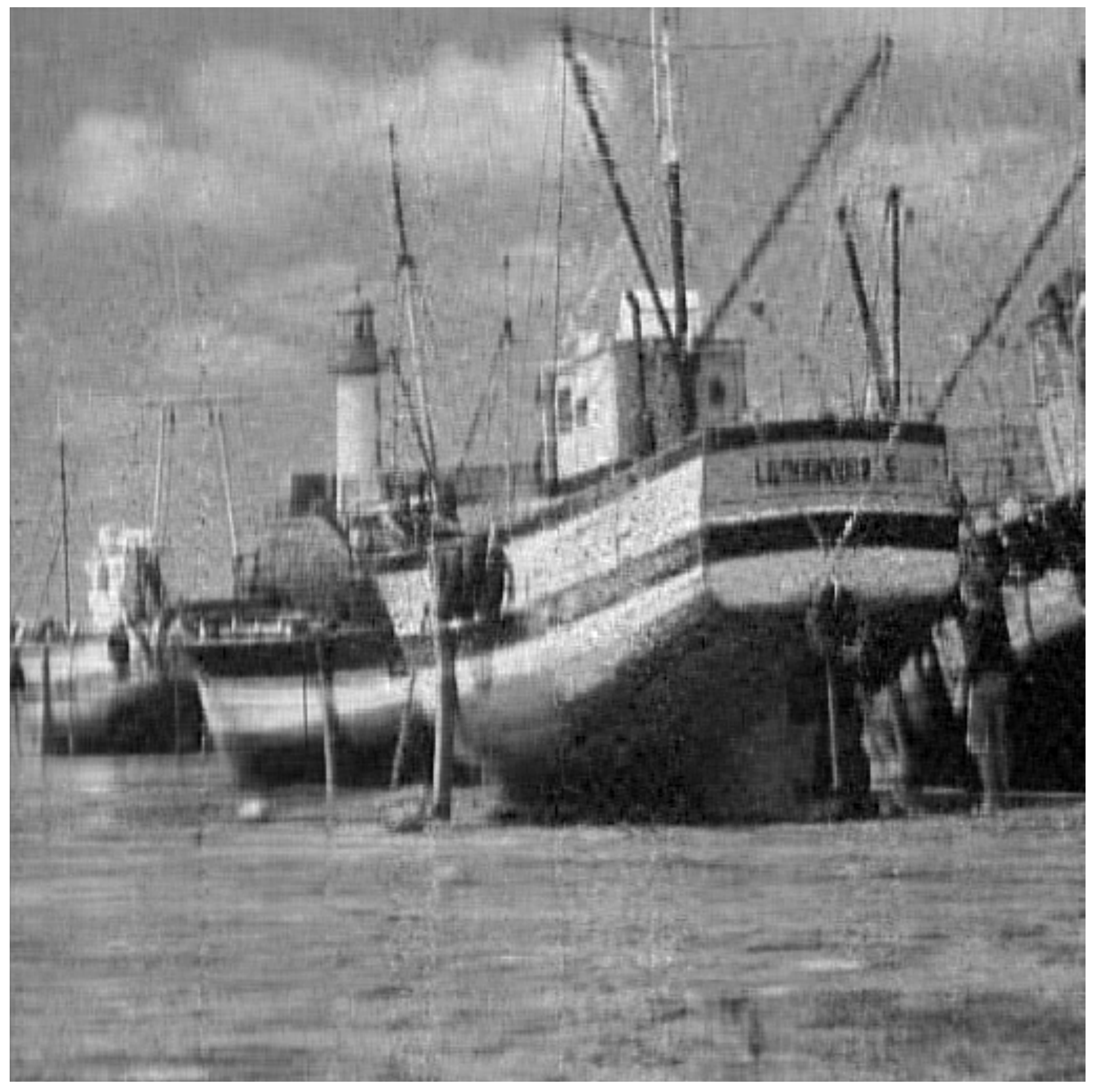}
    \end{subfigure}
    \begin{subfigure}[b]{0.138\textwidth}
		\centering
		\includegraphics[width=\textwidth]{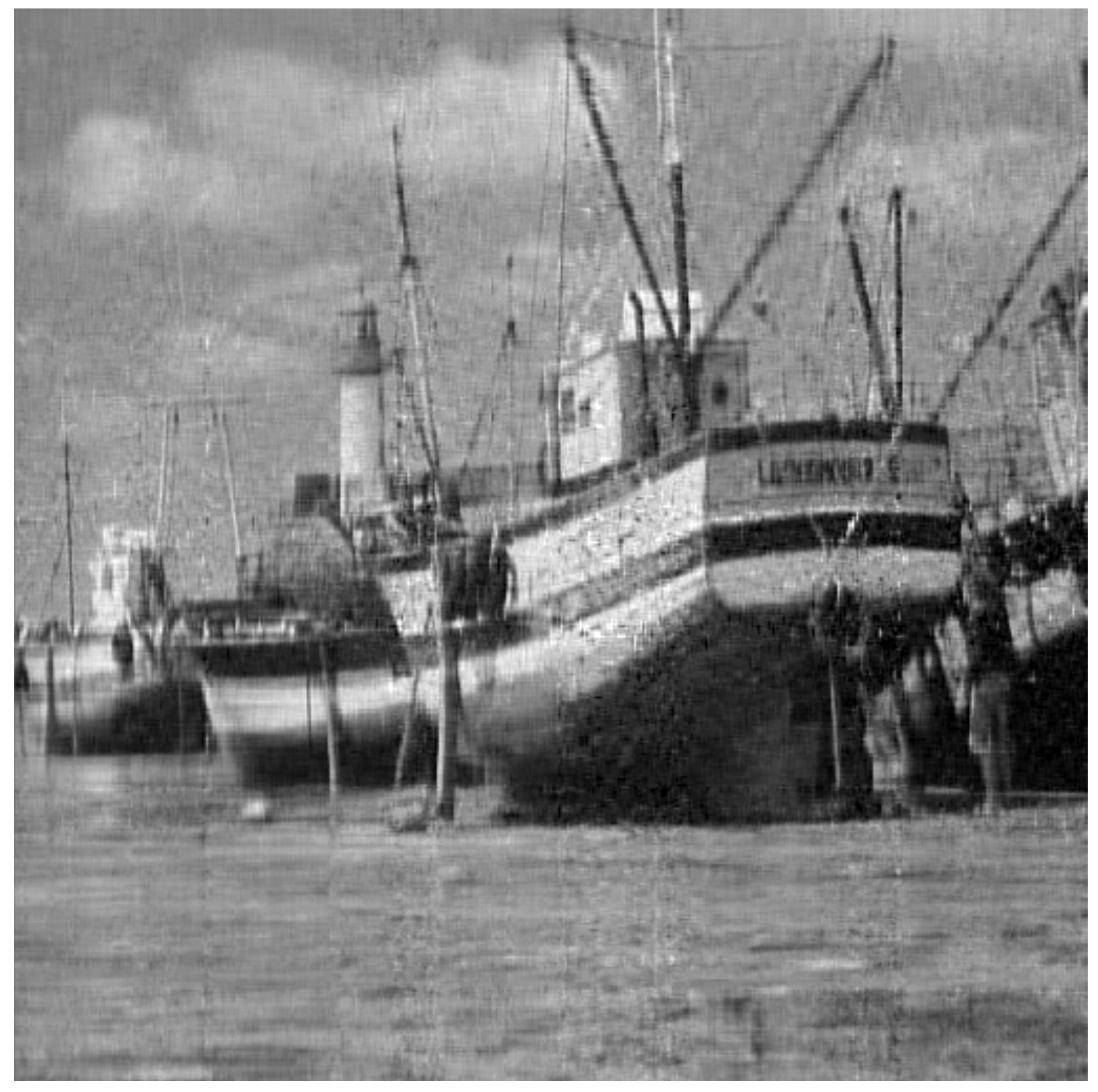}
    \end{subfigure}
    \begin{subfigure}[b]{0.138\textwidth}
		\centering
		\includegraphics[width=\textwidth]{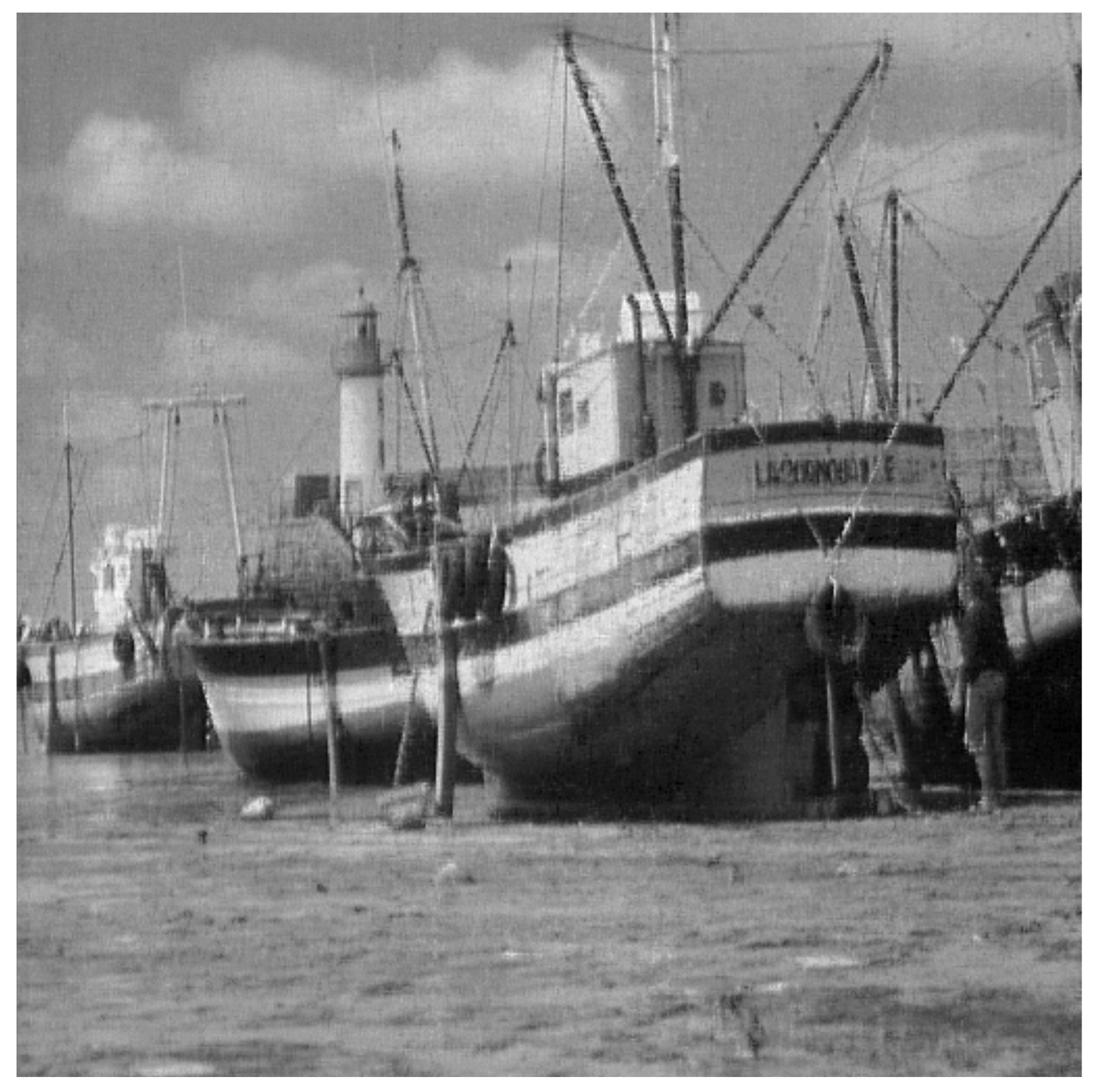}
    \end{subfigure}
    \begin{subfigure}[b]{0.138\textwidth}
		\centering
		\includegraphics[width=\textwidth]{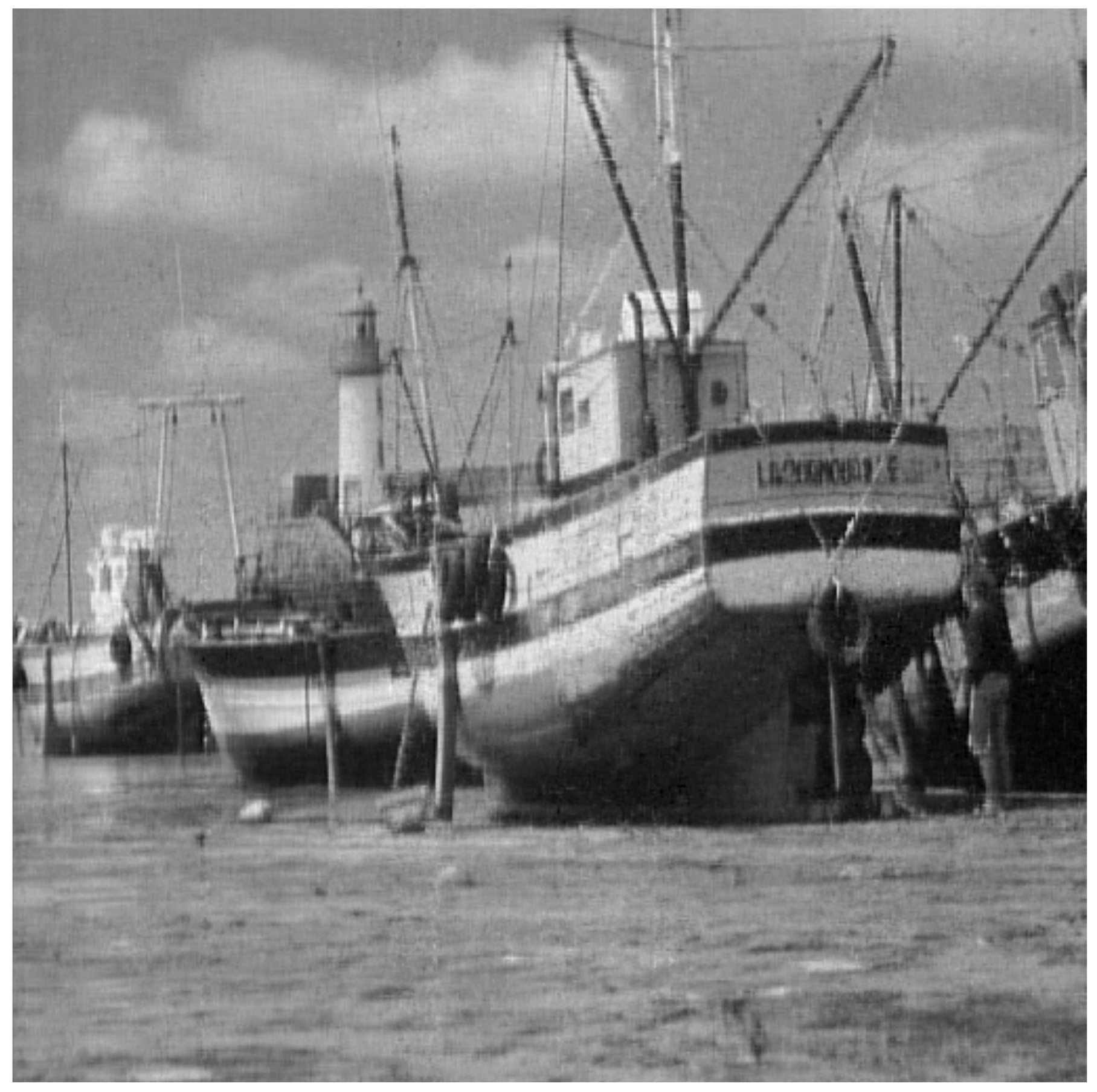}
    \end{subfigure}
	\begin{subfigure}[b]{0.138\textwidth}
		\centering
        \includegraphics[width=\textwidth]{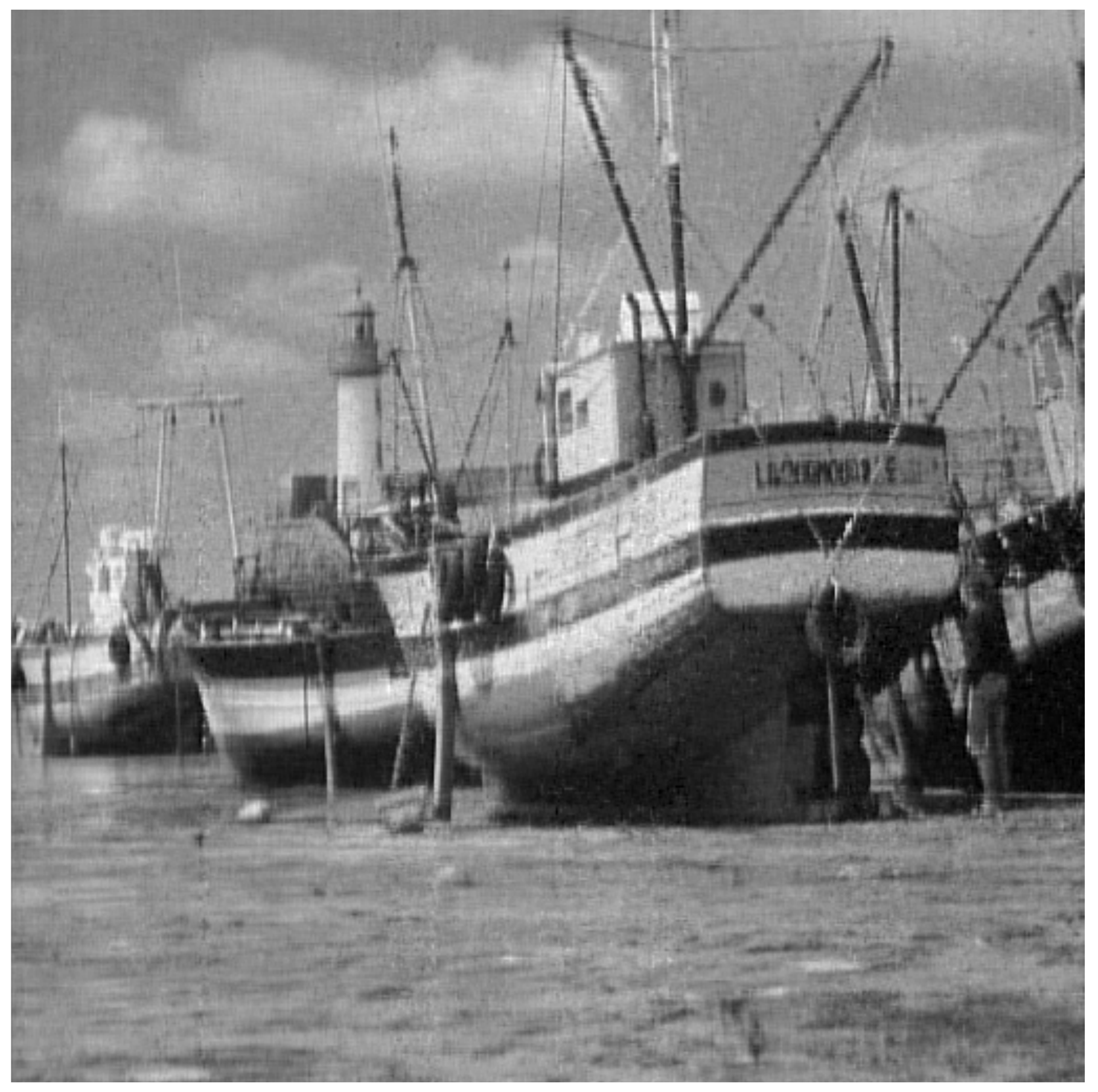}
	\end{subfigure}	
	
	\begin{subfigure}[b]{0.138\textwidth}
		\centering
        \includegraphics[width=\textwidth]{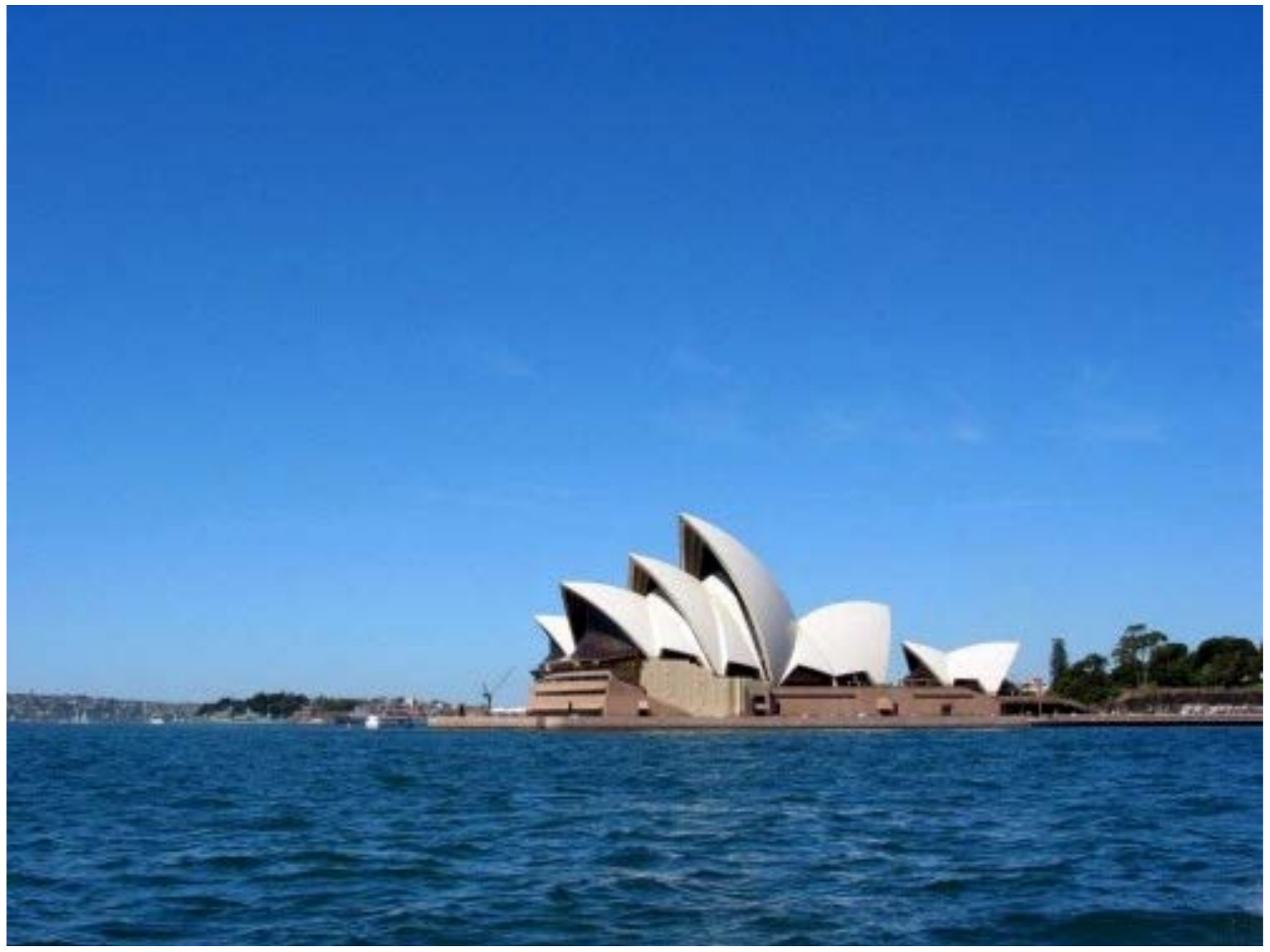}
        \caption{Original image}
	\end{subfigure}
	\begin{subfigure}[b]{0.138\textwidth}
		\centering
        \includegraphics[width=\textwidth]{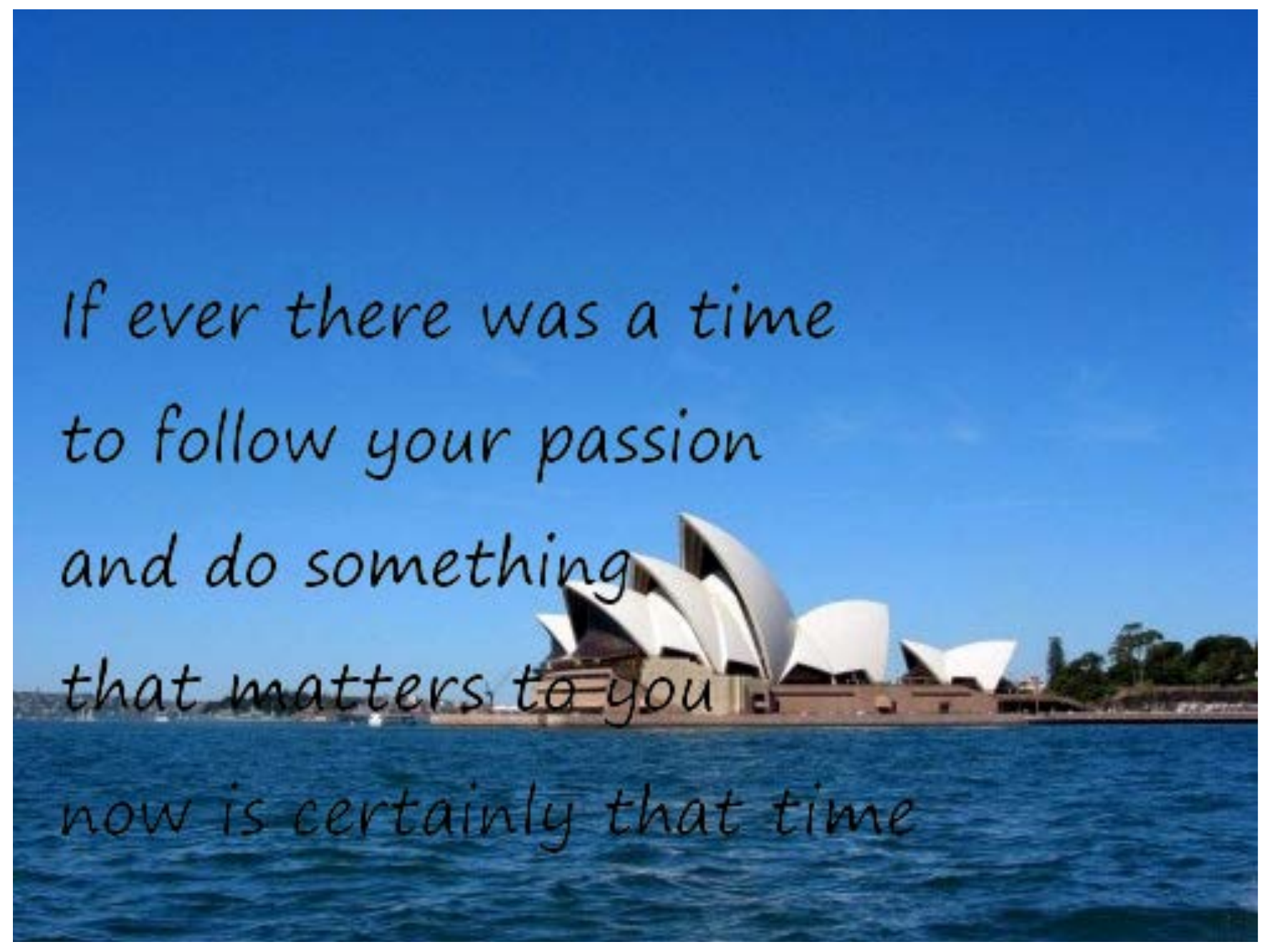}
        \caption{Noisy Image}\label{fig_gaussiannoise}
	\end{subfigure}	
	\begin{subfigure}[b]{0.138\textwidth}
		\centering
        \includegraphics[width=\textwidth]{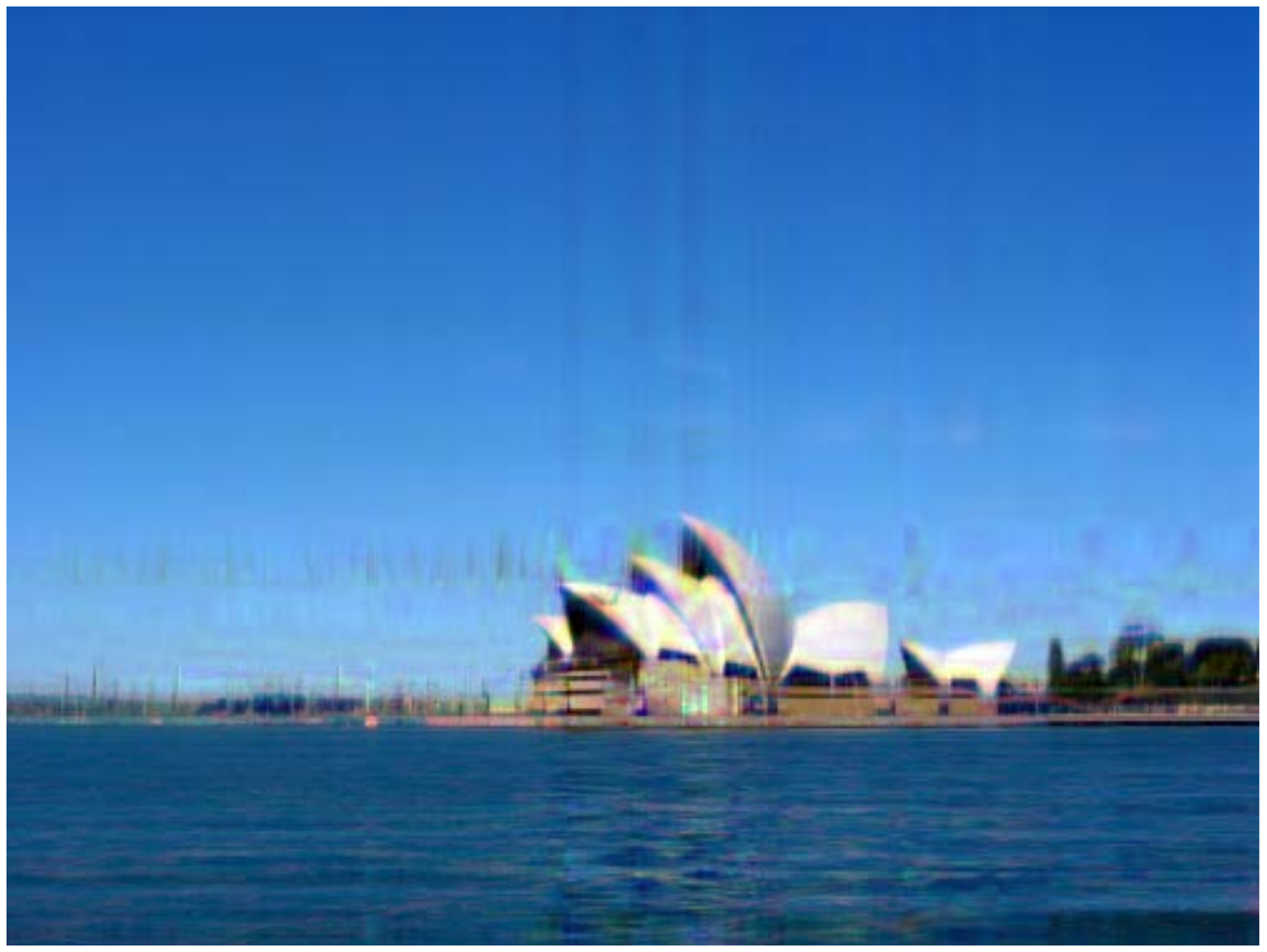}
        \caption{APGL}\label{fig_fig_imagerecovery_c}
	\end{subfigure}
	\begin{subfigure}[b]{0.138\textwidth}
		\centering
        \includegraphics[width=\textwidth]{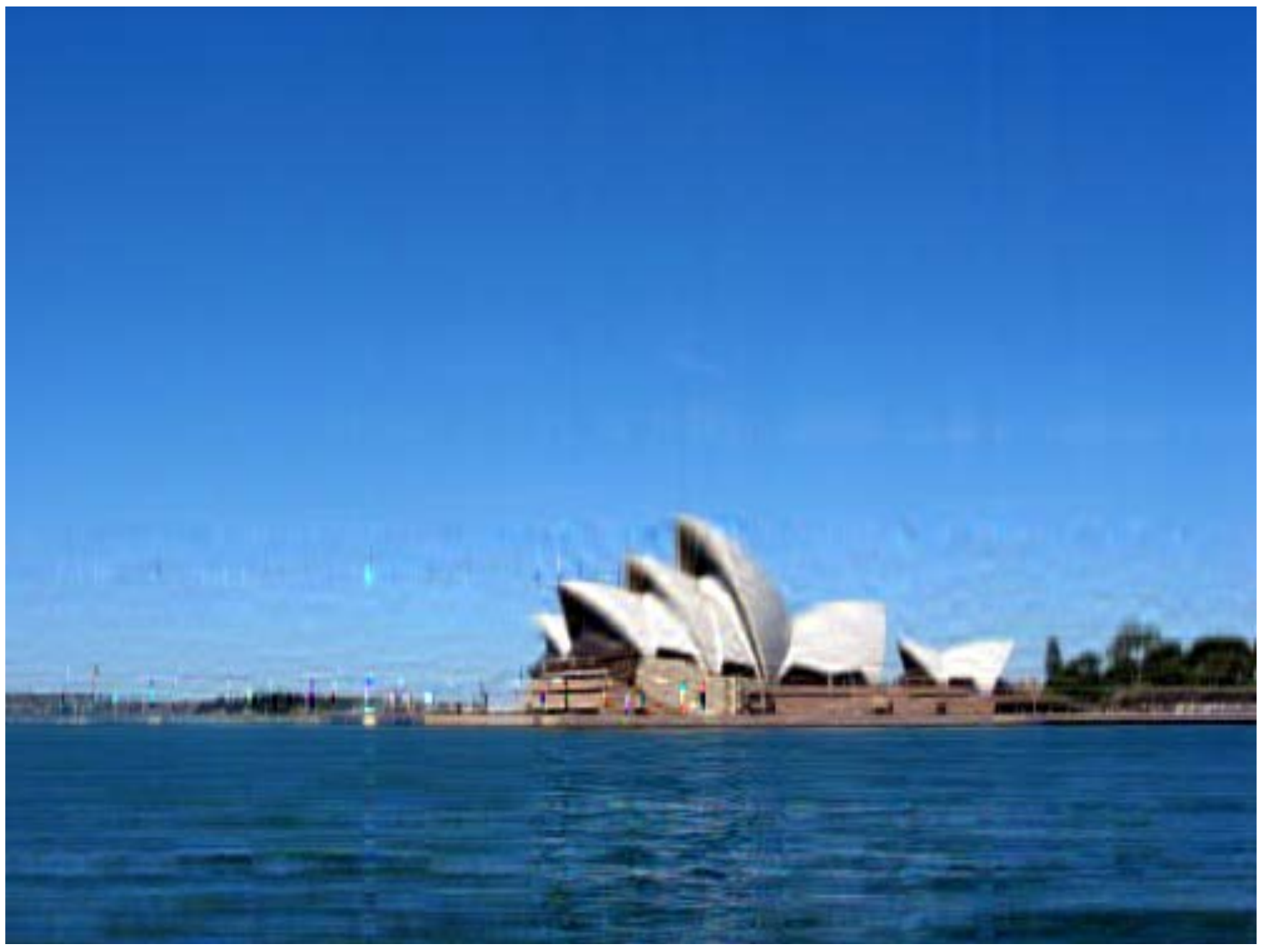}
        \caption{LMaFit}
	\end{subfigure}	
	\begin{subfigure}[b]{0.138\textwidth}
		\centering
        \includegraphics[width=\textwidth]{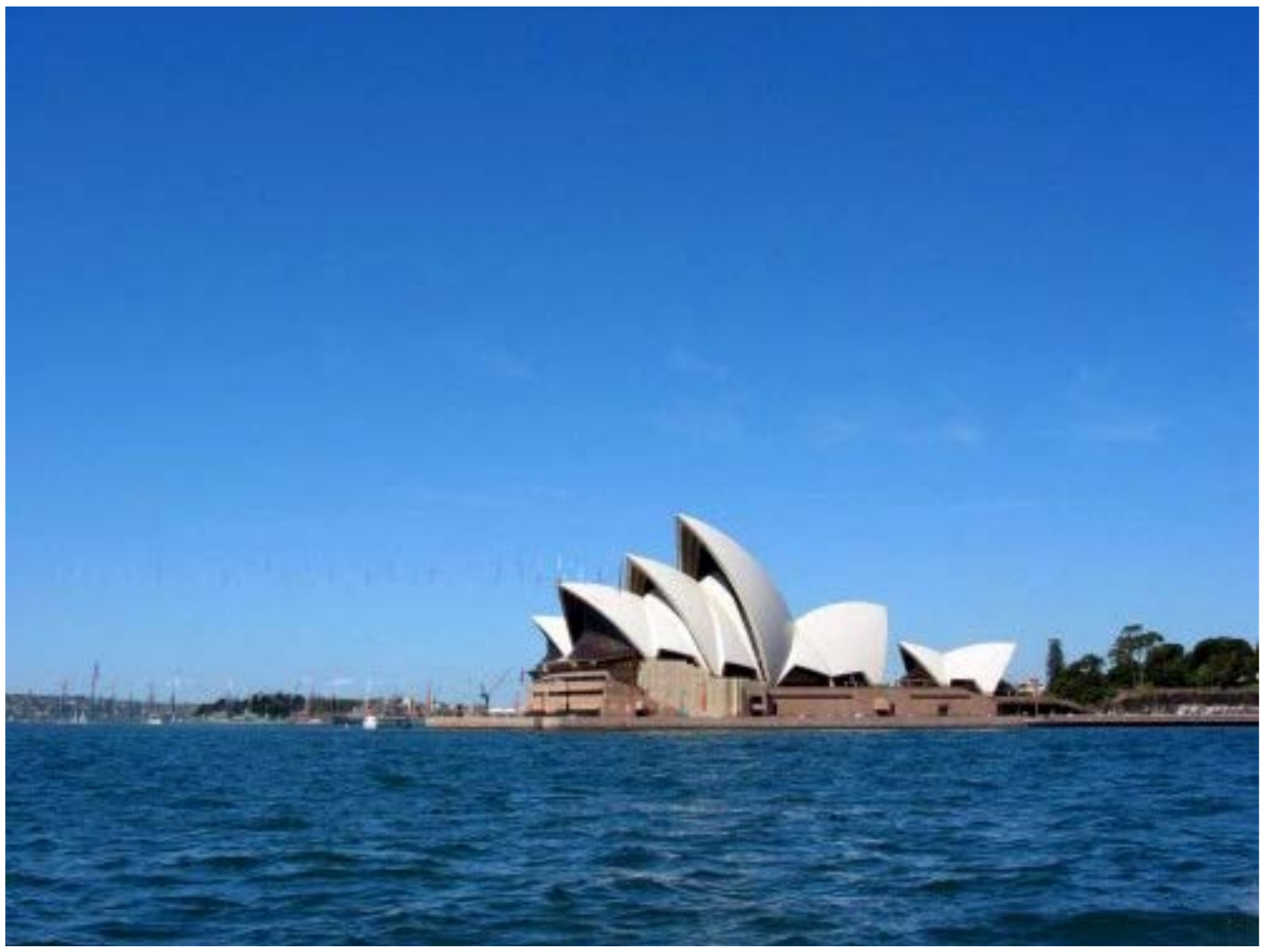}
        \caption{TNNR-ADMM}
	\end{subfigure}	
	\begin{subfigure}[b]{0.138\textwidth}
		\centering
        \includegraphics[width=\textwidth]{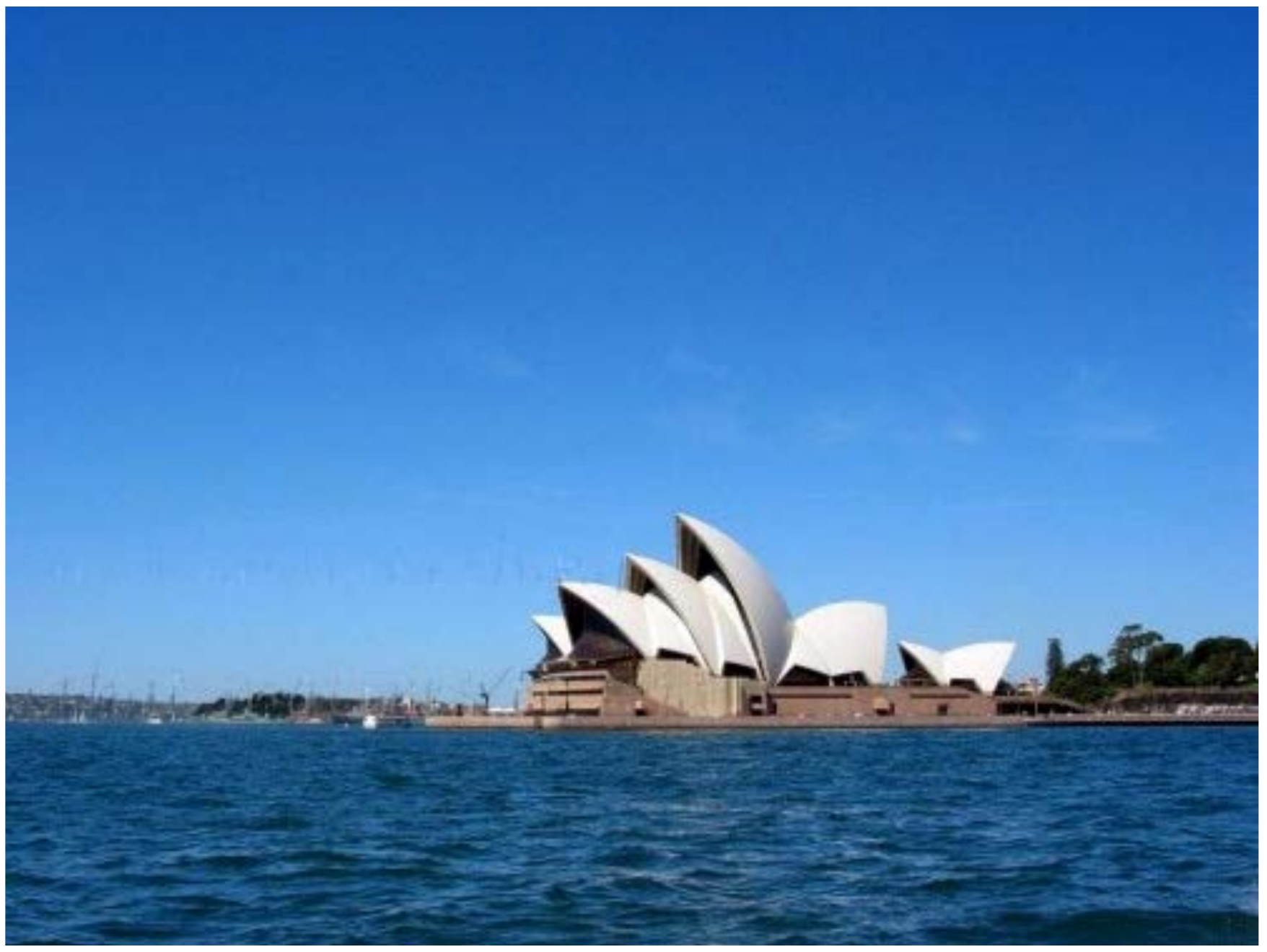}
        \caption{IRNN-$L_p$}
	\end{subfigure}
	\begin{subfigure}[b]{0.138\textwidth}
		\centering
        \includegraphics[width=\textwidth]{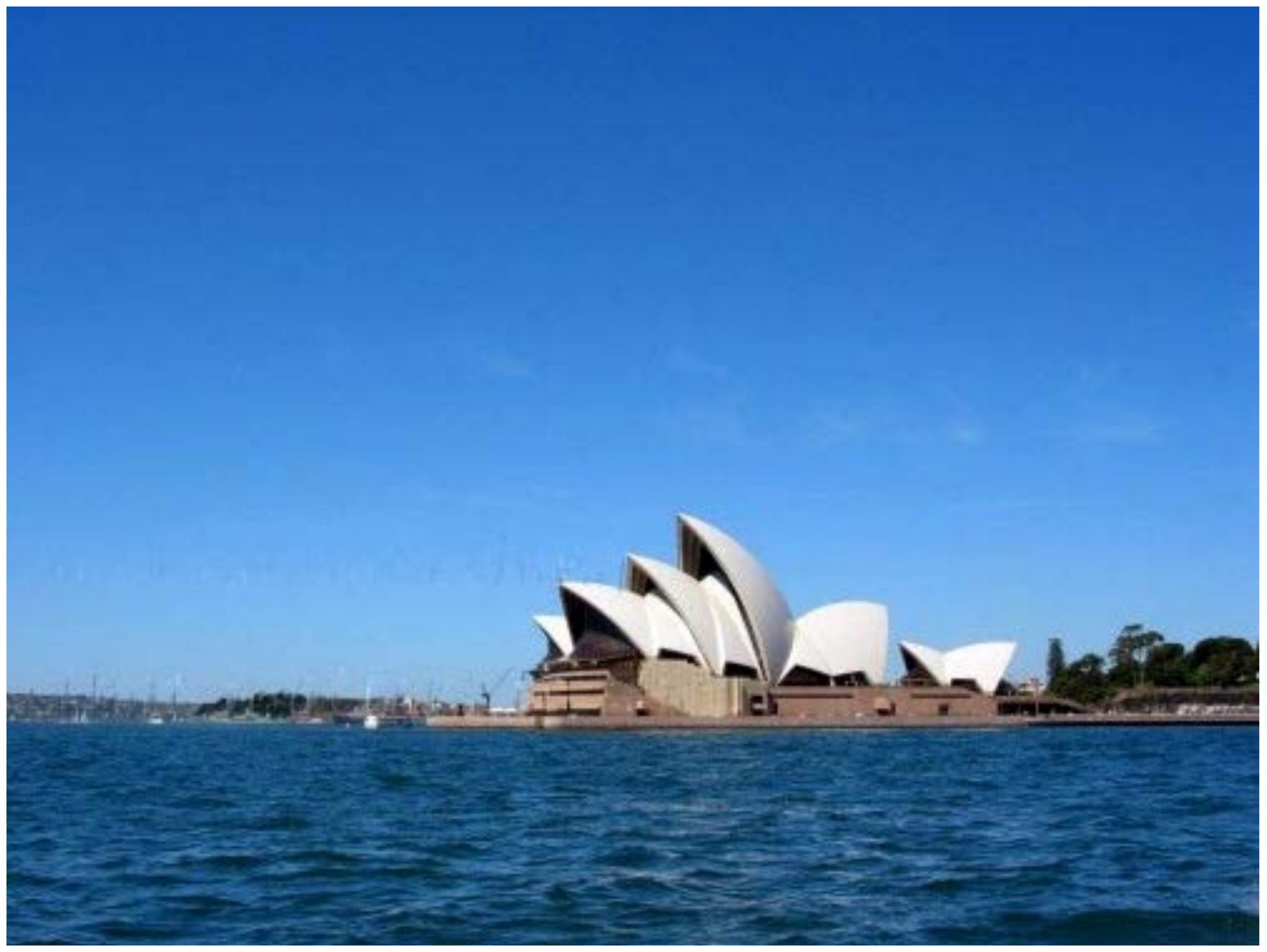}
        \caption{IRNN-SCAD}\label{fig_fig_imagerecovery_g}
	\end{subfigure} 	
	\caption{\small{Image recovery comparison by using different matrix completion algorithms. (a) Original image; (b) Image with Gaussian noise and text; (c)-(g) Recovered images by APGL, LMaFit, TNNR-ADMM, IRNN-$L_p$, and IRNN-SCAD, respectively. \textbf{Best viewed in $\times 2$ sized color pdf file.}}}\label{fig_imagerecovery}
	\vspace{-1.5em}
\end{figure*}

\section{Experiments}\label{sec_exp}

In this section, we present several experiments to demonstrate that the models with nonconvex rank surrogates outperform the ones with convex nuclear norm. We conduct three experiments. The first two aim to examine the convergence behavior of IRNN for the matrix completion problem \cite{candes2010matrix} on both synthetic data and real images. The last experiment is tested on the tensor low rank representation problem (\ref{tlrr}) solved by IRNN-PS for face clustering.  

For the first two experiments, we consider the nonconvex low rank matrix completion problem
\begin{equation}\label{pro_mc}
\min_{\X} \sum_{i=1}^mg(\sigma_i(\X))+\frac{1}{2}||\mathcal{P}_{\Omega}(\X-\M)||_F^2,
\end{equation}
where $\Omega$ is the set of indices of samples, and $\mathcal{P}_\Omega: \mathbb{R}^{m\times n}\rightarrow\mathbb{R}^{m\times n}$ is a linear operator that keeps the entries in $\Omega$ unchanged and those outside $\Omega$ zeros. The gradient of squared loss function in (\ref{pro_mc}) is Lipschitz continuous, with a Lipschitz constant $L(f)=1$. We set $\mu=1.1$ in IRNN. For the choice of $g$, we use five nonconvex surrogates in Table \ref{tab_nonpenlty}, including $L_p$-norm, SCAD, Logarithm, MCP and ETP. The other three nonconvex surrogates, including Capped $L_1$, Geman and Laplace, are not used since we find that their recovery performances are very sensitive to the choices of $\gamma$ and $\lambda$ in different cases. For the choice of $\lambda$ in $g$, we use a continuation technique to enhance the low rank matrix recovery. The initial value of $\lambda$ is set to a larger value $\lambda_0$, and dynamically decreased by $\lambda=\eta^k\lambda_0$ with $\eta<1$. It is stopped till reaching a predefined target $\lambda_t$. $\X$ is initialized as a zero matrix. For the choice of parameters (e.g., $p$ and $\gamma$) in $g$, we search them from a candidate set and use the one which obtains good performance in most cases.

\subsection{Low Rank Matrix Recovery on the Synthetic Data}
We first compare the low rank matrix recovery performances of nonconvex model (\ref{pro_mc}) with the convex one by using nuclear norm \cite{candes2009exact} on the synthetic data. We conduct two tasks. The first one is tested on the observed matrix $\M$ without noises, while the other one is tested on $\M$ with noises.

\begin{figure}
	\begin{subfigure}[b]{0.118\textwidth}
		\centering
		\includegraphics[width=\textwidth]{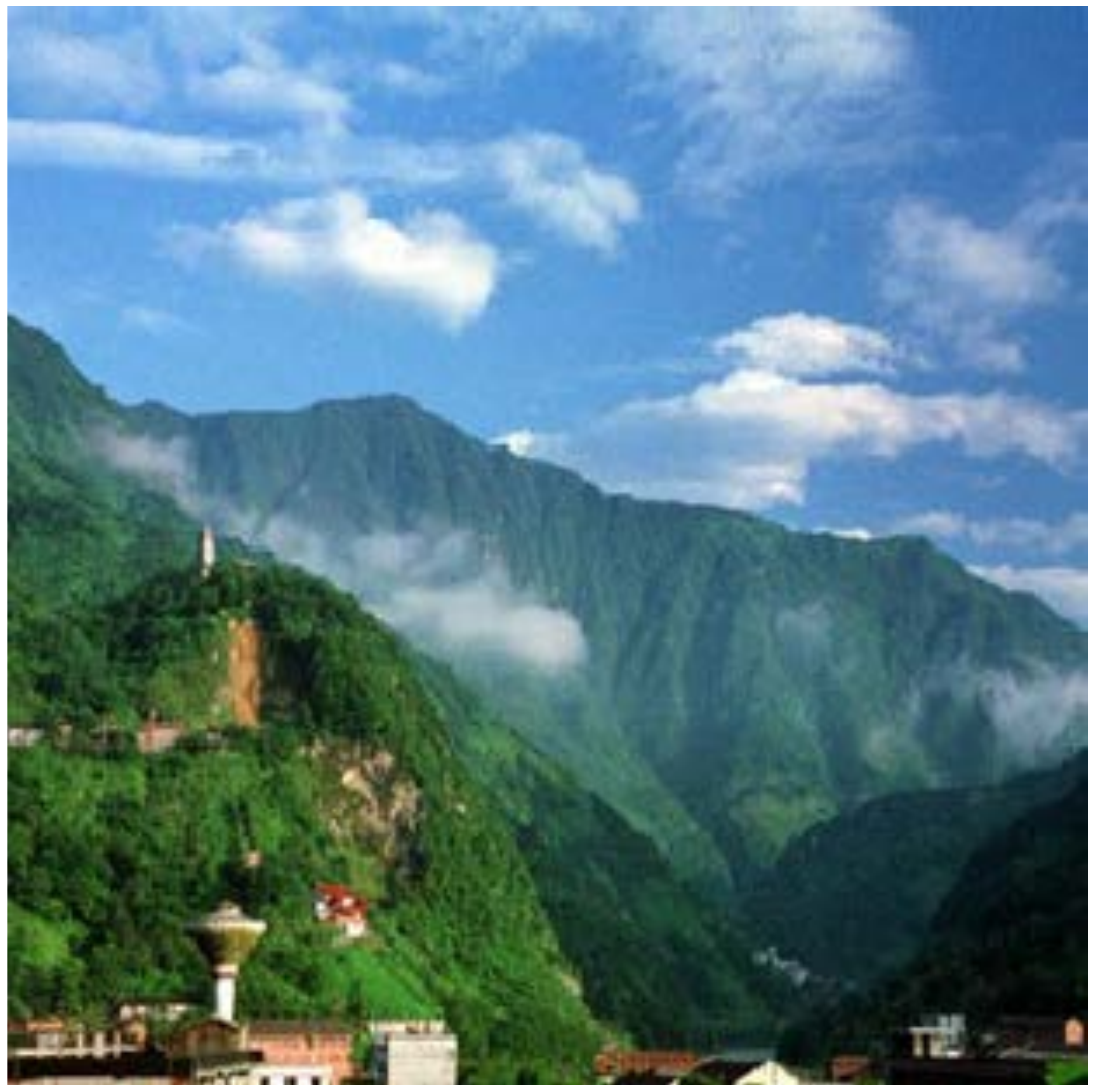}
	\end{subfigure}
	\begin{subfigure}[b]{0.118\textwidth}
		\centering
		\includegraphics[width=\textwidth]{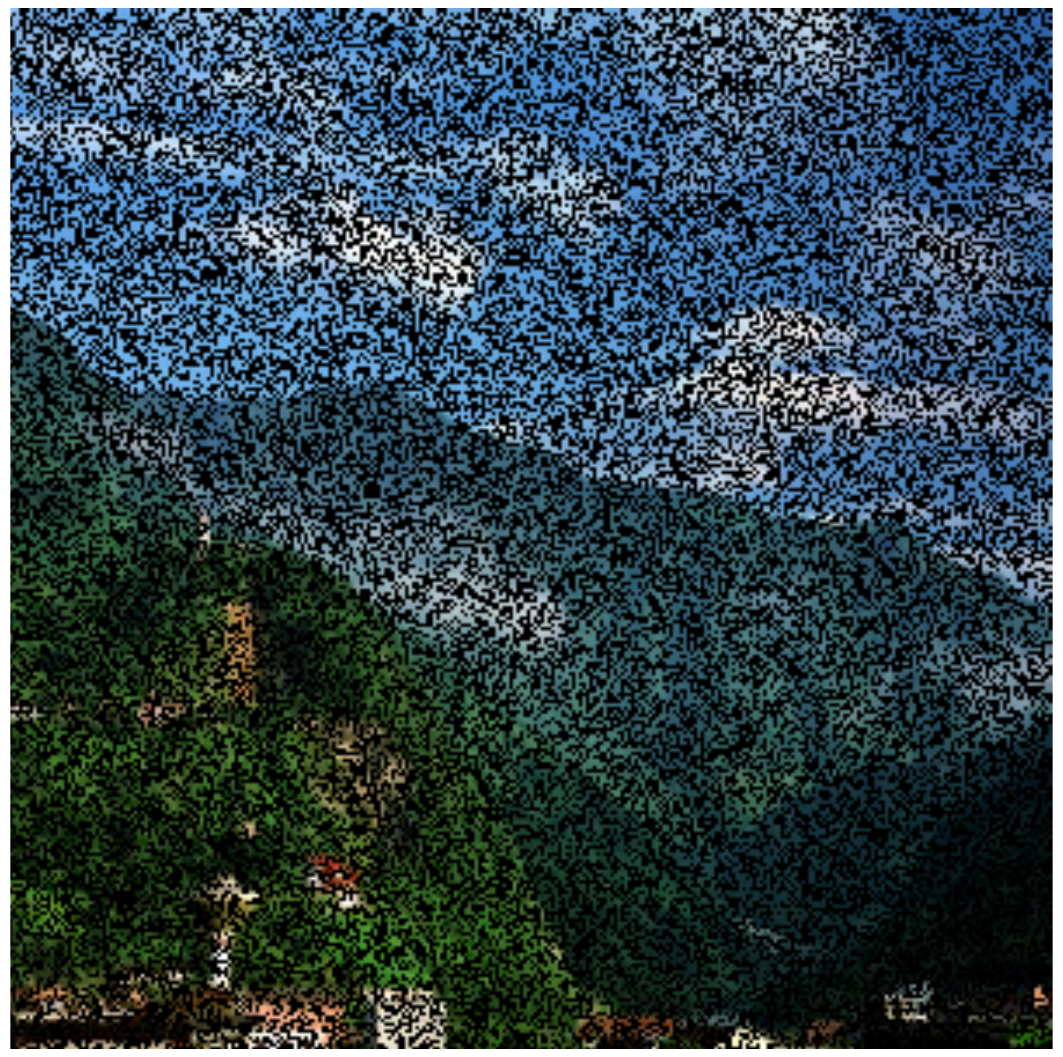}
	\end{subfigure}
	\begin{subfigure}[b]{0.118\textwidth}
		\centering
		\includegraphics[width=\textwidth]{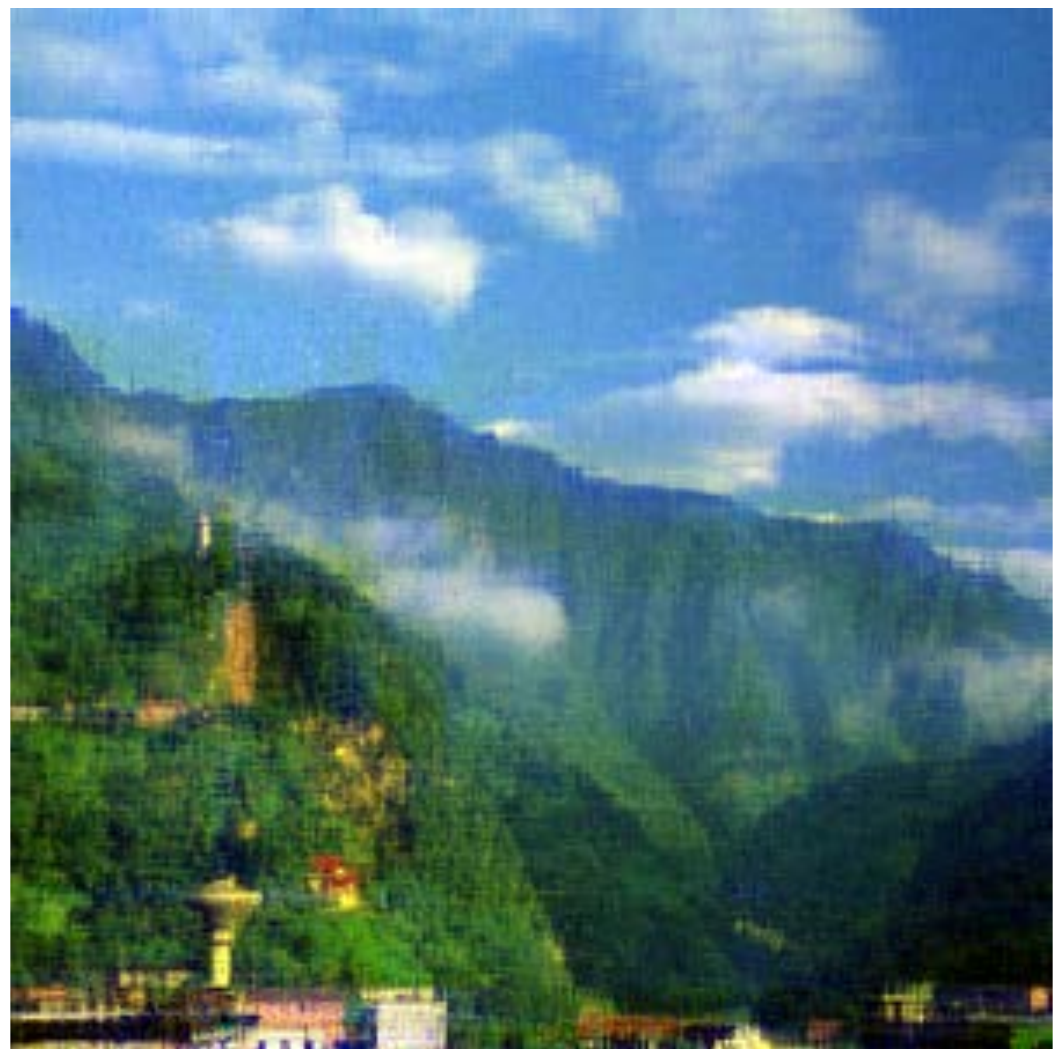}
	\end{subfigure}
	\begin{subfigure}[b]{0.118\textwidth}
		\centering
		\includegraphics[width=\textwidth]{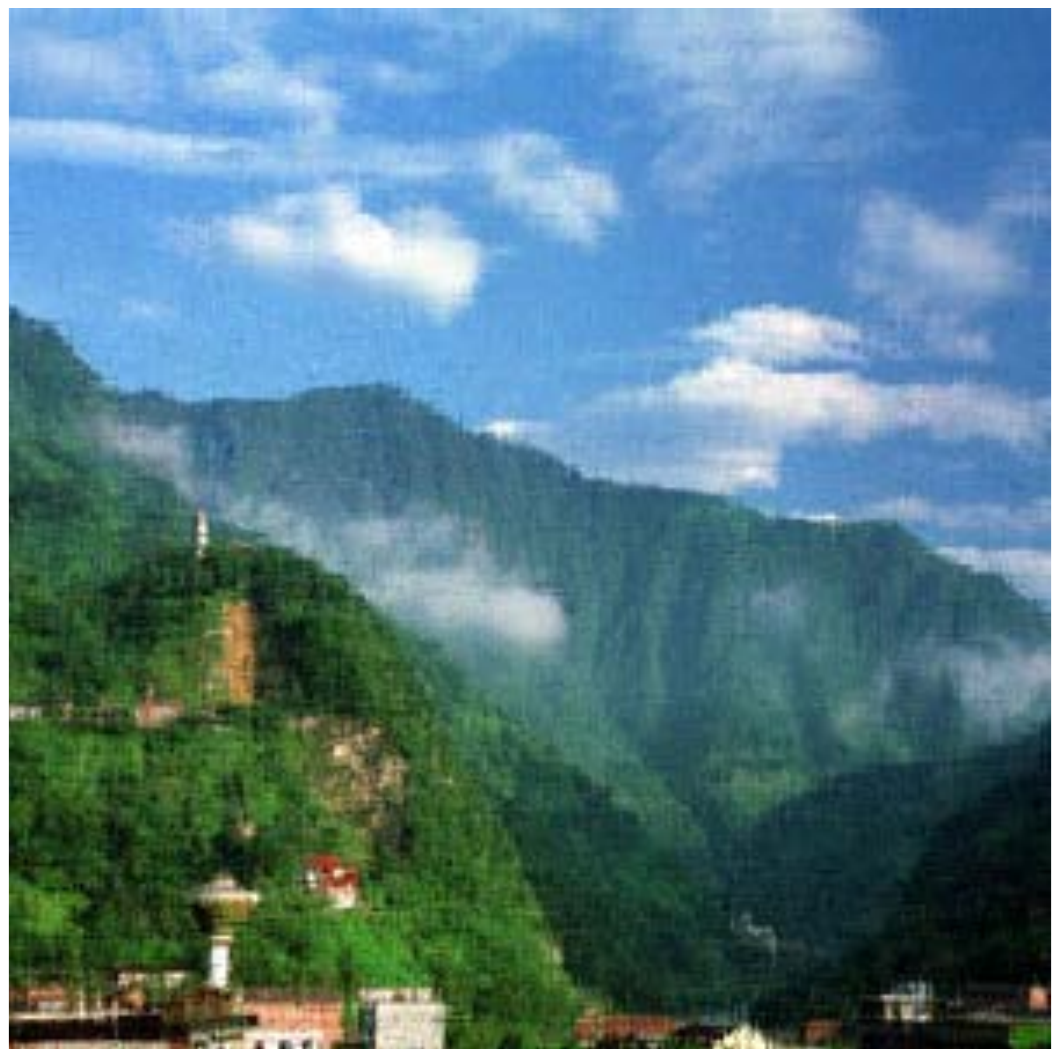}
	\end{subfigure}
	
	\begin{subfigure}[b]{0.118\textwidth}
		\centering
		\includegraphics[width=\textwidth]{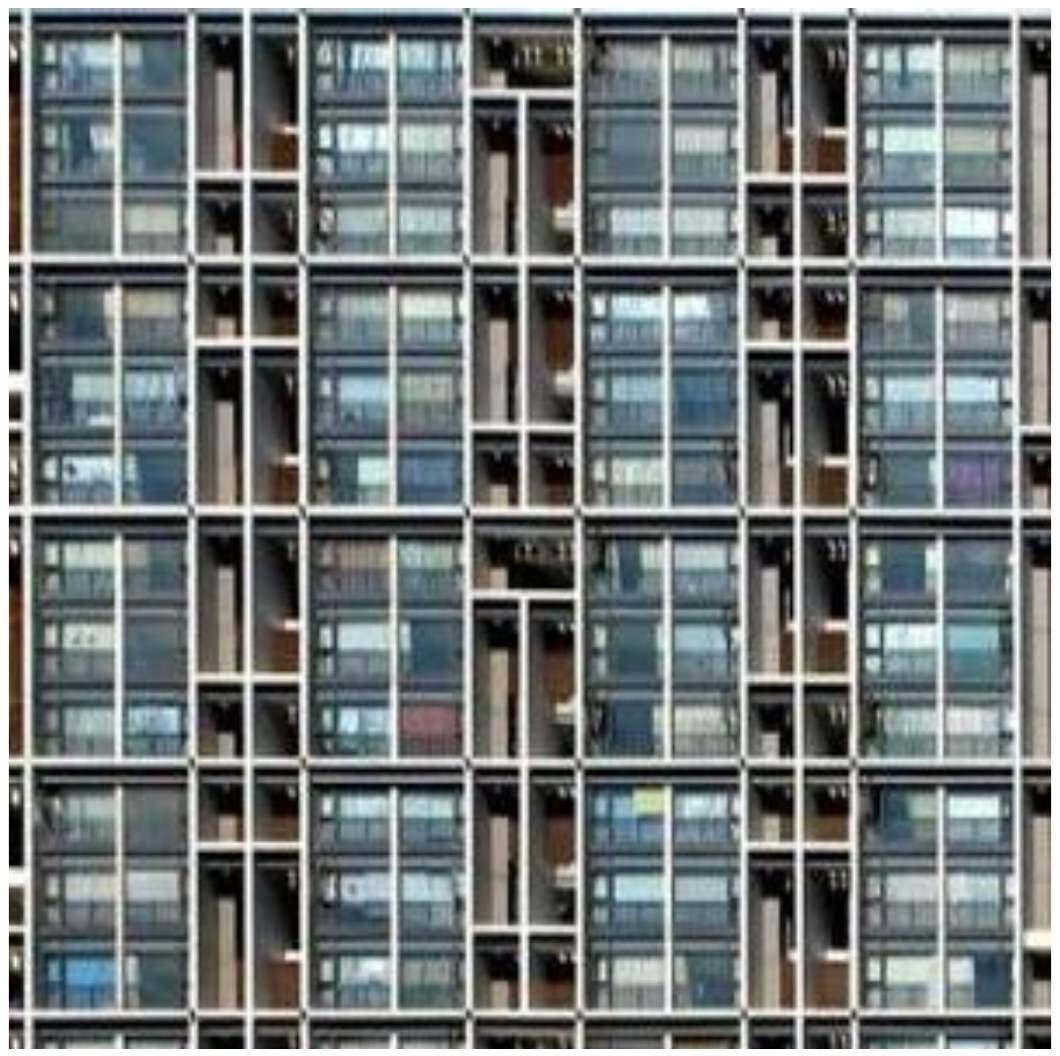}
	\end{subfigure}
	\begin{subfigure}[b]{0.118\textwidth}
		\centering
		\includegraphics[width=\textwidth]{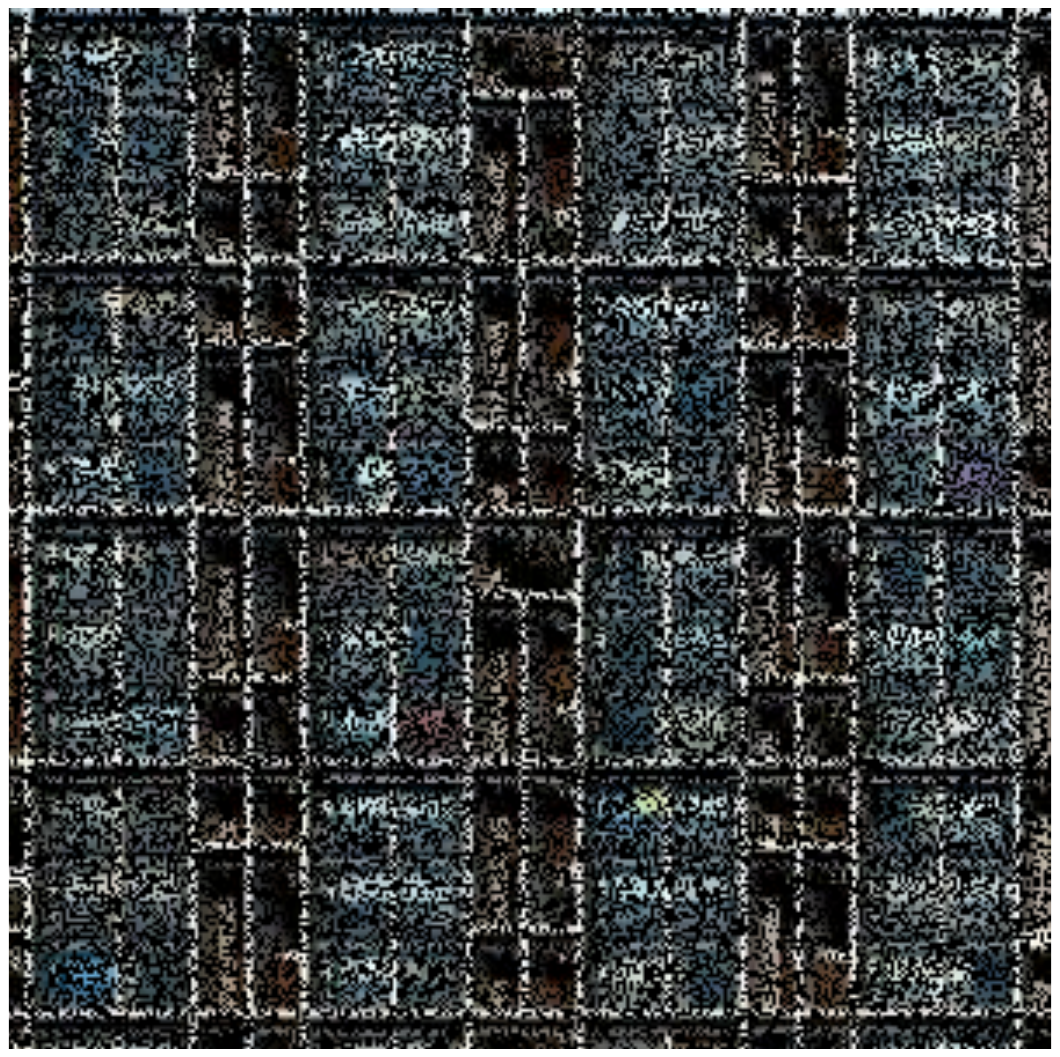}
	\end{subfigure}	
	\begin{subfigure}[b]{0.118\textwidth}
		\centering
		\includegraphics[width=\textwidth]{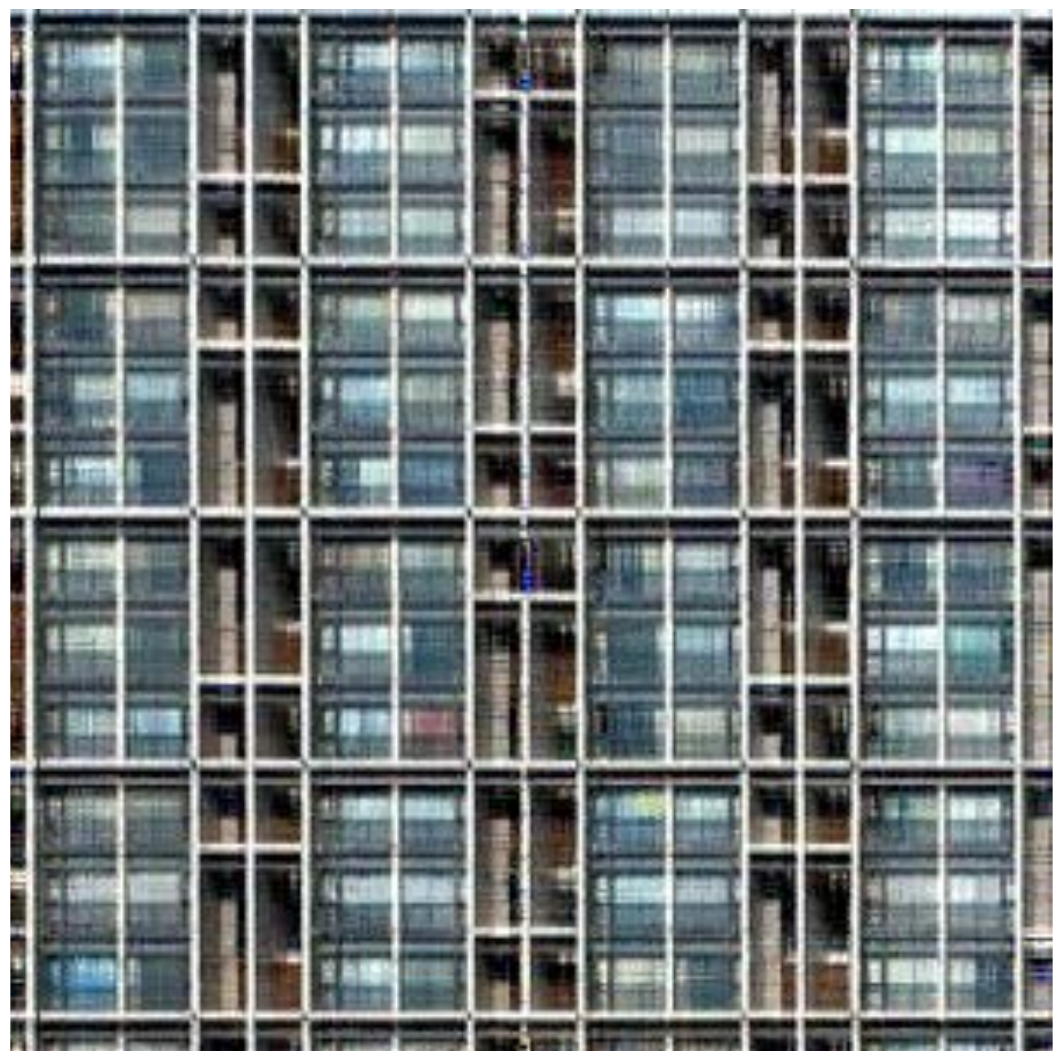}
	\end{subfigure}	
	\begin{subfigure}[b]{0.118\textwidth}
		\centering
		\includegraphics[width=\textwidth]{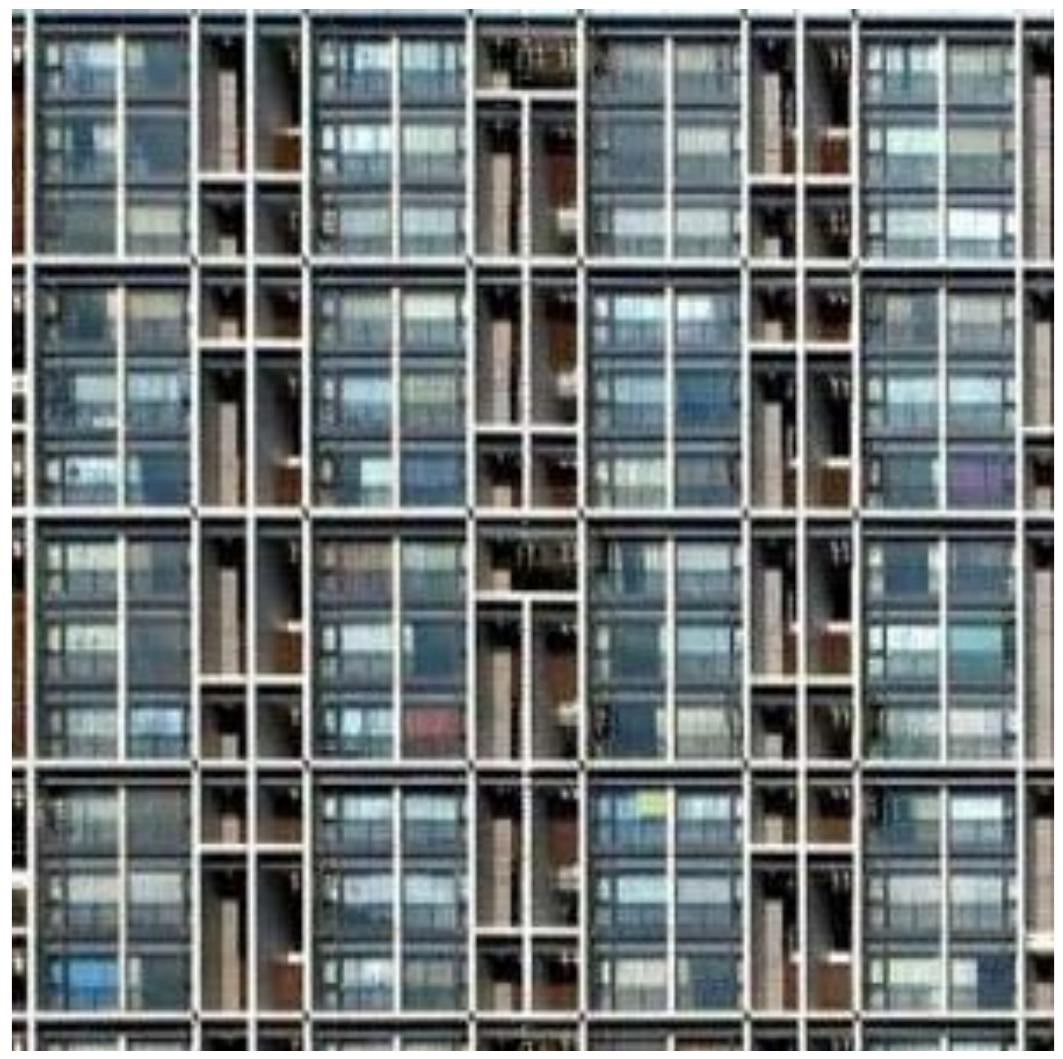}
	\end{subfigure}	
	
	\begin{subfigure}[b]{0.118\textwidth}
		\centering
		\includegraphics[width=\textwidth]{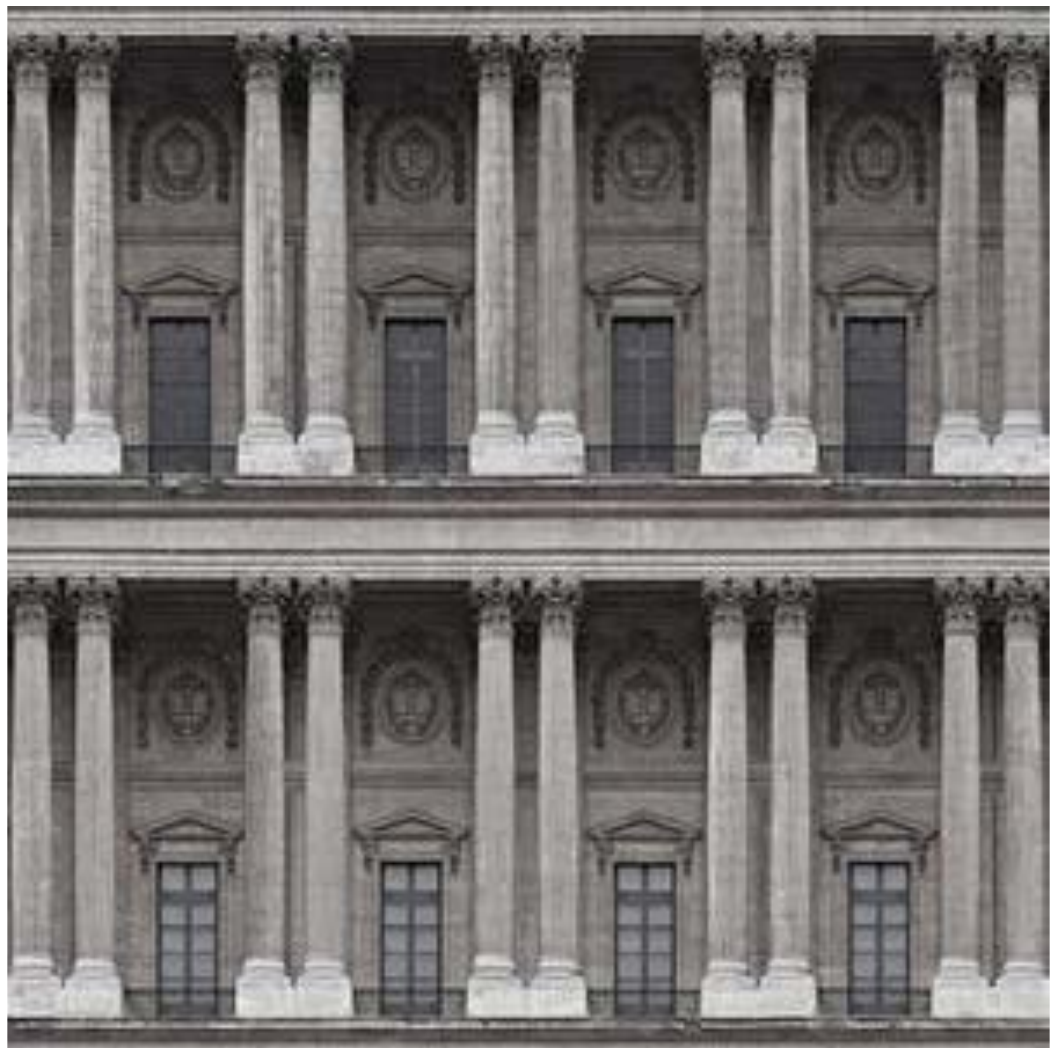}
	\end{subfigure}
	\begin{subfigure}[b]{0.118\textwidth}
		\centering
		\includegraphics[width=\textwidth]{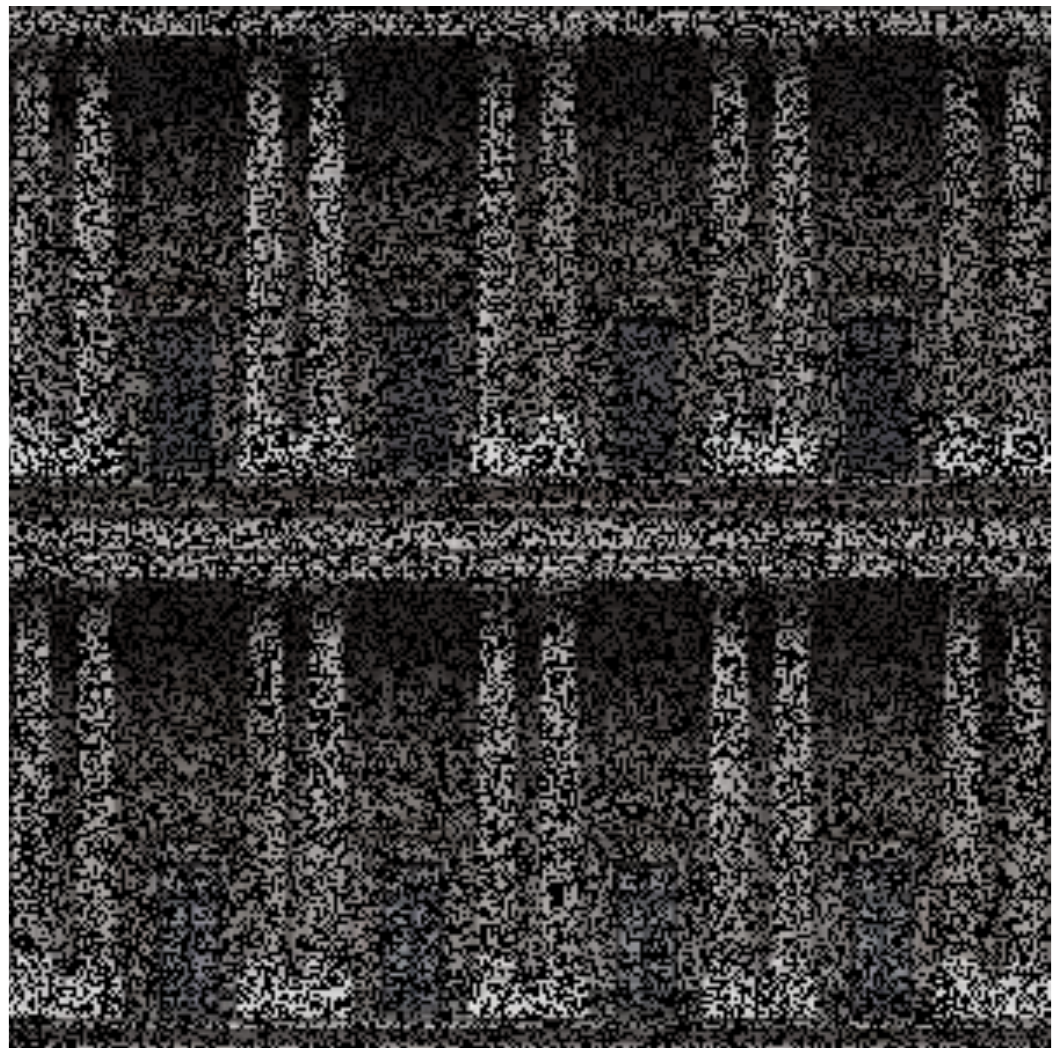}
	\end{subfigure}	
	\begin{subfigure}[b]{0.118\textwidth}
		\centering
		\includegraphics[width=\textwidth]{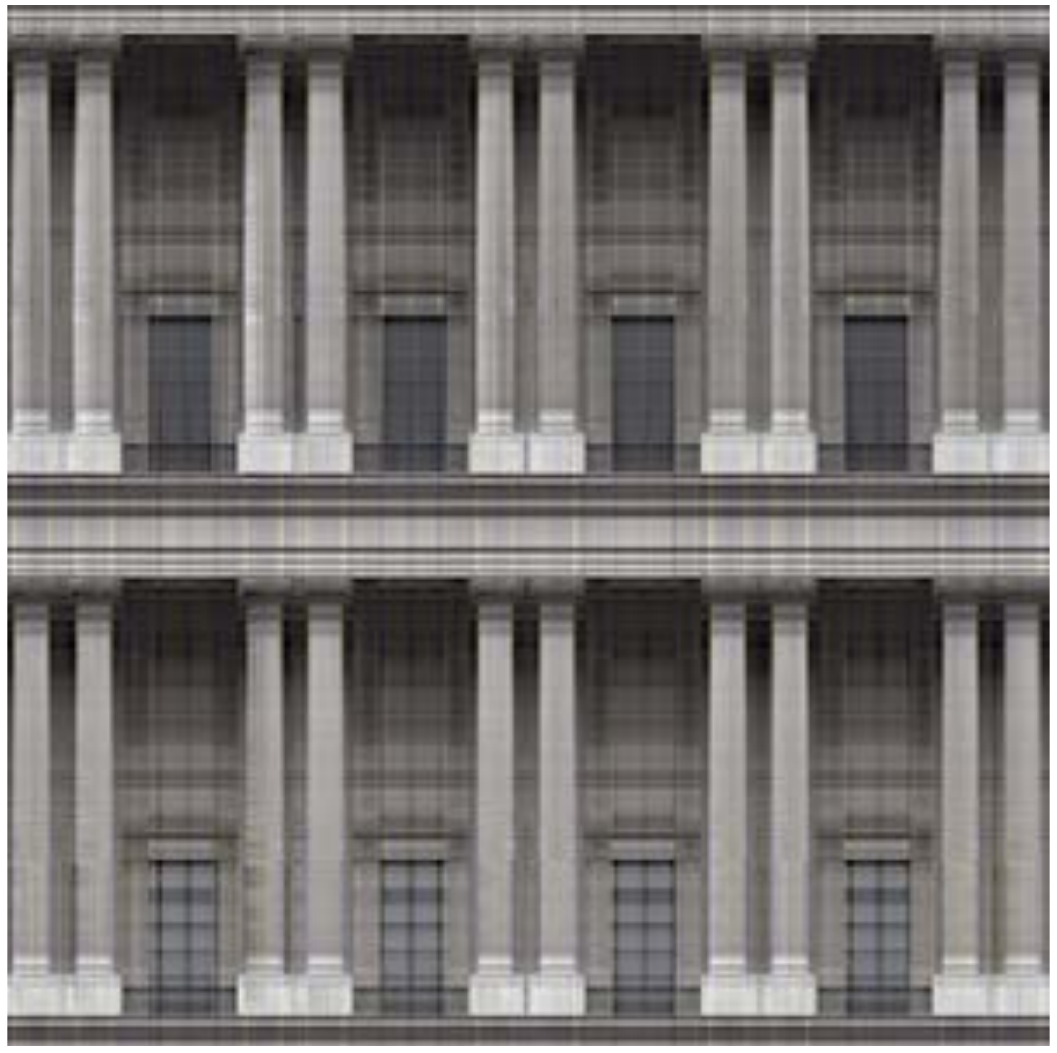}
	\end{subfigure}	
	\begin{subfigure}[b]{0.118\textwidth}
		\centering
		\includegraphics[width=\textwidth]{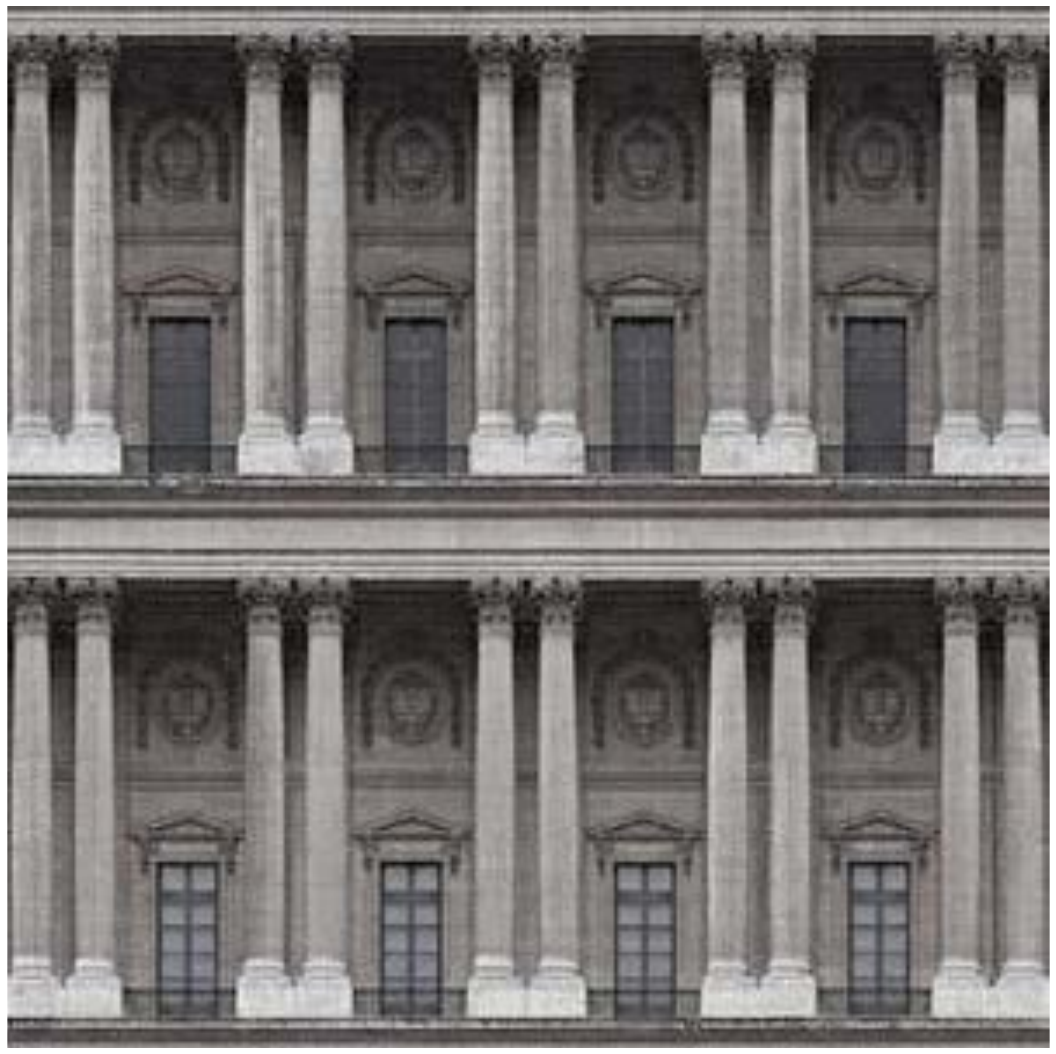}
	\end{subfigure}

	\begin{subfigure}[b]{0.118\textwidth}
		\centering
		\includegraphics[width=\textwidth]{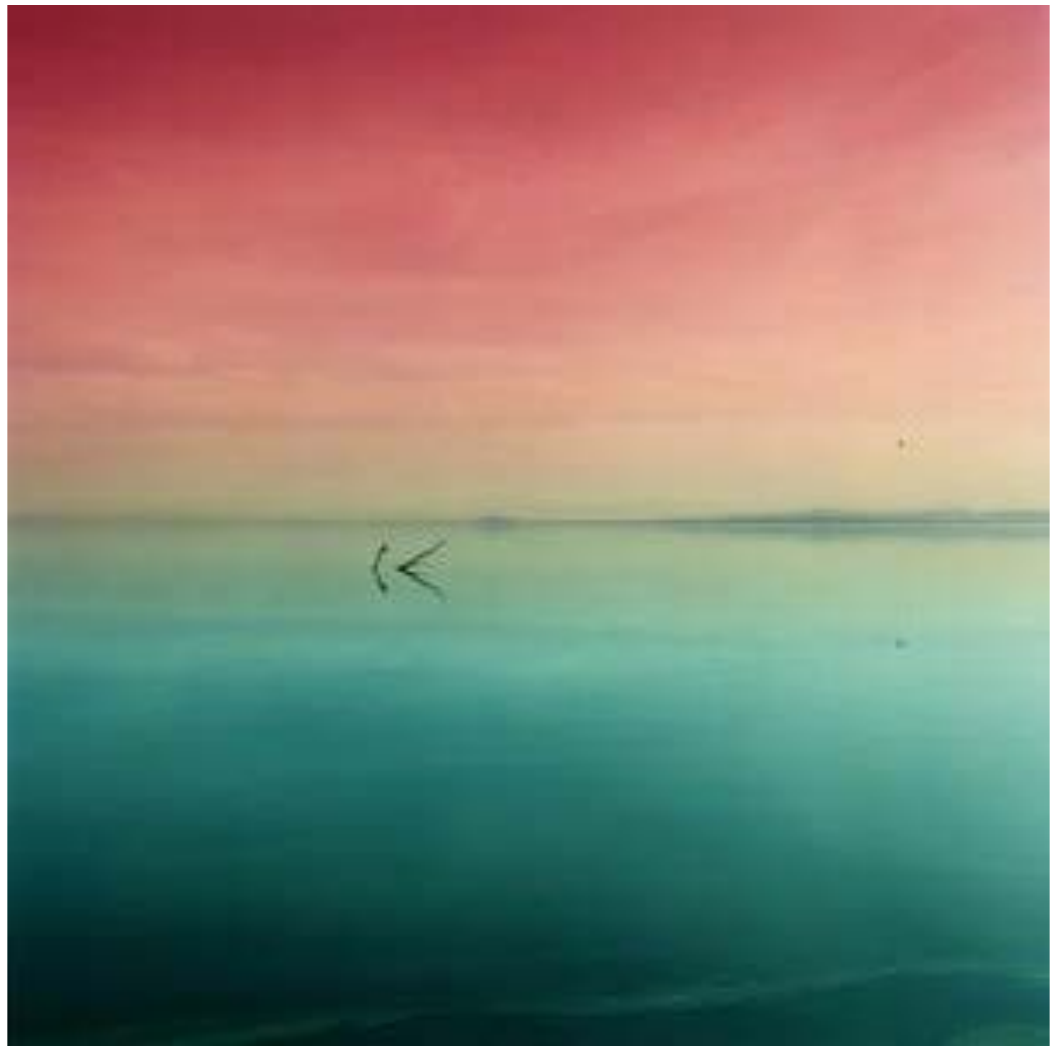}
	\end{subfigure}
	\begin{subfigure}[b]{0.118\textwidth}
		\centering
		\includegraphics[width=\textwidth]{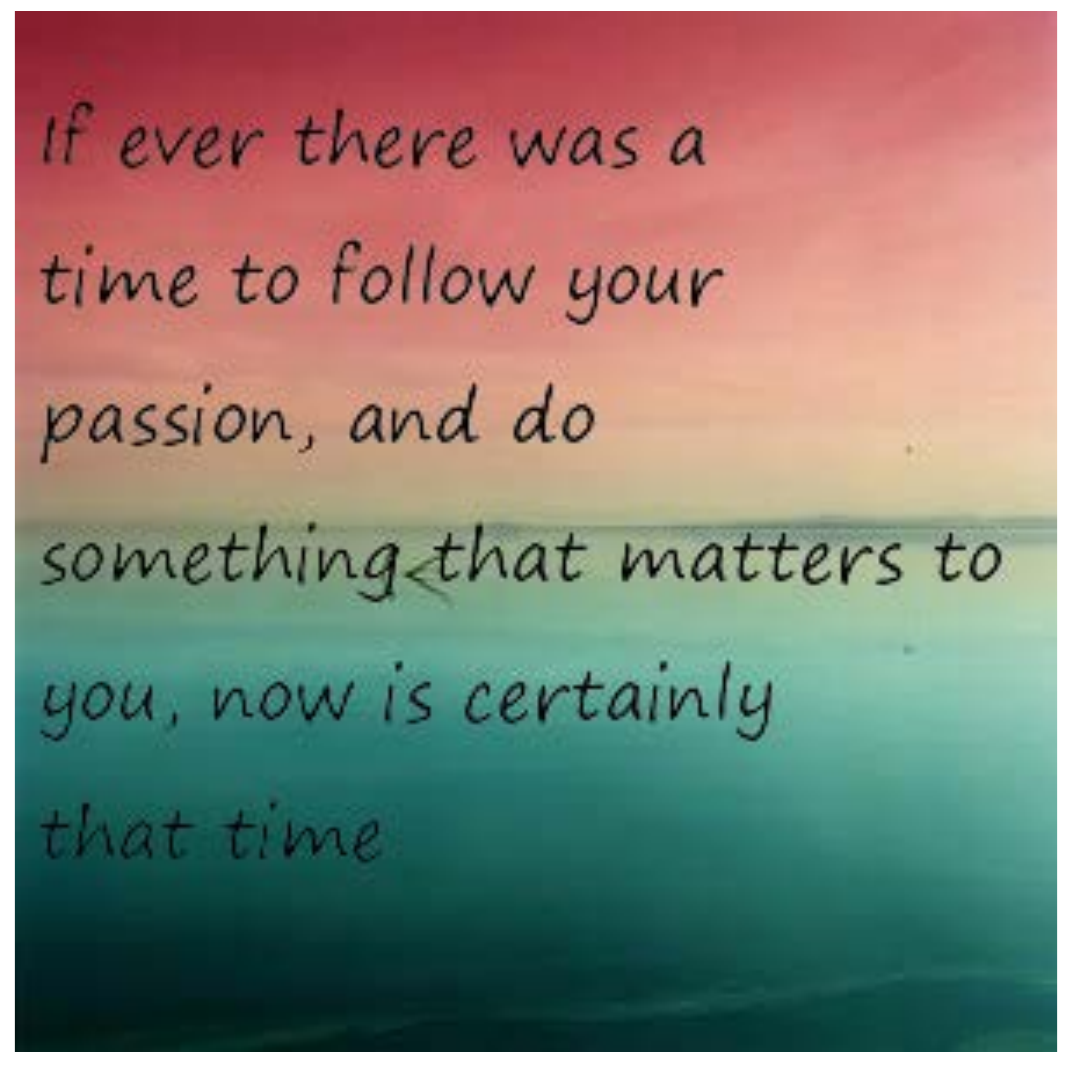}
	\end{subfigure}	
	\begin{subfigure}[b]{0.118\textwidth}
		\centering
		\includegraphics[width=\textwidth]{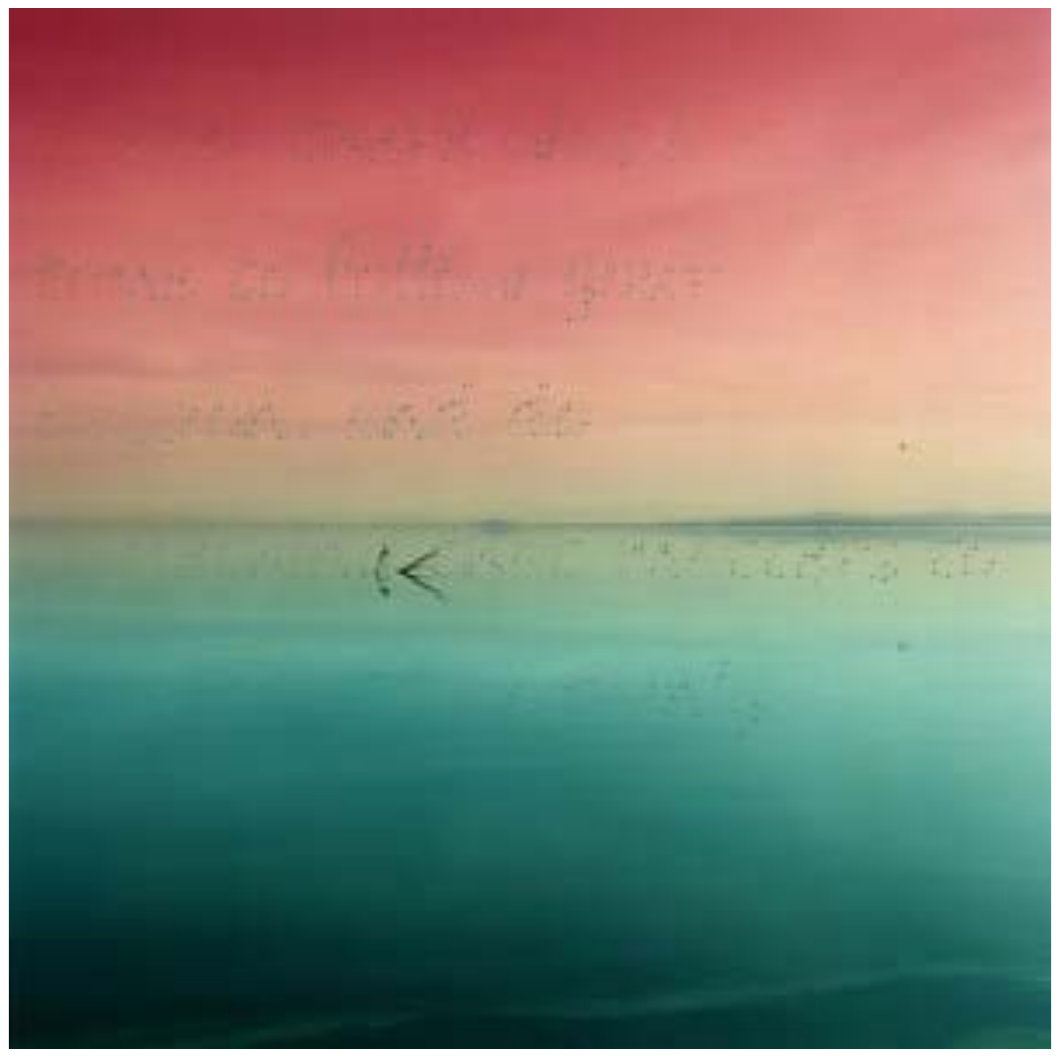}
	\end{subfigure}	
	\begin{subfigure}[b]{0.118\textwidth}
		\centering
		\includegraphics[width=\textwidth]{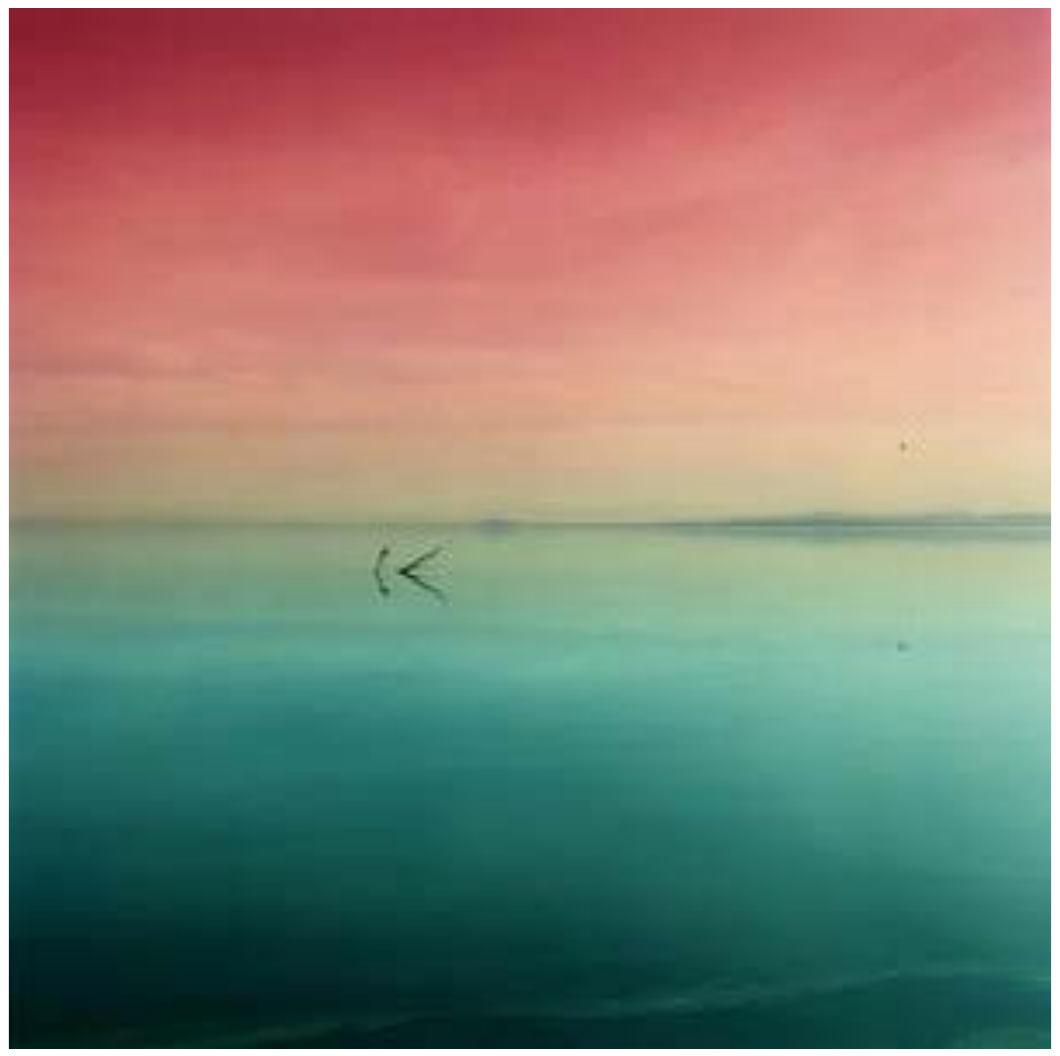}
	\end{subfigure}

	\begin{subfigure}[b]{0.118\textwidth}
		\centering
		\includegraphics[width=\textwidth]{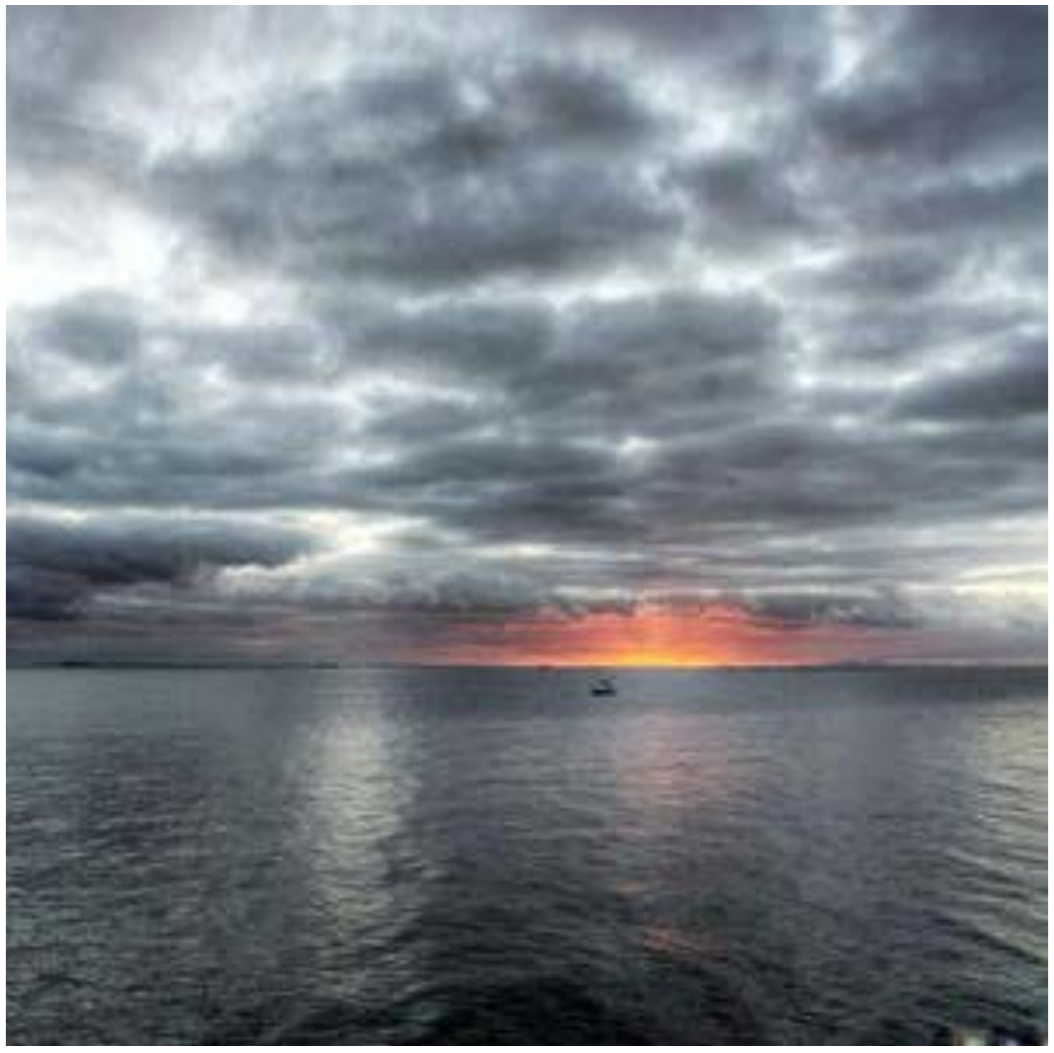}
	\end{subfigure}
	\begin{subfigure}[b]{0.118\textwidth}
		\centering
		\includegraphics[width=\textwidth]{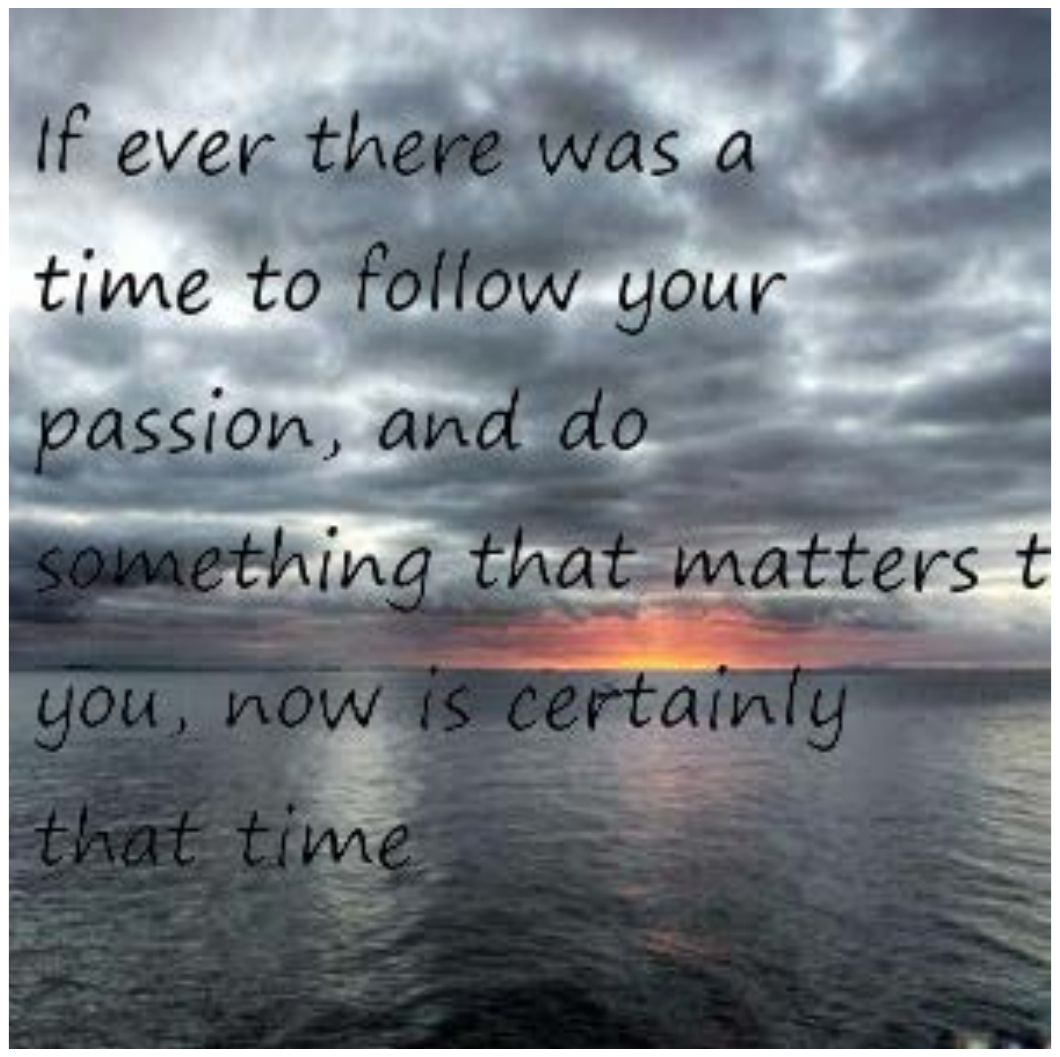}
	\end{subfigure}
	\begin{subfigure}[b]{0.118\textwidth}
		\centering
		\includegraphics[width=\textwidth]{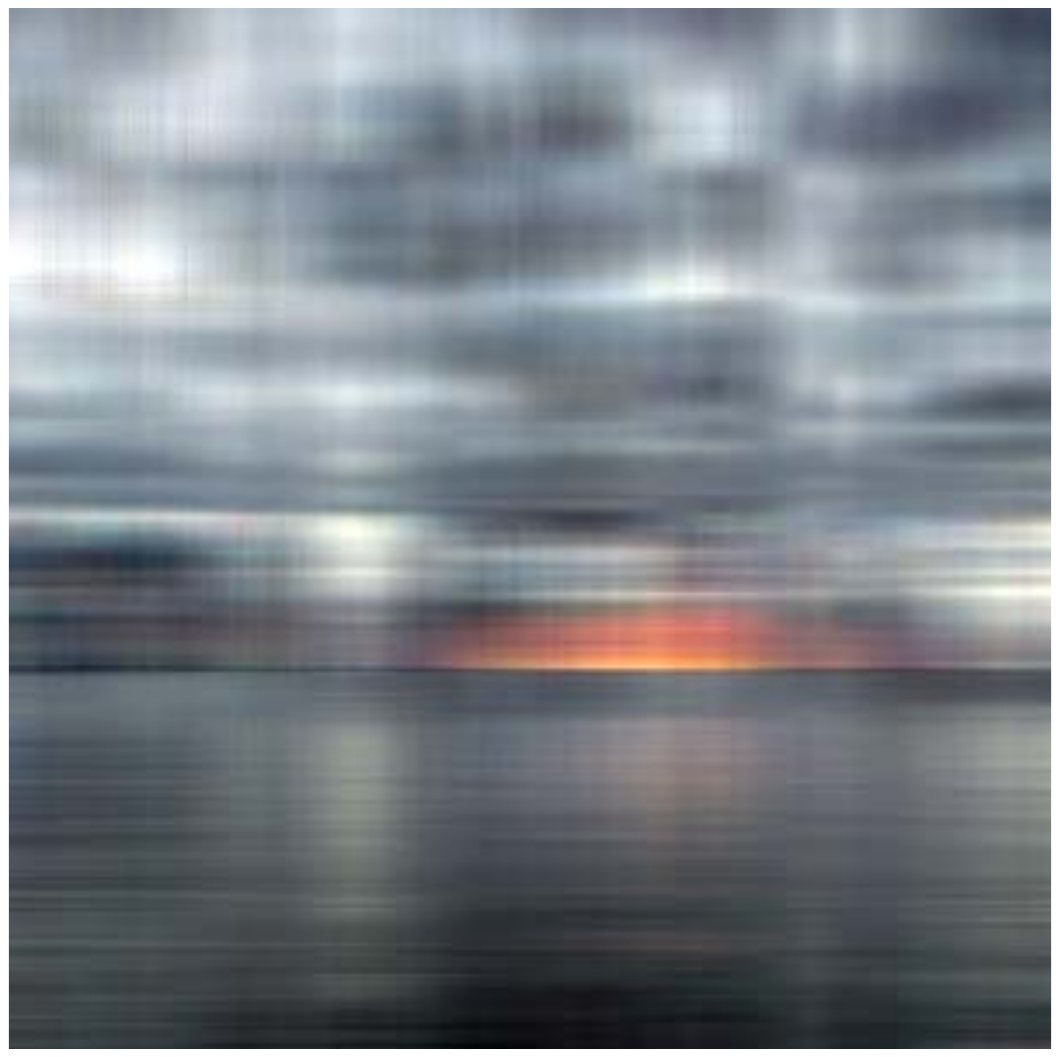}
	\end{subfigure}
	\begin{subfigure}[b]{0.118\textwidth}
		\centering
		\includegraphics[width=\textwidth]{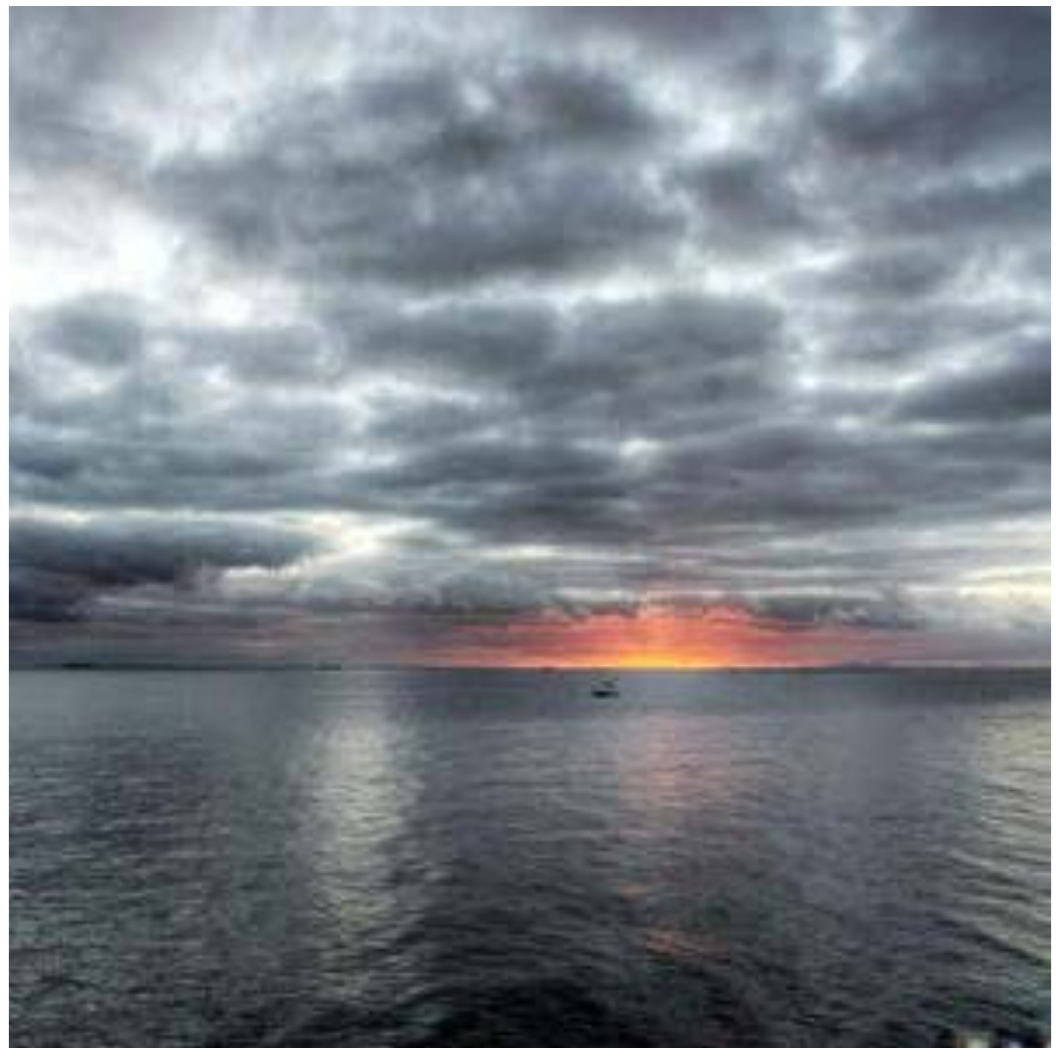}
	\end{subfigure}	
	
	\begin{subfigure}[b]{0.118\textwidth}
		\centering
		\includegraphics[width=\textwidth]{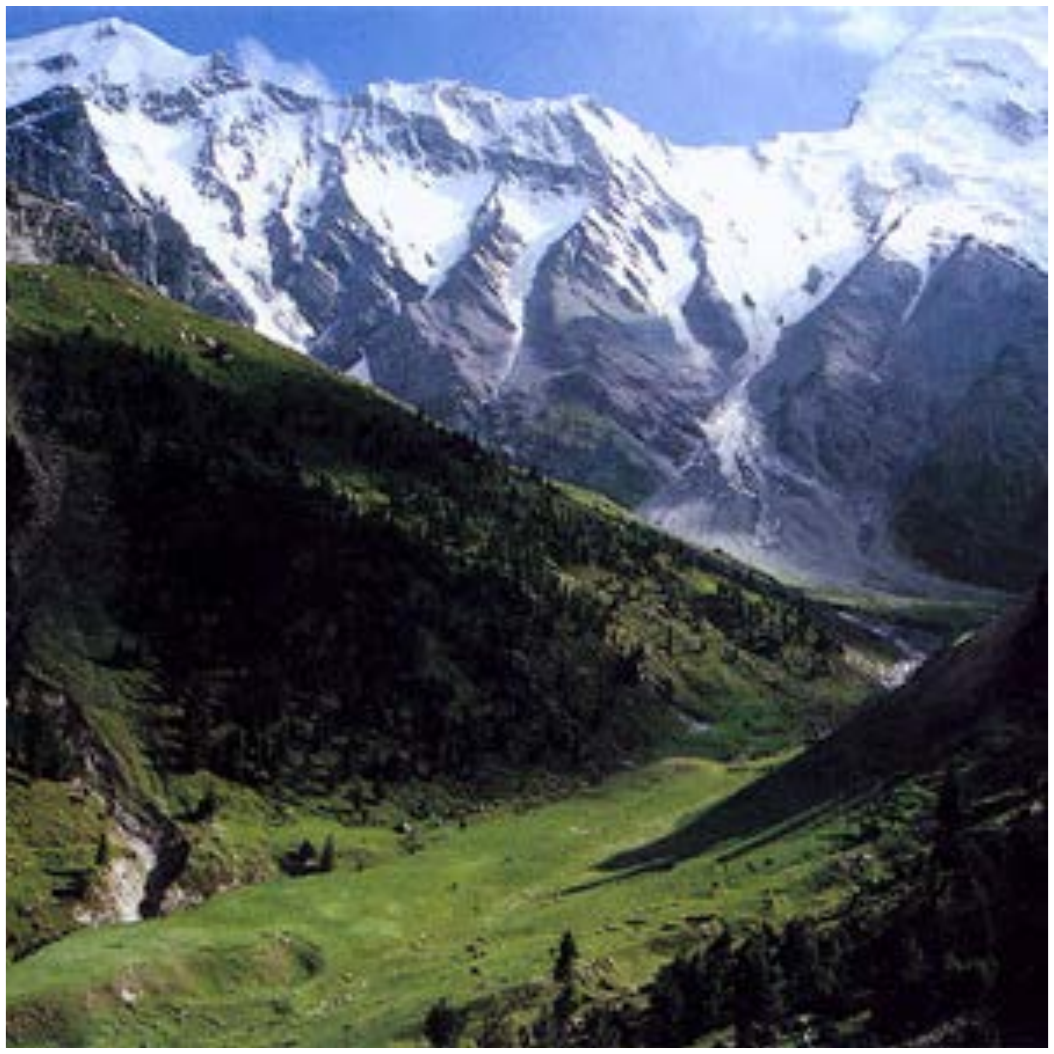}
		\caption{Original}
	\end{subfigure}
	\begin{subfigure}[b]{0.118\textwidth}
		\centering
		\includegraphics[width=\textwidth]{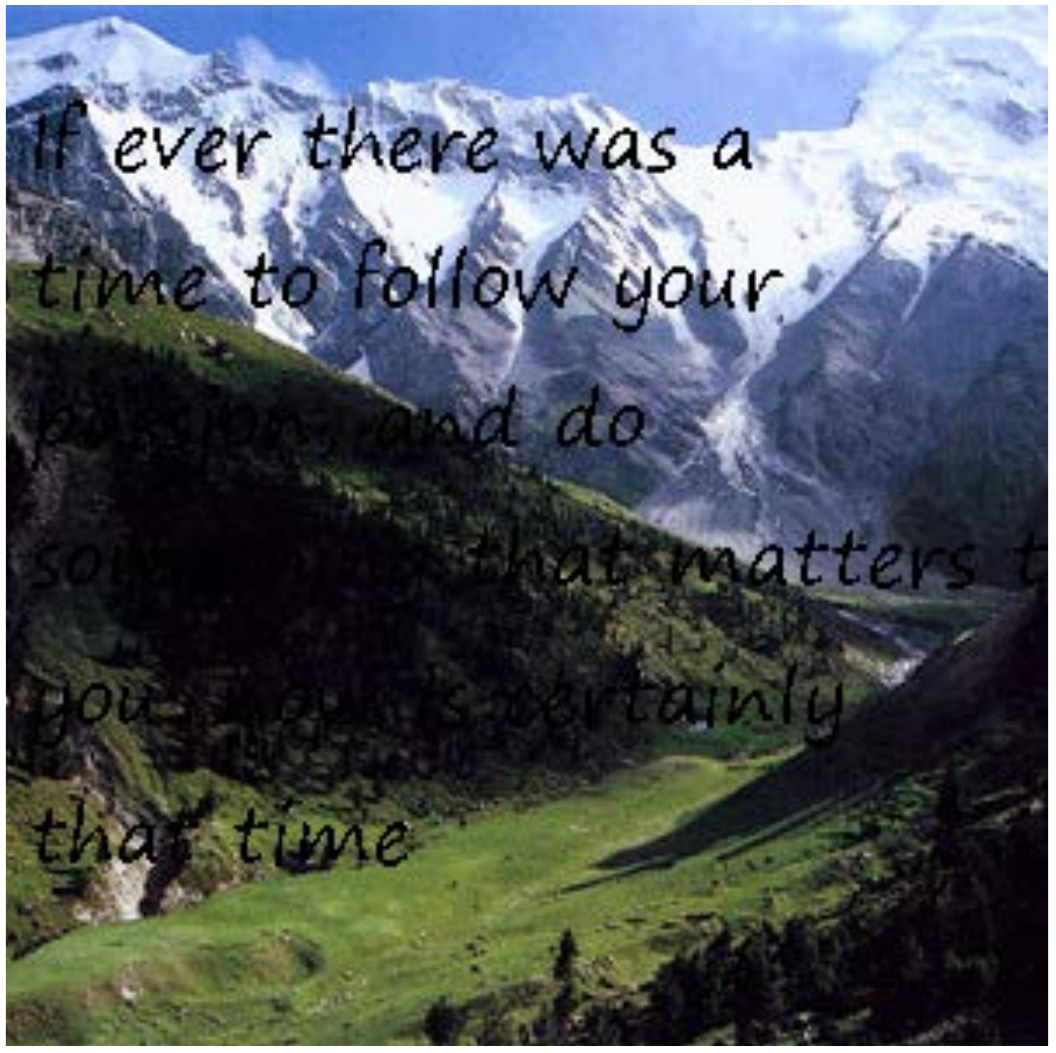}
		\caption{Noisy image}
	\end{subfigure}
	\begin{subfigure}[b]{0.118\textwidth}
		\centering
		\includegraphics[width=\textwidth]{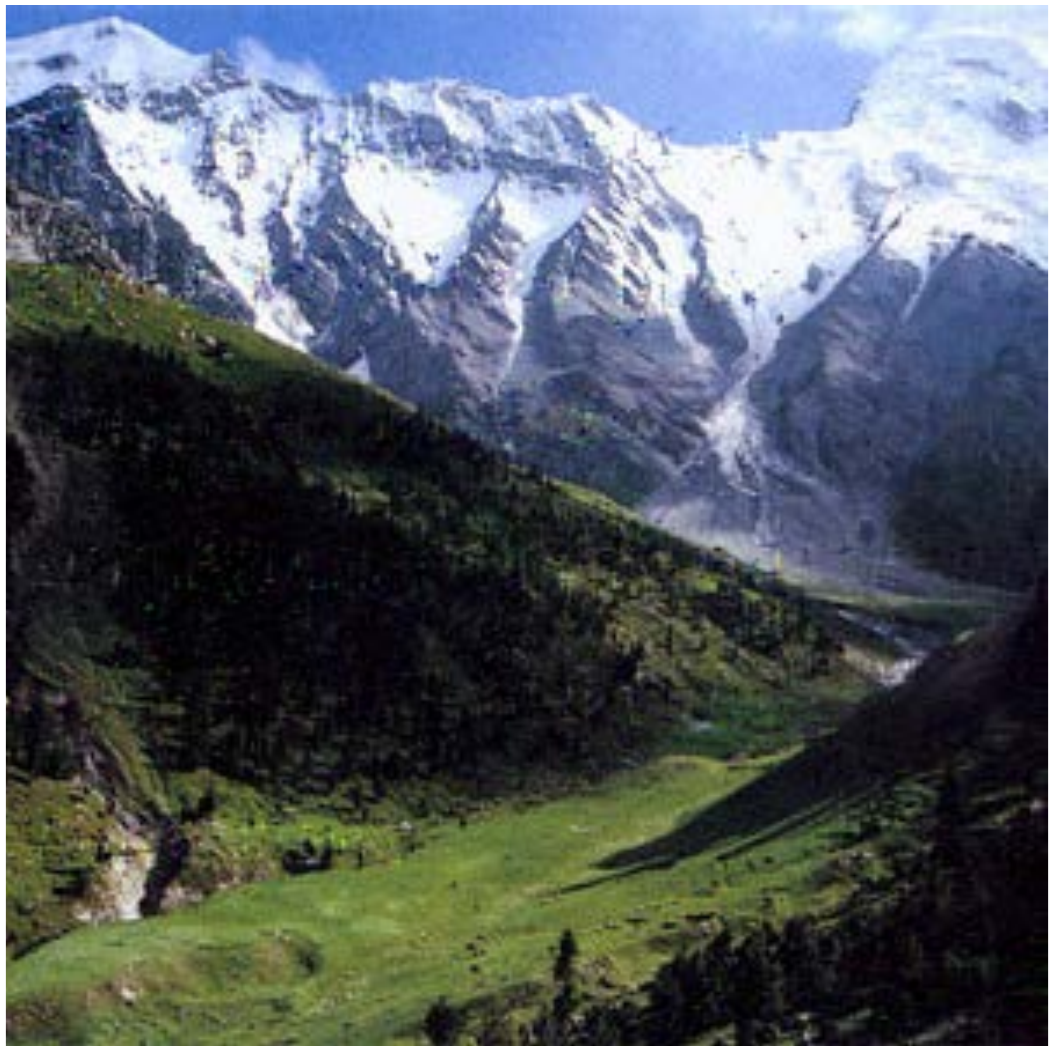}
		\caption{APGL}
	\end{subfigure}
	\begin{subfigure}[b]{0.118\textwidth}
		\centering
		\includegraphics[width=\textwidth]{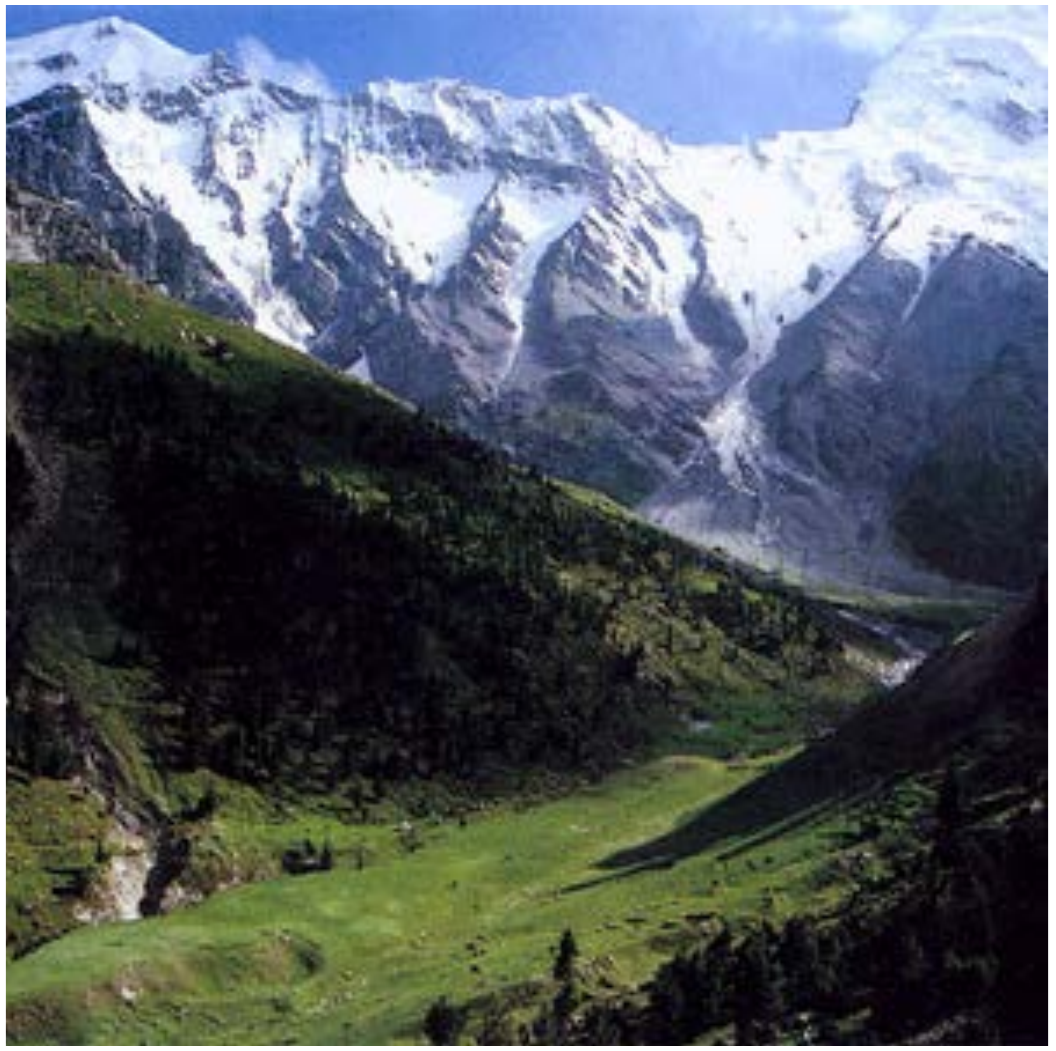}
		\caption{IRNN-$L_p$}
	\end{subfigure}	
	\caption{\small{Comparison of image recovery on more images. (a) Original images. (b) Images with noises. Recovered images by (c) APGL and (d) IRNN-$L_p$. \textbf{Best viewed in $\times 2$ sized color pdf file.}} }\label{fig_imagerecovery2}
	\vspace{-1.5em}
\end{figure}


For the noise free case, we generate the rank $r$ matrix $\M$ as $\M_L\M_R$, where $\M_L\in\mathbb{R}^{150\times r}$, and $\M_R\in\mathbb{R}^{r\times 150}$ are generated by the Matlab command \mcode{randn}. We randomly set $50\%$ elements of $\M$ to be missing. The Augmented Lagrange Multiplier (ALM) \cite{ALMlin} method is used to solve the noise free problem
\begin{equation}\label{noisefreeproblem}
\min_{\X} ||\X||_* \ \text{ s.t. } \ \mathcal{P}_\Omega(\X)=\mathcal{P}_\Omega(\M).
\end{equation}
The default parameters of in the released codes\footnote{Code: \footnotesize {\url{http://perception.csl.illinois.edu/matrix-rank/sample_code.html}.}} of ALM are used. For problem (\ref{pro_mc}), it is solved by IRNN with the parameters $\lambda_0=||\mathcal{P}_\Omega(\M)||_\infty$, $\lambda_t=10^{-5}\lambda_0$ and $\eta=0.7$. The algorithm is stopped when $||\mathcal{P}_\Omega(\X-\M)||_F\leq 10^{-5}$. The matrix recovery performance is evaluated by the Relative Error defined as 
\begin{equation}\label{relerr}
\text{Relative Error}=\frac{||\hat{\X}-\M||_F}{||\M||_F},
\end{equation}
where $\hat{\X}$ is the recovered matrix by different algorithms. If the Relative Error is smaller than $10^{-3}$, then $\hat{\X}$ is regarded as a successful recovery of $\M$. For each $r$, we repeat the experiments $s=100$ times. Then we define the $\text{Frequency of Success}=\frac{\hat{s}}{s}$, where $\hat{s}$ is the times of successful recovery.
We also vary the underlying rank $r$ of  $\M$ from 20 to 33 for each algorithm. We show the frequency of success in Figure \ref{fig_randrecov_noiseless}. The legend IRNN-$L_p$ in Figure \ref{fig_randrecov_noiseless} denotes the model  (\ref{pro_mc}) with $L_p$ penalty solved by IRNN. It can be seen that IRNN for (\ref{pro_mc}) with nonconvex rank surrogates  significantly outperforms ALM for (\ref{noisefreeproblem}) with convex rank surrogate. This is because the nonconvex surrogates approximate the rank function much better than the convex nuclear norm. This also verifies that our IRNN achieves good solutions of (\ref{pro_mc}), though its optimal solutions are in general not computable.

For the second task, we assume that the observed matrix $\M$ is noisy. It is generated by $\mathcal{P}_\Omega(\M)=\mathcal{P}_\Omega(\M_L\M_R)$+0.1$\times$\mcode{randn}. We compare IRNN for (\ref{pro_mc}) with convex  Accelerated Proximal Gradient with Line search (APGL)\footnote{Code: \footnotesize {\url{http://www.math.nus.edu.sg/~mattohkc/NNLS.html}.}} \cite{toh2010accelerated} which solves the noisy problem
\begin{equation}\label{noisypro}
\min_{\X} \lambda||\X||_*+\frac{1}{2}||\mathcal{P}_\Omega(\X)-\mathcal{P}_\Omega(\M)||_F^2.
\end{equation}
For this task, we set $\lambda_0=10||\mathcal{P}_\Omega(\M)||_\infty$ and $\lambda_t=0.1\lambda_0$ in IRNN. We run the experiments for 100 times and the underlying rank $r$ is varying from 15 and 35. For each test, we compute the relative error in (\ref{relerr}). Then we show the mean relative error over 100 tests in Figure \ref{fig_randrecov_noise}. Similar to the noise free case, IRNN with nonconvex rank surrogates achieves much smaller recovery error than APGL for convex problem (\ref{noisypro}). 

It is worth mentioning that though Logarithm seems to perform better than other nonconvex penalties for low rank matrix completion from Figure \ref{fig_randrecov}. It is still not clear which one is the best rank surrogate since the obtained solutions are not globally optimal. Answering this question is beyond  the scope of this work.

Figure \ref{fig_randrecov_time} shows the running times of the compared methods. It can be seen that IRNN is slower than the convex ALM. This is due to the reinitialization of IRNN when using the continuation technique. Figure \ref{fig_randrecov_conv} plots the objective function values in each iterations of IRNN with different nonconvex penalties. As verified in theory, it can be seen that the values are decreasing. 


\begin{figure*}
	\begin{subfigure}[b]{0.327\textwidth}
		\centering
		\includegraphics[width=\textwidth]{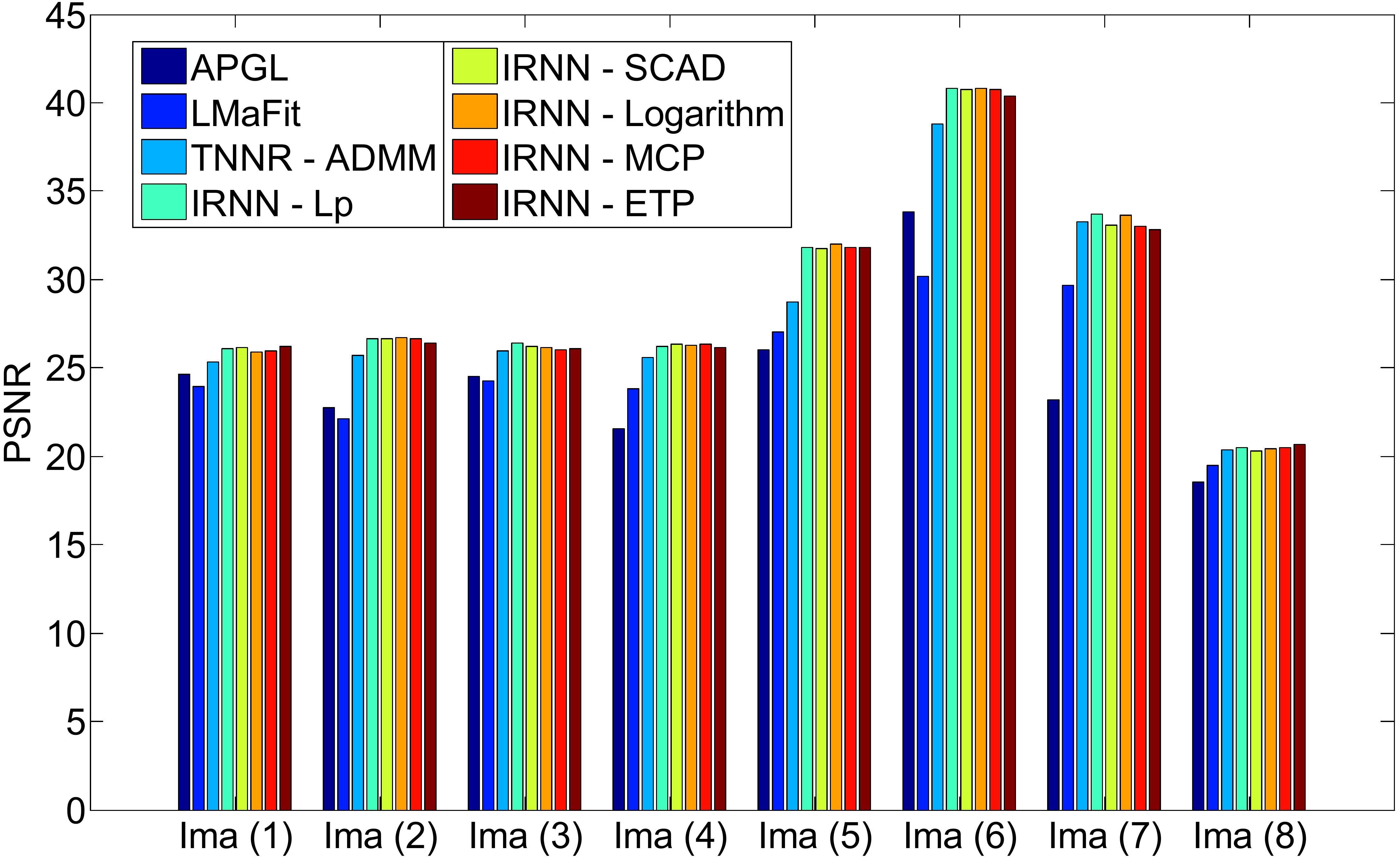}
		\caption{PSNR values}
		\label{fig_img_psnr}
	\end{subfigure}
	\begin{subfigure}[b]{0.32\textwidth}
		\centering
		\includegraphics[width=\textwidth]{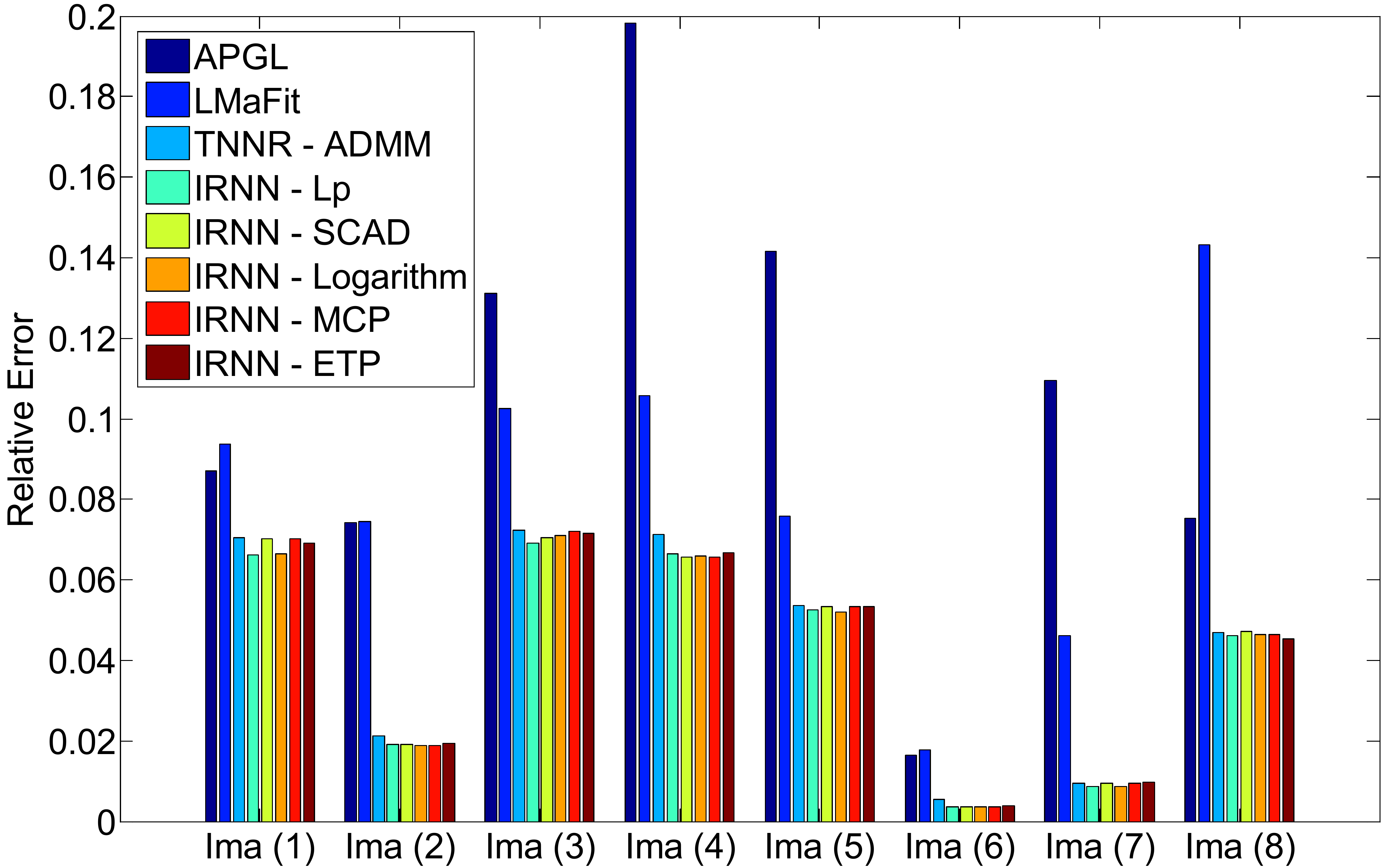}
		\caption{Relative error}
		\label{fig_img_err}
	\end{subfigure}
	\begin{subfigure}[b]{0.31\textwidth}
		\centering
		\includegraphics[width=\textwidth]{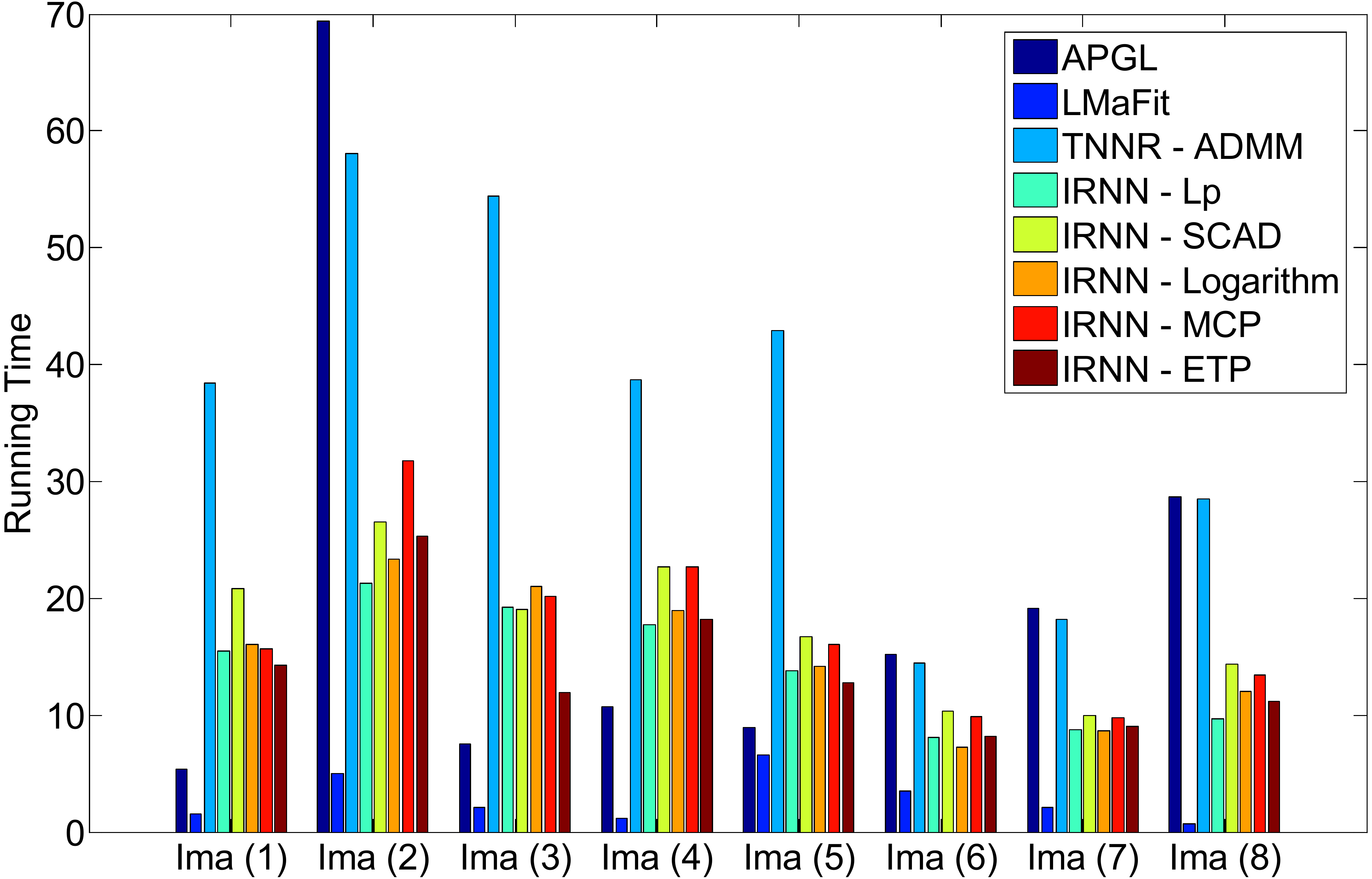}
		\caption{Running time}
		\label{fig_img_time}
	\end{subfigure}
	\caption{{Comparison of (a) PSNR values; (b) Relative error; and (c) Running time (seconds) for image recovery by different matrix completion methods.}}\label{fig_imagerecov_psnr_err}
\end{figure*}

\subsection{Application to Image Recovery}
In this section, we apply the low rank matrix completion models (\ref{noisypro}) and (\ref{eq_genpro}) for image recovery. We follow the experimental settings in \cite{hu2012fast}. Here we consider two types of noises on the real images. The first one replaces $50\%$ of pixels with random values (sample image (1) in Figure \ref{fig_gaussiannoise}). The other one adds some unrelated texts on the image (sample image (2) in Figure \ref{fig_gaussiannoise}). 
The goal is to remove the noises by using low rank matrix completion. Actually, the real images may not be of low-rank. But their top singular values dominate the main information. Thus, the image can be approximately recovered by a low-rank matrix. For the color image, there are three channels. Matrix completion is applied for each channel independently. 
We compare IRNN with some state-of-the-art methods on this task, including APGL, Low-Rank Matrix Fitting (LMaFit)\footnote{Code: \footnotesize{\url{http://lmafit.blogs.rice.edu/}.}} \cite{wen2012solving} and Truncated Nuclear Norm Regularization (TNNR)\footnote{Code:  \footnotesize{\url{https://sites.google.com/site/zjuyaohu/}.}} \cite{hu2012fast}. For the obtained solution, we evaluate its quality by the Peak Signal-to-Noise Ratio (PSNR) and the relative error (\ref{relerr}).

Figure \ref{fig_imagerecovery} (c)-(g) show the recovered images by different methods. It can be seen that our IRNN method for nonconvex models achieve much better recovery performance than APGL and LMaFit. The performances of low rank models (\ref{eq_genpro}) using different nonconvex surrogates are quite similar, so we only show the results by IRNN-$L_p$ and IRNN-SCAD due to the limit of space. Some more results are shown in Figure \ref{fig_imagerecovery2}. Figure \ref{fig_imagerecov_psnr_err} shows the PSNR values, relative errors and running time of different methods on all the tested images. It can be seen that IRNN with all the evaluated nonconvex functions achieves higher PSNR values and smaller relative error. This verifies that the nonconvex penalty functions are effective in this situation. The nonconvex truncated nuclear norm is close to our methods, but its running time is 3$\sim$5 times of ours.

  \begin{figure}
  	\centering
  	\begin{subfigure}[b]{0.48\textwidth}
  		\centering 
  		\includegraphics[width=\textwidth]{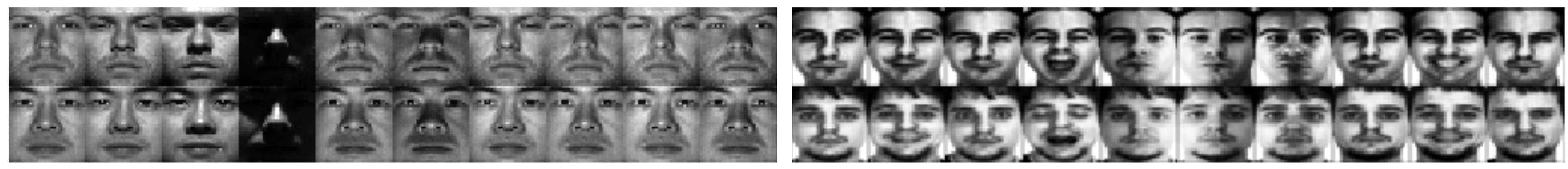} 
  		\caption{\small}
  	\end{subfigure}
  	\begin{subfigure}[b]{0.48\textwidth}
  		\centering
  		\includegraphics[width=\textwidth]{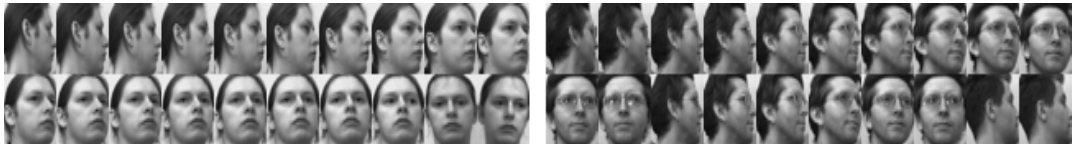} 
  		\caption{\small}
  	\end{subfigure} 
  	\caption{\small Some example face images from (a) Extended Yale B and (b) UMIST databases.}\label{fig_faceimages}
  \end{figure}
 
\subsection{Tensor Low-Rank Representation}

In this section, we consider to use the Tensor Low-Rank Representation (TLRR) (\ref{tlrr}) for face clustering \cite{LRR,liu2011latent}. Problem (\ref{tlrr}) can be solved by the Accelerated Proximal Gradient (APG) \cite{beck2009fast} method with the optimal convergence rate $O(1/K^2)$, where $K$ is the number of iterations. The corresponding Nonconvex TLRR (NTLRR) related  to (\ref{tlrr}) is
\begin{equation}\label{ntlrr}
\min_{\PP_j\in\mathbb{R}^{m_j\times m_j}} \sum_{j=1}^p\sum_{i=1}^{m_j}g(\sigma_i(\mathbf{\PP}_j))+\frac{1}{2}\left\|\Xt-\sumj\Xt\times_j{\PP}_j\right\|_F^2,
\end{equation}
where we use the Logarithm function $g$ in Table \ref{tab_nonpenlty}, since we find it achieves the best performance in the previous experiments. Problem (\ref{ntlrr}) has more than one block of variable, and thus it can be solved by IRNN-PS.

In this experiment, we use TLRR and NTLRR for face clustering. Assume that we are given $m_3$ face images from $k$ subjects with size $m_1\times m_2$. Then we can construct an 3-way tensor $\Xt\in\mathbb{R}^{m_1\times m_2\times m_3}$. After solving (\ref{tlrr}) or (\ref{ntlrr}), we follow the settings in \cite{LRR} to construct the affinity matrix by $\mathbf{W}=(|\PP_3|+|\PP_3^T|)/2$. Finally, the Normalized Cuts (NCuts) \cite{NormaCut} is applied based on $\mathbf{W}$ to segment the data into $k$ groups. 

Two challenging face databases, Extended Yale B \cite{YaleBdatabase} and UMIST\footnote{\url{http://www.cs.nyu.edu/~roweis/data.html}.}, are used for this test. Some sample face images are shown in Figure \ref{fig_faceimages}. Extended Yale B consists of 2,414 frontal face images of 38 subjects under various lighting, poses and illumination conditions. Each subject has 64 faces. We construct two clustering tasks based on the first 5 and 10 subjects face images of this database. The UMIST database contains 564 images of 20 subjects,
each covering a range of poses from profile to frontal views. All the images in UMIST are used for clustering. For both databases, the images are resized into $m_1\times m_2=28\times 28$. 

Table \ref{tab_fc} shows the face clustering accuracies of NTLRR, compared with LRR, LatLRR and TLRR. The performances of LRR and LatLRR are consistent with previous work \cite{LRR,liu2011latent}. Also, it can be seen that TLRR achieve better performance than LRR and LatLRR, since it exploits the inherent spatial structures among samples. More importantly, NTLRR futher improves TLRR. Such an improvement is similar to those in previous experiments, though the support in theory is still open.
\begin{table}[t]
	\centering
	\caption{Face clustering accuracy ($\%$) on Extended Yale B and UMIST databases.}
	\begin{tabular}{c|c|c|c|c}
		\hline
		&  LRR  & LatLRR & TLRR  &     NTLRR      \\ \hline
		YaleB 5  & 83.13 & 83.44  & 92.19 & \textbf{95.31} \\ \hline
		YaleB 10 & 62.66 & 65.63  & 66.56 & \textbf{67.19} \\ \hline\hline
		UMINST  & 54.26 & 54.09  & 56.00 & \textbf{58.09} \\ \hline
	\end{tabular}	\label{tab_fc}
\end{table}

\section{Conclusions and Future Work}\label{sec_con}
This work targeted for nonconvex low rank matrix recovery by applying the nonconvex surrogates of $L_0$-norm on the singular values to approximate the rank function. We observed that all the existing nonconvex surrogates are concave and monotonically increasing on $[0,\infty)$. Then we proposed a general solver IRNN to solve the nonconvex nonsmooth low rank minimization problem (\ref{eq_genpro}). We also extend IRNN to solve problem (\ref{eq_genpro2}) with multi-blocks of variables. In theory, we proved that any limit point is a stationary point. Experiments on both synthetic data and real data demonstrated that IRNN usually outperforms the state-of-the-art convex algorithms.

There are some interesting future work. First, it is still unclear which nonconvex surrogate is the best. It is possible to provide some support in theory under some conditions. Second, one may consider to use the alternating direction method of multiplier to solve the nonconvex problem with the affine constraint and to prove the convergence. 
Second, one may consider to solve the following problem by IRNN
\begin{equation}\label{pro_stru}
\min_{\X} \sum_{i=1}^m g(h(\sigma_i(\X)))+f(\X),
\end{equation}
when $g(y)$ is concave and the following problem
\begin{equation}
\min_{\X } w_ih(\sigma_i(\X))+||\X-\mathbf{Y}||_F^2,
\end{equation}
can be cheaply solved. An interesting application of (\ref{pro_stru}) is to extend the group sparsity on the singular values. By dividing the singular values into $k$ groups, i.e., $G_1=\{1,\cdots,r_1\}$, $G_2=\{r_1+1,\cdots,r_1+r_2-1\}$, $\cdots$, $G_k=\{\sum_i^{k-1}r_i+1,\cdots,m\}$, where $\sum_ir_i=m$, we can define the group sparsity on the singular values as $||\X||_{2,g}=\sum_{i=1}^kg(||\mathbf{\sigma}_{G_i}||_2)$. This is exactly the first term in (\ref{pro_stru}) by letting $h$ be the $L_2$-norm of a vector. $g$ can be nonconvex functions satisfying the assumption \textbf{A1} or specially the absolute convex function.

\section*{Acknowledgements}

This research is supported by the Singapore National Research Foundation under its International Research Centre @Singapore Funding Initiative and administered by the IDM Programme Office. Z. Lin is supported by NSF of China (Grant nos. 61272341, 61231002, and 61121002) and MSRA.

{ \small
\bibliographystyle{IEEEbib}
\bibliography{IRNN}
}

\textbf{Canyi Lu} received the bachelor degree in mathematics from the Fuzhou University in 2009, and the master degree in the pattern recognition and intelligent system from the University of Science and Technology of China in 2012. He is currently a Ph.D. student with the Department of Electrical and Computer Engineering at the National University of Singapore. His current research interests include computer vision, machine learning, pattern recognition and optimization. He was the winner of the Microsoft Research Asia Fellowship 2014.

\textbf{Jinhui Tang} is currently a Professor of School of Computer Science and Engineering, Nanjing University of Science and Technology. He received his B.E. and Ph.D. degrees in July 2003 and July 2008 respectively, both from the University of Science and Technology of China (USTC). From July 2008 to Dec. 2010, he worked as a research fellow in School of Computing, National University of Singapore. During that period, he visited School of Information and Computer Science, UC Irvine, from Jan. 2010 to Apr. 2010, as a visiting research scientist. From Sept. 2011 to Mar. 2012, he visited Microsoft Research Asia, as a Visiting Researcher. His current research interests include multimedia search, social media mining, and computer vision. He has authored over 100 journal and conference papers in these areas. He serves as a editorial board member of  Pattern Analysis and Applications, Multimedia Tools and Applications, Information Sciences, and Neurocomputing. Prof. Tang is a recipient of ACM China Rising Star Award in 2014, and a co-recipient of the Best Paper Award in ACM Multimedia 2007, PCM 2011 and ICIMCS 2011. He is a senior member of IEEE and a member of ACM.

\textbf{Shuicheng Yan}
	is currently an Associate Professor at the Department of Electrical and Computer Engineering at National University of Singapore, and the founding lead of the Learning and Vision Research Group (http://www.lv-nus.org). Dr. Yan's research areas include machine learning, computer vision and multimedia, and he has authored/co-authored hundreds of technical papers over a wide range of research topics, with Google Scholar citation $>$19,000 times and H-index 60. He is ISI Highly-cited Researcher, 2014 and IAPR Fellow 2014. He has been serving as an associate editor of IEEE TKDE, TCSVT and ACM Transactions on Intelligent Systems and Technology (ACM TIST). He received the Best Paper Awards from ACM MM'13 (Best Paper and Best Student Paper), ACM MM’12 (Best Demo), PCM'11, ACM MM’10, ICME’10 and ICIMCS'09, the runner-up prize of ILSVRC'13, the winner prize of ILSVRC’14 detection task, the winner prizes of the classification task in PASCAL VOC 2010-2012, the winner prize of the segmentation task in PASCAL VOC 2012, the honourable mention prize of the detection task in PASCAL VOC'10, 2010 TCSVT Best Associate Editor (BAE) Award, 2010 Young Faculty Research Award, 2011 Singapore Young Scientist Award, and 2012 NUS Young Researcher Award.

\textbf{Zhouchen Lin}
	received the Ph.D. degree in Applied Mathematics from Peking University, in 2000. He is currently a Professor at Key Laboratory of Machine Perception (MOE), School of Electronics Engineering and Computer Science, Peking University. He is also a Chair Professor at Northeast Normal University and a Guest Professor at Beijing Jiaotong University. Before March 2012, he was a Lead Researcher at Visual Computing Group, Microsoft Research Asia. He was a Guest Professor at Shanghai Jiaotong University and Southeast University, and a Guest Researcher at Institute of Computing Technology, Chinese Academy of Sciences. His research interests include computer vision, image processing, computer graphics, machine learning, pattern recognition, and numerical computation and optimization. He is an Associate Editor of IEEE Trans. Pattern Analysis and Machine Intelligence and International J. Computer Vision, an area chair of CVPR 2014, ICCV 2015, NIPS 2015 and AAAI 2016, and a Senior Member of the IEEE.

\end{document}